%% file: arxiv.tex
\documentclass{article}

\usepackage[final]{template}
\usepackage{graphicx}
\usepackage{multirow}

\usepackage[utf8]{inputenc} 
\usepackage[T1]{fontenc}    
\usepackage{hyperref}       
\usepackage{url}            
\usepackage{booktabs}       
\usepackage{amsfonts,amsmath,dsfont}       
\usepackage{nicefrac}       
\usepackage{microtype}      
\usepackage{xcolor}         
\usepackage{verbatim}
\usepackage{algorithm}
\usepackage{algorithmic}
\usepackage{colortbl}

\newtheorem{proposition}{Proposition}
\newtheorem{theorem}{Theorem}

\newtheorem{lemma}{Lemma}

\newtheorem{remark}{Remark}

\newtheorem{assumption}{Assumption}
\newtheorem{corollary}{Corollary}

\title{\vspace{-0.1in} 
Learning under Commission and Omission Event Outliers
\vspace{-0.1in}}

\author{%
 Yuecheng Zhang ~~ Guanhua Fang ~~ Wen Yu\\
Fudan University, 220 Handan Rd., Yangpu District, Shanghai, China
}

\input{macro}

\begin{document}

\bibliographystyle{plainnat}
\maketitle

\begin{abstract}
Event stream is an important data format in real life. 
The events are usually expected to follow some regular patterns over time.
However, the patterns could be contaminated by unexpected absences or occurrences of events.
In this paper, we adopt the temporal point process framework for learning event stream and we provide a simple-but-effective method to deal with both commission and omission event outliers.
In particular, we introduce a novel weight function to dynamically adjust the importance of each observed event so that the final estimator could offer multiple statistical merits.
We compare the proposed method with the vanilla one in the classification problems, where event streams can be clustered into different groups. Both theoretical and numerical results confirm the effectiveness of our new approach.
To our knowledge, our method is the first one to provably handle both commission and omission outliers simultaneously.

\end{abstract}

\section{Introduction}


In modern user recommender systems, the data collected for each individual can be viewed as an irregular time-stamped event sequence/stream.
For example, in e-commerce \citep{xu2014path}, the merchants want to recommend related items to customers based on their activity history. The actions taken by a customer in viewing and purchasing the items on the website can form an event sequence. 
In electronic health \citep{enguehard2020neural}, patients with similar behaviors may be given similar treatment. 
The messages sent by a patient through an AI medical assistant can be viewed as a sequence of events.
In online testing \citep{xu2018latent}, the students take steps to answer the interactive problem-solving questions on the computer, and their response history can be treated as an event stream. The educators can make accurate diagnoses based on students' responses.
In mobile music service \citep{carneiro2011towards}, 
the users can search and play different song tracks. Their listening history will be recorded and hence be treated as an event sequence. Then the music apps can recommend favourite songs to each user. 
Event sequence data is more complicated than the panel data and contains a lot of individual-level information. 
It is of great interest to classify distinct event sequences to different groups, which can be useful for personalized or group-wise treatment, diagnosis, and recommendation \citep{hosseini2017recurrent, wang2021modeling, cao2021deep}.

In the literature, the existing methods on event sequence clustering can be mainly divided into two categories, namely metric-based clustering and model-based clustering. 
The methods in the former category \citep{berndt1994using, pei2013clustering} measure the similarity or dissimilarity among pairs of event sequences based on extracted features or pre-specified metrics.
The methods in the second category \citep{xu2017dirichlet, yin2021row, zhang2022learning} adopt a temporal point process (TPP) framework, where the event sequences are assumed to follow a mixture of point process models.
However, these works do not take into account the outlier events, which frequently happen in real-world scenarios. For example, in an e-commerce platform, a customer purchases grocery products every weekend. An outlier event happens if he/she forgets to shop on a particular weekend.
In electronic health, the patient who suffers from chronic disease may buy some prescribed medicines on a regular schedule. An outlier event happens if he/she suddenly has an acute disease. 
In online testing, the student follows some proper strategies to solve the challenging questions. An outlier happens if he/she takes some unexpected actions which leads to wrong answers.
In the mobile music apps, the user regularly plays the song tracks. An outlier event happens when he/she listens to music that was disliked in the past.

In this work, we tackle the event sequence data with potential event outliers.
In \cite{liu2021event}, they introduce two types of outliers that may exist in the event streams.
The first type is called ``omission" that an event, supposed to happen, is missing or overdue.
The second type is called ``commission" that the unexpected events happen.
Our main task is to provide a unified framework for handling the event stream with the presence of outliers. 
The goals include the following four parts.
(i) The methodology is simple-but-effective and it could be integrated with any existing TPP models with low additional computational cost.
(ii) The proposed algorithm can return robust estimation results when a large number of outliers are allowed.
(iii) The algorithm can provide more consistent clustering results. 
(iv) The method can also detect the event outliers as a side product.

Our solution is as follows. 
Given a pre-determined $K$-mixture TPP model, we modify it by introducing weight to each observed event. 
A lower (higher) weight means the event is more likely (unlikely) to be an outlier.
The weight function is carefully constructed through an influence function $\phi(x)$ which has several important properties. 
Firstly, $\phi(x)$ has twice continuous derivative that allows for easy computation.
Secondly, the derivative $\phi'(x)$ has a bounded support set that makes the algorithm robust and has a high break-down point. 
Thirdly, $\phi'(x)$ can properly balance the weights for both omission outliers and commission outliers at the same time.
After re-weighting the TPP model, we then alternatively update the latent class labels and the model parameters. 
In the computation of the (approximate) posterior of latent labels, we use the \textit{overall} weight function that combines all $K$ class information.
When updating the model parameters, we use the \textit{class-specific} weight function that can lead to a better gradient direction.
Moreover, we also allow additional tuning parameters in the construction of weight functions.
The tuning parameters can adjust the overall impact of the re-weighting procedure so that sufficient event information is reserved and the non-identifiable situations can be avoided.

We want to point out a few key differences between our method and those in the existing literature. 
\cite{zhang2024robust} recently propose a robust clustering algorithm that can consistently perform better than the classical methods. However, their setting is different in the sense that they treat the whole event sequence as the outlier. By contrast, our setting, treating the outlier on the event level, is more general and flexible. 
\cite{zhang2021learning2} consider a situation where only commission outliers are allowed. They propose a best subset selection method to detect the potential outliers. Unfortunately, such a method is obviously NP-hard and cannot be used for omission outliers. 


Our technical contributions can be summarized as follows.
(a) The proposed methodology is easy-to-implement and can be integrated into any existing (parametric, nonparametric, neural network-based) TPP models. The additional computational cost lies in the calculation of weight function which scales linearly with the sample size. 
(b) Although the proposed method is simple, it is shown to work adaptively. To be specific, when there are no outliers, the algorithm behaves similarly to the original one with no weight modification. 
When there exist outliers, the algorithm can still return the asymptotically consistent estimator.  
(c) We provide a relatively complete theory including, a non-asymptotic local convergence result,
theoretical explanations for the advantages of using weight functions, and the detection guarantee for outlier events.
A brief summary is given in Table \ref{tab:thm}.
Here we would also like to point out that our new weight functions
could be applied to different tasks of learning event streams, including but not limited to, the classification of event sequences, the prediction of the next event arrival, the change point detection of user behaviors, etc.

\begin{minipage}{\textwidth}
\centering
\begin{minipage}[t]{0.9\textwidth}
\makeatletter\def\@captype{table}
\centering
\scalebox{0.9}{
\begin{tabular}{lll}
        \hline
        \hline
         Scope & Result & Explanation  \\
        \hline
        \multirow{3}{*}{Impact of weights} & Theorem \ref{Thm:Mgrad} & consistency with the absence of outliers \\
         & Theorem \ref{Thm:Mout} & robustness with the presence of outliers \\
         & Theorem \ref{cor:grad} & smaller gradient bias \\
        \hline
       Parameter estimates  & Theorem \ref{thm:local} & a local linear convergent rate \\
        \hline 
       Outlier detection & Theorem \ref{thm:tpr} & guarantees of detecting outliers \\
        \hline 
        \hline
    \end{tabular}\label{tb:thm}
    }
\caption{{ An overview of our theoretical results. "Scope": the theoretical aspect that we want to investigate; "Result": the detailed theorem number; "Explanation": the main conclusions or messages from the theorem.}}
\label{tab:thm}
\end{minipage}
\end{minipage}

The rest of the paper is organized as follows. 
In Section \ref{sec:2}, we review the definitions of event sequence data and TPP models. We also discuss the related works and potential challenges.
The main methodology and the detailed algorithm are described in Section \ref{sec:3}. 
Theoretical analyses are provided in Section \ref{sec:4} to help readers to better understand each component of our new method.
Simulation studies and real data applications are given in Section \ref{sec:5} and Section \ref{sec:6}, respectively, to show the effectiveness of the proposed method.
Finally, a concluding remark is given in Section \ref{sec:7}.

\textbf{Notation}.
In this paper, we use $\mathbb E$ and $\mathbb P$ to denote the generic expectation and probability
and use $[N] = \{1, 2, ..., N\}$ for any positive integer. 
Sub-scripts $n$, $i$, and $k$ are referred to the index of the event sequence, event number, and the class label, respectively.
We say $a_n = O(b_n), O_p(b_n) (\text{or}~\Theta(b_n), \Theta_p(b_n))$ if there exists a constant $c$ that $a_n \leq c b_n$ (or $\frac{1}{c} a_n \leq b_n \leq c a_n$) holds or holds with high probability. Symbol $\tilde O, \tilde O_p$ hides all logarithmic terms. 
We use $\lambda(t) \downarrow 0$ (or $\lambda(t) \uparrow \infty$) to represent that a sequence $\{\lambda_n\}$ satisfies $\lambda_n (t) = c_n \cdot \lambda(t)$ with $c_n \rightarrow 0$ (or $c_n \rightarrow \infty$).

\section{Preliminary}\label{sec:2}

\subsection{Data Description}

We observe the following event sequences, $\Big\{ (t_{n,1}, ..., t_{n,i}, ..., t_{n, M_n}); n = 1, ..., N \Big\}$, where $t_{n,i}$ is the $i$-th event time stamp of the $n$-th sequence, $M_n$ is the number of events observed for sequence $n$, and $N$ is the total number of event sequences. 
For the notional simplicity, we may use $S_n$ to denote the $n$-th observed sequence, i.e., $S_n = (t_{n,1}, ..., t_{n,i}, ..., t_{n, M_n})$.
Then, the whole dataset becomes $\boldsymbol{S}=\left\{S_n\right\}_{n=1}^N$.
Moreover, all time sequences are observed within the time horizon $[0, T_0]$, i.e., $0 \leq t_{n,i} \leq T_0$ for all $i$ and $n$.
We further assume that total the time horizon can be divided into $L$ time periods, i.e., $T_0 = L \cdot T$, where $T$ is the length of a single time period.
To help readers to have more intuition, a real data example is given in Table \ref{tab:example1}, which shows the event stream sequence of a randomly selected user from the internet protocol television (IPTV) data.

\begin{minipage}{\textwidth}
\centering
\begin{minipage}[t]{0.7\textwidth}
\makeatletter\def\@captype{table}
\centering
\scalebox{0.9}{
\begin{tabular}{ccc}
        \toprule
          & id & time \\
        \midrule
        1 & 65659245 & 2012/01/01 15:37:15 \\
        2 & 65659245 & 2012/01/01 16:54:40 \\
        $\cdots$   & $\cdots$   & $\cdots$   \\
        2760 & 65659245 & 2012/11/30 18:37:11 \\
        2761 & 65659245 & 2012/11/30 18:45:09 \\
        \bottomrule
    \end{tabular}
    }
\caption{{ IPTV dataset. "id": user identifier. "time": the time stamp when the user started to watch a TV program. All events are recorded during the period, 2012.1.1 - 2012.11.30. Total time horizon $T_0 = $  336 days, $T = $ 7 days (a week), and $L = 48$.}}
\label{tab:example1}
\end{minipage}
\end{minipage}

To mathematically describe the event sequence data, we adopt the TPP methodology \citep{daley2003introduction}, also known as the counting process method or recurrent event analysis \citep{yamaguchi1991event}. 
For any increasing event time sequence $0 < t_1 < t_2 < ... < t_M$, we let $N(t) := \sharp\{i: t_i \leq t \}$ be the number of events observed up to time $t$. Then we can define the conditional intensity function,
$
\lambda^{\ast}(t) := \lim_{dt \rightarrow 0}
\mathbb E[N[t, t+dt) | \mathcal H_t]/dt
$,
where $N[t, t+dt) := N(t+dt) - N(t)$ represents the number of events happening within $[t, t+dt)$ and
$\mathcal H_t := \sigma(\{N(s); s < t\})$ is the history filtration before time $t$.
Intensity $\lambda^{\ast}(t)$ characterizes the dynamic of the event process and is of great importance and practical interest for statistical modelling.


\subsection{Event Outliers}

\begin{figure}[ht!]
	\centering
	\includegraphics[width=0.46\textwidth]{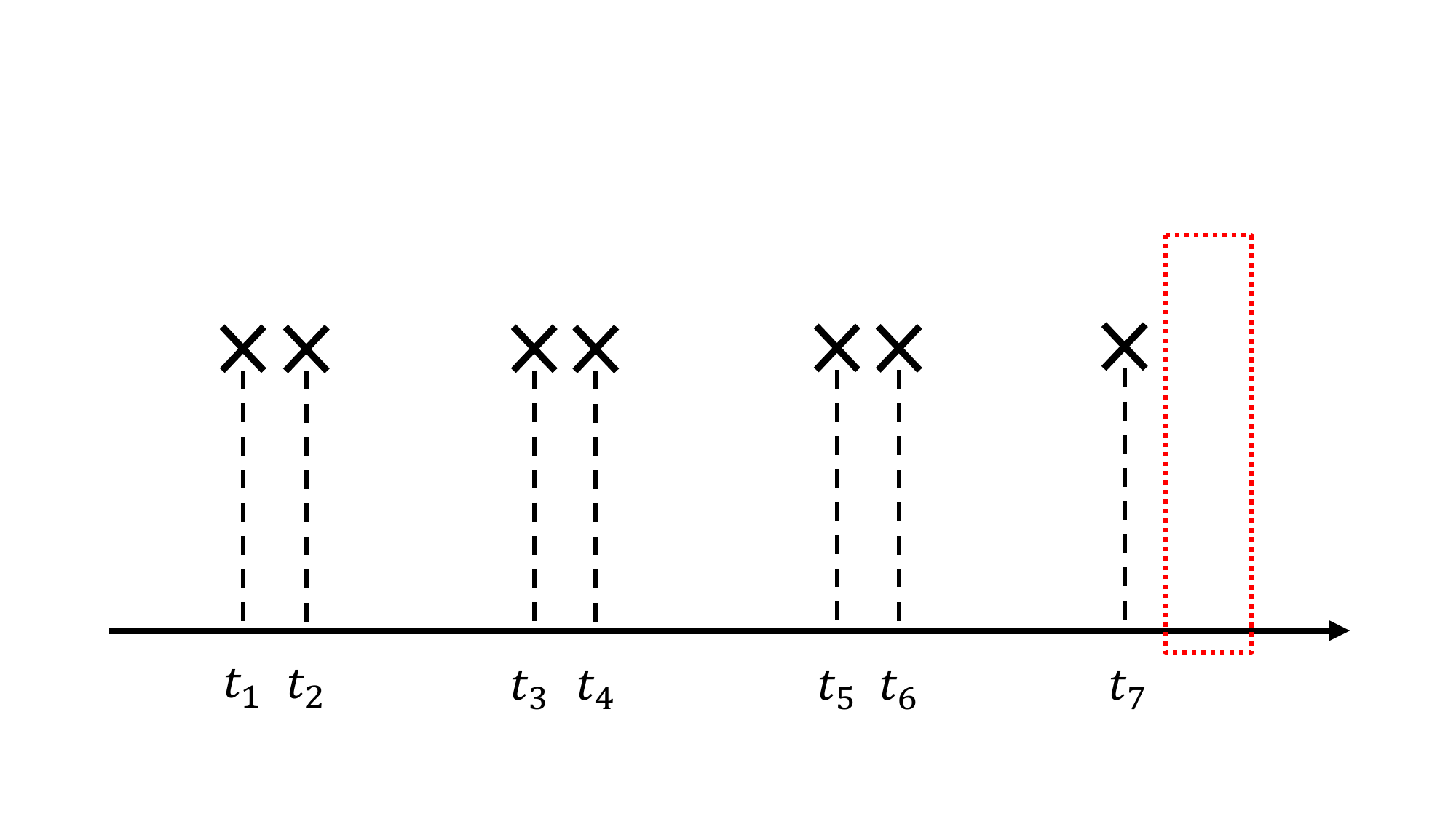}
	\includegraphics[width=0.46\textwidth]{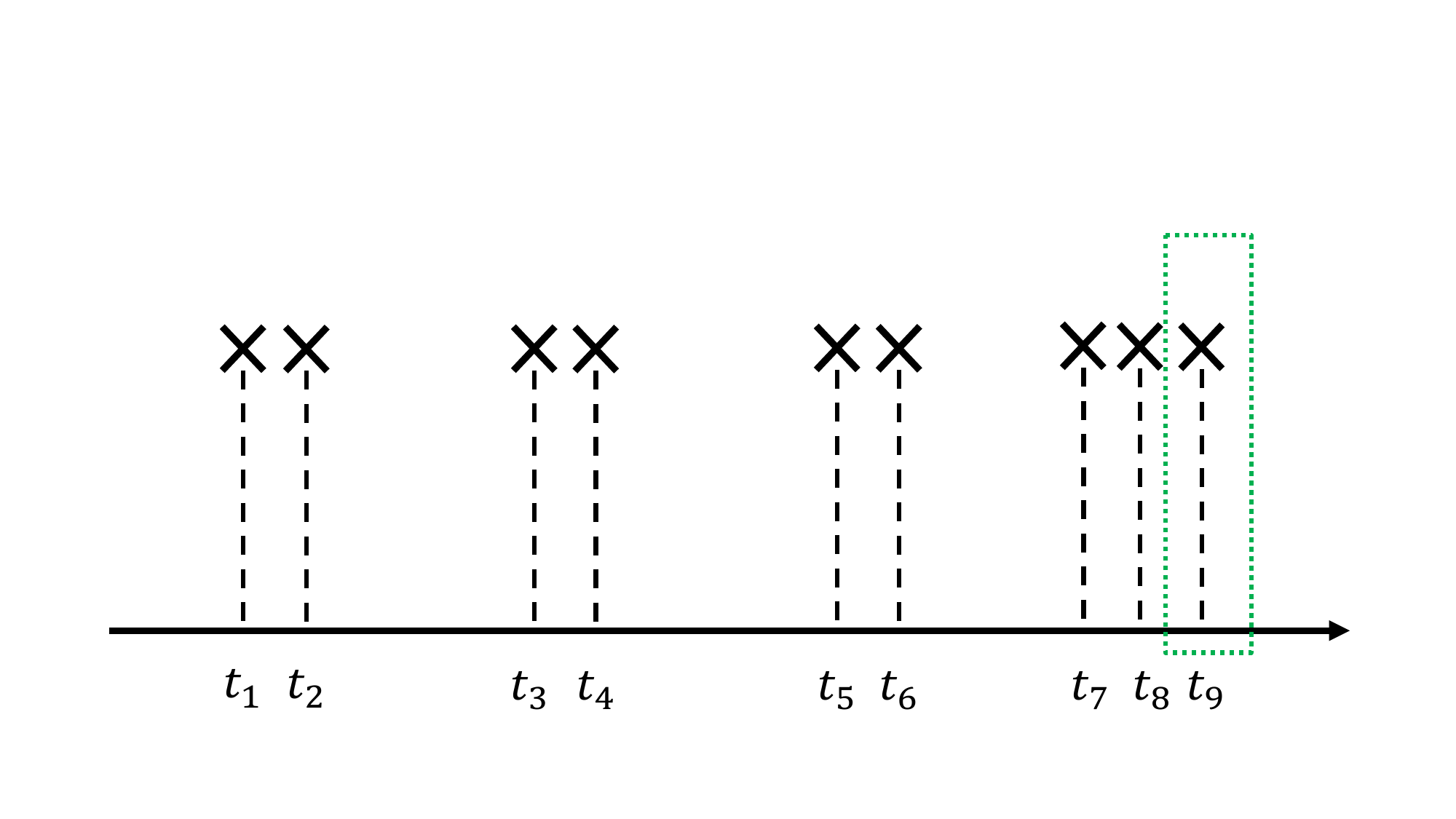} 

 \caption{The visualization of the two types of event outliers. The left plot is for the omission case, where the red box indicates a possible missing event. The right plot is for the commission case, where the green box indicates a potential unexpected event.}\label{fig:outlier}
\end{figure}

In applications, such as disease outbreak detection \citep{buckeridge2007outbreak},  fraud detection \citep{rajeshwari2016real, carcillo2018scarff}, medical error detection \citep{hauskrecht2016outlier, kirkendall2019data}, and network monitoring system \citep{chen2016streaming, cronie2024cross},  unusual occurrences or absence of events frequently happen in real-time event sequences.
\cite{liu2021event} introduce that two types of outliers may arise in continuous-time event
sequences.
The first type is called as ``omission outlier" which refers to the scenario that the event, supposed to happen, is missing or overdue.
The second type is known as ``commission outlier" which describes the situation that the newly happened event is unexpected: it either arrives too early or is not expected to occur at all given the historical information. 
The visualization of the two types of event outliers is given in Figure \ref{fig:outlier}.

Both types of event outliers are related to problems of practical interest.
Consider a patient who suffers from
a chronic disease and takes specific medications on a regular schedule.
An omission event happens if the patient does not take the medicine for unexpectedly long time.
A commission event happens if he/she takes the medicine too early or too frequently compared to the normal schedule.
Consider a TV user who regularly watches the television at home in the evening.
An omission event happens if the user is too busy to watch the TV program in the evening on a particular day.
A commission event happens if the user starts watching TV in the morning or at noon. 

\subsection{Clustering Problem}

The clustering effects exist ubiquitously in the event stream data. 
The individuals can be classified into groups according to whether their corresponding event sequences show similar behaviors or not.
For classifying event sequence data, the existing methods can also be divided into two main categories, distance-based clustering \citep{berndt1994using, bradley1998refining, peng2008distance} and model-based clustering \citep{luo2015multi, xu2017dirichlet, yin2021row, zhang2024robust}.
The former one extracts the features from the sequences to transform the data into the matrix form and then applies classical 
clustering algorithms.
The latter one makes the assumption that event sequences follow some underlying parametric mixture models of point processes so that the log-likelihood can be used as the objective function and EM \citep{dempster1977maximum} or VI algorithm \citep{blei2017variational} could be applied.

However, there is little existing work that takes into account the noisy or outlier events for clustering event streams in the literature.
In the present work, we provide an \textit{easy-to-implement} tool that can make classification results adaptive and robust to both types of event outliers.


\subsection{Challenges}

There are several challenges that may be encountered in developing the new TPP clustering methodology.
In this subsection, we discuss a few related technical difficulties.

In the literature, most works \citep{liu2021event, zhang2023multiple} focus on event outlier detection only, while our main task is to classify different event sequences and to provide the estimation for the latent intensity functions. 
In other words, the proposed algorithm needs to go beyond detecting an event to be the outlier or not and simultaneously return robust and consistent classification and estimation results.

\begin{figure}[ht!]
	\centering
	\includegraphics[width=0.49\textwidth]{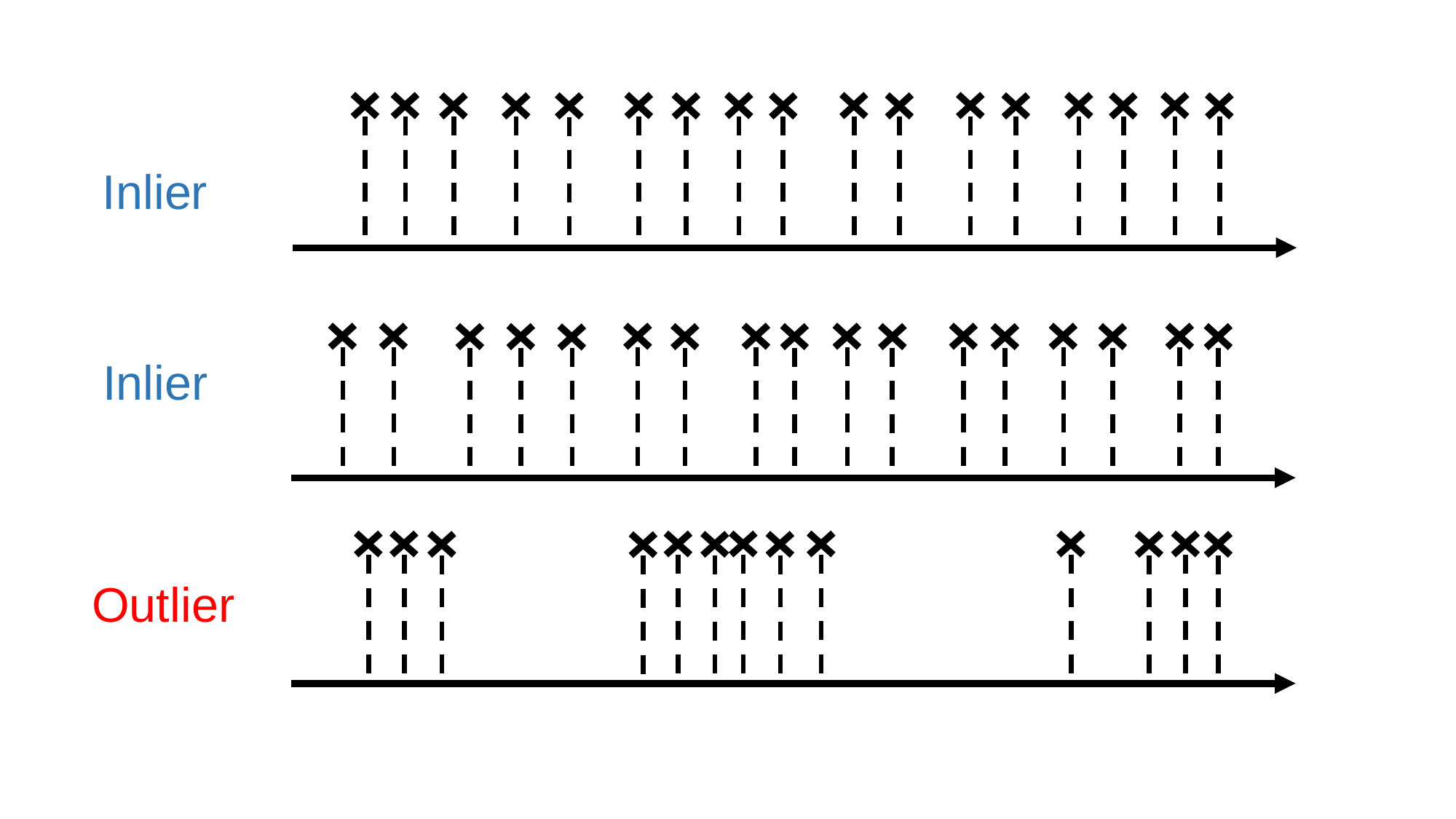}
	\includegraphics[width=0.49\textwidth]{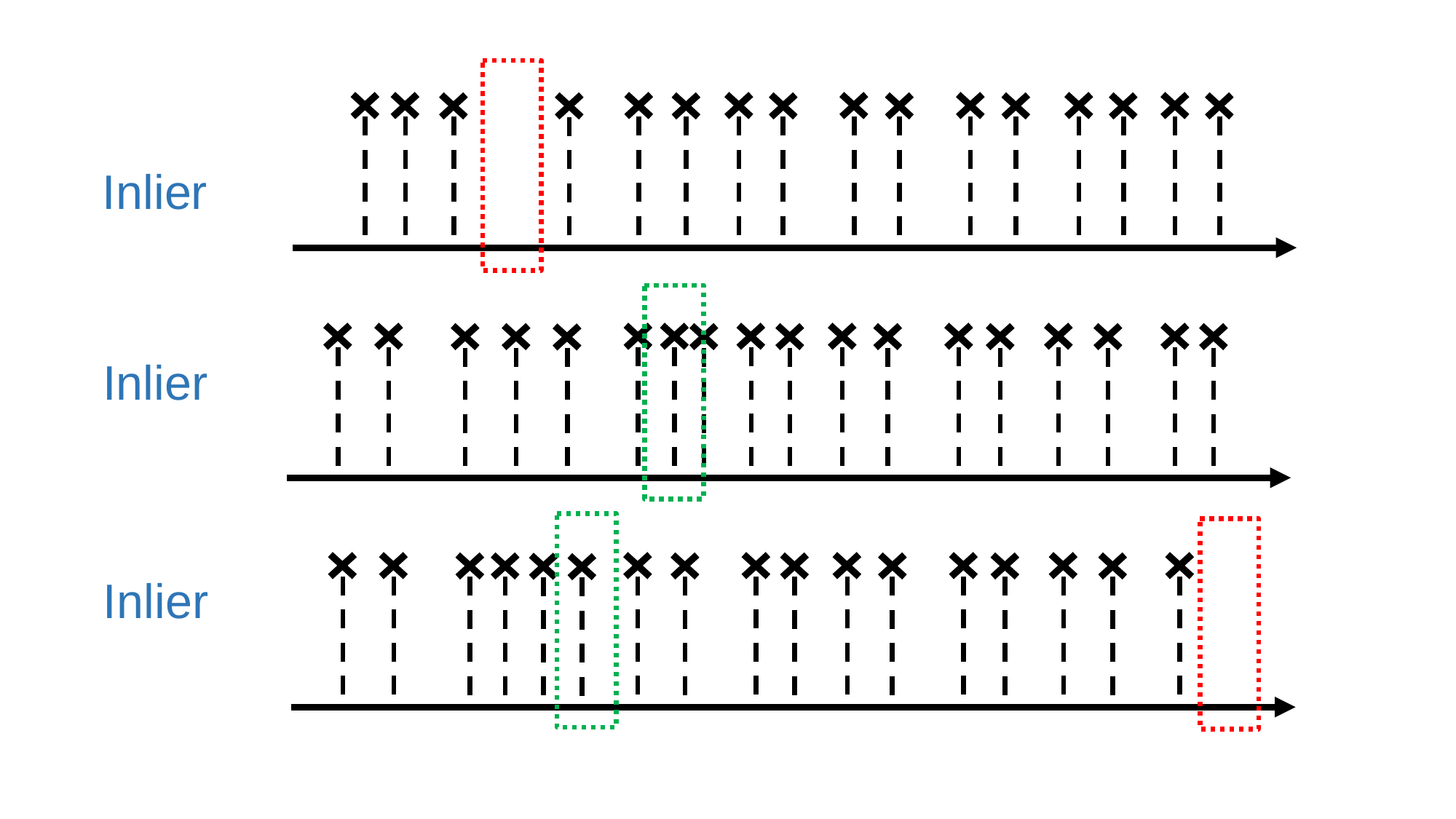} 

 \caption{Left plot: the setting of \cite{zhang2024robust} where the first two event sequences are inliers while the third one is assumed to be the outlier. Right plot: our setting where all three event sequences are inliers but each of them may contain several event outliers.}\label{fig:compare}
\end{figure}

A recent work \citep{zhang2024robust} considers a robust clustering framework allowing the existence of outlier event sequences. 
However such outlier assumption on sequence level is a bit too strong. It may not be practical to treat some entire event sequences as inliers and others as outliers. 
In the present work, we assume every event sequence may have outlier events. 
In other words, our setting is on the event level and is more practically useful. 
Therefore, \cite{zhang2024robust}'s method cannot be directly applied here.
A visualization comparison is given in Figure \ref{fig:compare}.

\cite{zhang2021learning2} introduce a model-based method that can select exogenous events.
However, they only assume the existence of commission outliers but no omission outliers.
Their method is a best subset selection-type algorithm and, hence is NP-hard without convergence guarantees.
Therefore, we need to seek a different type of approach to make it more computationally friendly and more flexible to handle both omission and commission outliers.    

In addition to methodological development, it is also critical to provide guarantees of the clustering results.
Especially, we hope the new method can provably ensure 
(i) robustness: the method works even with a large amount of outliers;
(ii) convergence: the method can at least enjoy fast local convergent speed.
(iii) adaptivity:  the method works effectively whenever there exist outliers or no outliers.
(iv) outlier detection power: the method can return high true positive rates (low false negative rates) of detecting event outliers.

\section{Weighted Clustering Approach}\label{sec:3}

In this section, we propose a weighted clustering algorithm to deal with potentially outlier events. 

\subsection{Framework}\label{sec:mod}

We assume the observed event sequences are generated from a $K$-mixture of temporal point processes with potential event contamination.
To be specific, we let $Z_n \in [K]$ denote the latent label for the $n$-th event sequence. 
In other words, $Z_n = k$ indicates that the $n$-th event sequence belongs to the $k$-th class.
Moreover, its corresponding underlying intensity without event contamination is denoted by $\lambda_k^{\ast}(t)$, which is a periodic function with period length equal to $T_0$. 
At this stage, we do not impose any other structural assumptions on $\lambda_k^{\ast}(t)$. 

In our algorithm, we use the working model $\lambda_k(t)$ to approximate $\lambda_k^{\ast}(t)$ and
$\lambda_k(t)$ has the same period $T_0$ as $\lambda_k^{\ast}(t)$ does.
For reader convenience, we provide several possible choices for  $\lambda_k(t)$ as follows.
\begin{itemize}
    \item \textit{Non-homogeneous Poisson Process}.~ 
    \begin{eqnarray}\label{eq:model1}
    \lambda_k(t) :=\sum_{h=1}^H b_{k,h} \kappa_h(t) ~~~~ \text{for}~ t \in [0, T_0], 
    \end{eqnarray}
    where $\kappa_h(t)$ is the $h$-th basis function.
    \item \textit{Self-exciting Process}.~ 
    \begin{eqnarray}\label{eq:model2}
    \lambda_k(t) :=\sum_{h=1}^H b_{k,h} \kappa_h(t)
    + \sum_{t_j < t} \sum_{h'=1}^{H'} \alpha_{k,h'} g_{h'}(t - t_j) ~~~~ \text{for}~ t \in [0, T_0],
    \end{eqnarray}
    where both $\kappa_h(t)$ and $g_{h'}(t)$ are basis functions.
\end{itemize}
 
 For notational simplicity, we denote $\boldsymbol{B}_k$ as the working parameter set for class $k$.
 In other words, $\boldsymbol{B}_k := \left[b_{k,h}\right] \in \mathbb{R}_{0+}^{H}$ for the non-homogeneous Poisson process and 
 $\boldsymbol{B}_k := \left[b_{k,h}, a_{k,h'}\right] \in \mathbb{R}_{0+}^{H + H'}$ for the self-exciting process. We further write $\boldsymbol{B} := \{\boldsymbol{B}_k\}_{k=1}^K$ as the complete parameter set.
 For later analyses, we abuse $H$ to denote the dimension of $\boldsymbol{B}_k$ and the results could apply to both model \eqref{eq:model1} and  \eqref{eq:model2}.

\begin{remark}
    In practice, we can choose $\{\kappa_h(t)\}$ and $\{g_{h'}(t)\}$ to be cubic spline functions or Gaussian kernel functions.
\end{remark}

\begin{remark}
    The working model $\lambda_k$ here can be taken as any temporal point process in the literature. Our framework does not need to specify the underlying $\lambda_k^{\ast}(t)$ correctly. 
\end{remark}


Apart from the intensity specification, the model also allows the event contamination.
The contamination mechanism is described as follows. 
\begin{itemize}
    \item[a] The event contamination for distinct event sequences are independent of each other. 
    \item[b] For each event sequence, we assume there is a set of non-overlapping sub-time intervals that could possibly contain outlier events. The total length of these sub-time intervals is at most $\eta \cdot T$, where $\eta \in (0,1)$.
    The locations of sub-time intervals are randomly generated from $[0,T]$.
    \item[c] During those contaminated time intervals, two types of contamination may happen: (Type-i) The original events may be eliminated / missing. (Type-ii) There are multiple new events inserted between two consecutive original events.
\end{itemize}

\begin{figure}[ht!]
	\centering
	\includegraphics[width=0.96\textwidth]{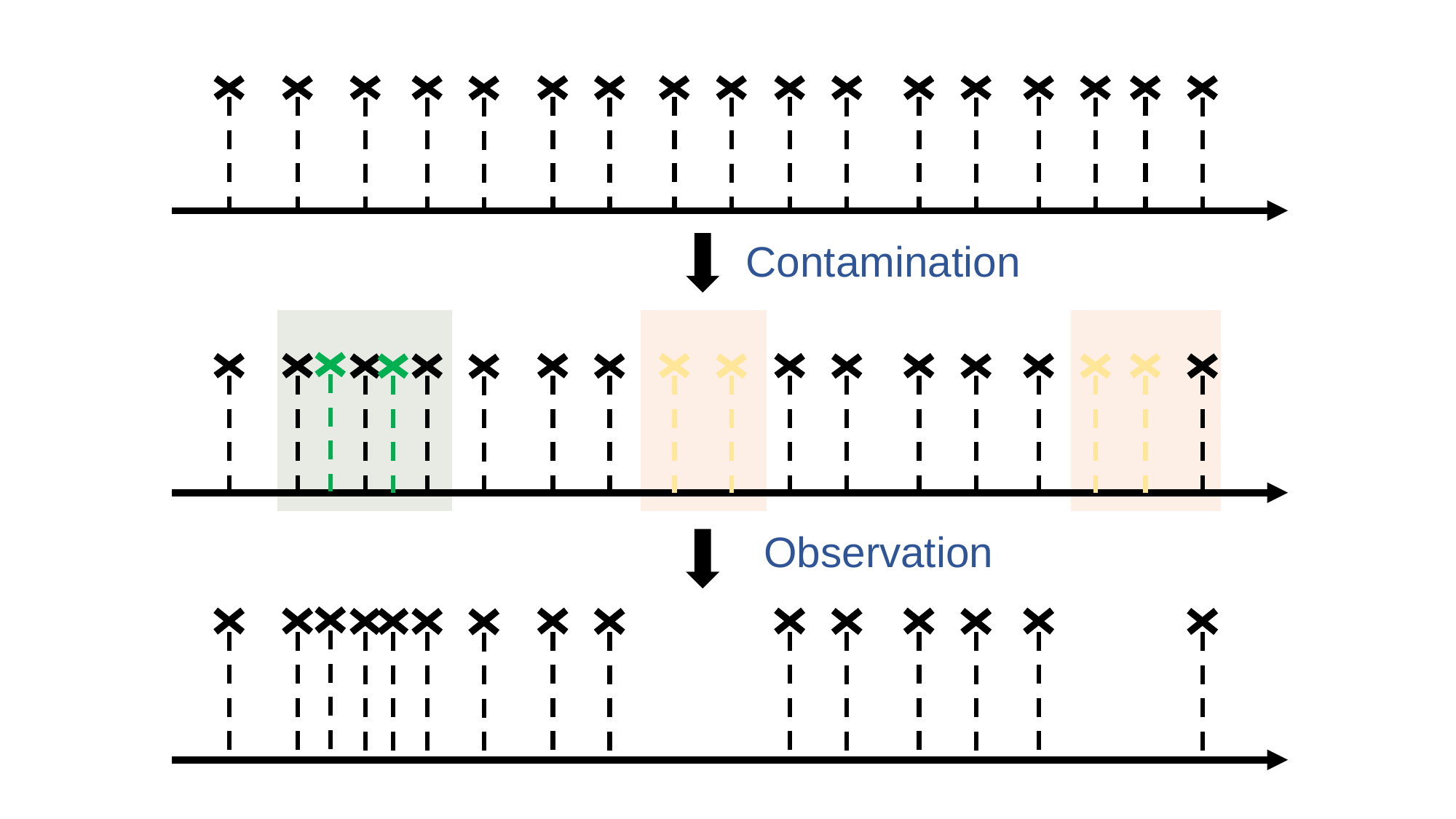}

    \caption{The visualization of event contamination process. The upper plot: the original event sequences generated from the true underlying intensity function. The middle plot: the event contamination on the event sequences. (Green area indicates the Type-ii time interval and Orange areas indicate Type-i time intervals.) The bottom plot: the observed event sequence after event contamination.}\label{fig:contam}
\end{figure}

According to the descriptions of the mechanism, different event sequences may have distinct lengths or numbers of contaminated sub-time intervals.
Within each sub-time interval, the number of events could either increase or decrease.
Type-i contamination can be viewed as the omission adversary. Type-ii contamination can be treated as the commission adversary.
To help readers gain more intuition, the above contamination procedure is visualized in Figure  \ref{fig:contam}.

\subsection{Weight Function}

According to the classical mixture models \citep{xu2017dirichlet, zhang2022learning} with no contamination, we can easily give out the probability of observing an event stream $S$ as
\begin{eqnarray}
   p(S; \boldsymbol{B}) &=& \sum_{k=1}^K \pi_k \cdot \operatorname{TPP}\left(S |  \boldsymbol{B}_k\right), \label{eq:prob0} \\ ~\text{with}~ \operatorname{TPP}\left(S | \boldsymbol{B}_k\right) &:=& \left( \prod_{i=1}^{M} \lambda_k(t_i) \right) \cdot \exp (- \int_{0}^{T_0} \lambda_k(u) d u) \cdot , \label{eq:obj:0}
\end{eqnarray}
where $\pi_k$ 's are class probabilities, $\operatorname{TPP}\left(S | \boldsymbol{B}_k\right)$ is the conditional probability of the event sequence $S$ if it belongs to class $k$ with working parameter $\boldsymbol{B}_k$ being plugged in, and $M$ is the number of events in the sequence $S$. 

Unfortunately, due to the presence of outliers, 
directly optimizing over \eqref{eq:obj:0} may lead to biased estimation and unreliable classification results.
Instead, we consider the following weighted version, that is,
\begin{eqnarray}\label{eq:obj:1}
   p_W(S; \boldsymbol{B}) 
   &=& \sum_{k=1}^K \pi_k \cdot \operatorname{WTPP}\left(S |  \boldsymbol{B}_k\right), \label{eq:prob} \\ ~\text{with}~ \operatorname{WTPP}\left(S | \boldsymbol{B}_k\right) 
   &:=& 
   \left(\prod_{i=1}^{M} \lambda_k(t_i)^{W_{i+1}(S;\boldsymbol{B})} \right) \cdot \left( \prod_{i=1}^{M + 1} [\exp (- \int_{t_{i-1}}^{t_i} \lambda_k(u) d u)]^{W_i(S;\boldsymbol{B})} \right)
   \nonumber \\
   &=& \prod_{i=1}^{M+1} \left(\lambda_k(t_{i-1}) \exp (- \int_{t_{i-1}}^{t_i} \lambda_k(u) d u)\right)^{W_i(S;\boldsymbol{B})}, \label{eq:quasi-density}
\end{eqnarray}
where $t_0 = 0, t_{M + 1} = T_0$, and $\lambda_k(t_0) \equiv 1$.
The most important modification in \eqref{eq:quasi-density} compared with \eqref{eq:obj:0} is that we introduce the weight $W_i(S;\boldsymbol{B})$ to time interval $(t_{i-1}, t_i]$ to adaptively adjust the importance of different events. 
The main purpose of the weights is presented as follows.
On the one hand, if $t_{i-1}$ and $ t_i$ are two original events, then the weight $W_i(S;\boldsymbol{B})$ should more likely to be close to 1.
On the other hand, if this time interval is contaminated, then the weight $W_i(S;\boldsymbol{B})$ tends to be close to 0. In order to achieve such purposes, the weight function is designed through the following procedure.

Given a set of working parameters $\{\boldsymbol{B}_k\}$'s, we first compute the following class-specific weight,
\begin{eqnarray}\label{eq:weight}
w_i(S;\boldsymbol B_k):=
	\begin{cases}
	\phi'_{\rho_1,\rho_2}\left( \int_{t_{i-1}}^{t_i} \lambda_k(u) d u-1\right); &~ i \leq M\\
	\phi'_{\rho_1,\rho_2}\left( \int_{t_{M}}^{T} \lambda_k(u) d u-1\right); &~ i = M + 1,\\
	\end{cases}    
\end{eqnarray}
where 
\begin{eqnarray}\label{eq:robust_fun}
\phi'_{\rho_1,\rho_2}(x):=
	\begin{cases}
	\phi'(x/\rho_2) &~ 0\leq x\\

        \phi'(x'/\rho_1) &~ -1\leq x <0, \\
	\end{cases}    
\end{eqnarray}
and
$\phi'(x)$ is the derivative function of an influence function $\phi(x)$.
Throughout the paper, we assume it has 
the following form for $x>0$,
\begin{eqnarray}\label{eq:robust_fun0}
\phi'(x):=
	\begin{cases}
	\frac{1+x}{1+x+x^2/2} &~ 0\leq x \leq a\\
	\phi'(a)\cdot(b-x)^2/(b-a)^2 &~ a<x\leq b\\
        0  &~ x > b,\\
	\end{cases}    
\end{eqnarray}
and $\phi'(x)=\phi'(x')$ for $1\leq x<0$, where $x':=\{x'|(x'+1)\exp(-x'-1)=(x+1)\exp(-x-1),x'\geq 0\}$. 
$\rho_1$ and $\rho_2$ are two positive tuning parameters.

We can easily see that $\phi'(x)$ achieves the maximum value at $x = 0$ and has the compact support. The influence function $\phi(x)$ is increasing and twice continuously differentiable. 
Thanks to these properties, $w_i(S;\boldsymbol B_k)$ will be closer to 1 if $\int_{t_{i-1}}^{t_i} \lambda_k(u) du$ is close to 1, while $w_i(S;\boldsymbol B_k)$ will be reduced to zero if $\int_{t_{i-1}}^{t_i} \lambda_k(u) du$ is far away from 1.
Moreover, by the time transformation theory, $\int_{t_{i-1}}^{t_i} \lambda_k^{\ast}(u) du \sim \text{Exp}(1)$ if event sequence $S$ is generated from the point process with intensity $\lambda_k^{\ast}(t)$. In other words, the class-specific weight $w_i(S;\boldsymbol B_k)$ is around one when time interval $(t_{i-1}, t_i]$ is not contaminated and sequence $S$ belongs to class $k$.

\begin{remark}
    It can be checked that $\phi(x)$ is also a Catoni-style influence function, which enjoys many statistical merits. See \cite{catoni2012challenging}, \cite{fang2023empirical}, and the references therein.
\end{remark}

With the constructions of $w_i(S_n;\boldsymbol B_k)$'s, we then choose the overall weight
\begin{eqnarray}\label{eq:weightall:1}
W_i(S;\boldsymbol{B}) := \max_{k \in [K]} w_i(S_n; \boldsymbol{B}_k)
\end{eqnarray}
or 
\begin{eqnarray}\label{eq:weightall:2}
W_i(S;\boldsymbol{B}) := \sum_{k \in [K]} r_{k} \cdot w_i(S; \boldsymbol{B}_k),
\end{eqnarray}
where $r_k$ is the probability of event sequence $S$ belonging to class $k$. Later, see \eqref{eq:weightall} for a concrete example of $r_k$'s. 
\eqref{eq:weightall:1} indicates that, for the $i$-th time interval of event sequence $S$, the weight is determined by the maximum weight across all $K$ classes. 
\eqref{eq:weightall:2} implies that, the weight of $i$-th time interval is determined by the combination of $K$ class-specific weights.
Both choices can ensure that the time interval $(t_{i-1},t_i]$ can have larger weight if at least one of $\int_{t_{i-1}}^{t_i} \lambda_k(u) du$ is close to 1. 



For the influence function defined in \eqref{eq:robust_fun0}, we can further prove that it possesses the following ``unbiasedness" properties. 

\begin{lemma}\label{lem:robust_exp}
When $X$ follows the standard exponential distribution $\text{Exp}(1)$, it holds that $ \mathbb{E}  \left[(X-1)\cdot\phi'(X-1) \right]=0 $ .
\end{lemma}

Note that $\int_{t_{i-1}}^{t_i} \lambda_k(u) du$ is left-skewed, i.e. has the longer right tail. 
By  Lemma \ref{lem:robust_exp}, our construction of $\phi(x)$ guarantees that the proposed weight function can carefully balance the left and right tails of the integral of the intensity function so that they will have the equal impacts on the log-likelihood function. Lemma \ref{lem:robust_exp} also directly leads to the following corollary.

\begin{corollary}
When $\rho_1=\rho_2$, it also holds that $\mathbb{E}_X  \left[(X-1)\cdot\phi'_{\rho_1,\rho_2}(X-1) \right]=0 $.
\end{corollary}

\subsection{Computation}

Denote the full latent label vector $\mathbf Z = \{Z_n\}_{n=1}^N$. 
The objective of the complete data is given as follows,
\begin{eqnarray}
    \text{WL}(\mathbf S, \mathbf Z; \boldsymbol{B}, \boldsymbol{\pi})
    = \sum_{n=1}^N \sum_{k=1}^K \mathbf 1\{Z_{n} = k\} \cdot \log \pi_k \cdot \log \text{WTPP}(S_n | \boldsymbol{B}_k). 
\end{eqnarray}
Based on the objective, we alternatively update the pseudo posterior \footnote{} of $\mathbf Z$ and the parameters $\boldsymbol{B}$ and $\boldsymbol{\pi}$ as follows.


\textbf{Update of $\mathbf Z$}.~ 
At time step $t$,
we compute the pseudo posterior $q(\mathbf Z | \mathbf S; \boldsymbol B^{(t-1)})$, where $\boldsymbol B^{(t-1)}$ is the parameter estimate in the previous step. It is not hard to find that 
\begin{eqnarray}\label{eq:posterior:e-step}
    q(\mathbf Z | \mathbf S; \boldsymbol B^{(t-1)}) 
    & = & \prod_{n=1}^N \prod_{k=1}^K (r_{nk}^{(t)})^{\mathbf 1\{Z_n = k\}}
\end{eqnarray}
with 
\begin{eqnarray}\label{eq:def:rnk}
    r_{nk}^{(t)} = \frac{\rho_{nk}^{(t)}}{\sum_{k'} \rho_{nk'}^{(t)}},
\end{eqnarray}
where $\rho_{nk}^{(t)} := \pi_{k}^{(t-1)} \cdot \operatorname{WTPP}(S_n|\boldsymbol B_k^{(t-1)})$. For simplicity, we may write $q(\mathbf Z | \mathbf S; \boldsymbol B^{(t-1)})$ as $q^{(t)}(\mathbf Z)$ in the following sections.

\textbf{Update of $\boldsymbol{B}$ and $\boldsymbol{\pi}$}.~
Given $q^{(t)}(\mathbf Z)$, we update the class probabilities by
\begin{eqnarray}\label{eq:update:pi}
\pi_k^{(t)} = \frac{1}{N}\sum_{n=1}^N r_{n k}^{(t)}.
\end{eqnarray}
Additionally, $\boldsymbol B^{(t)}$ is obtained by performing the gradient descent-type algorithm,
\begin{eqnarray}\label{eq:mstep}
\boldsymbol B_k^{(t)} := \boldsymbol B_k^{(t-1)} -lr\cdot \varrho_k^{(t)}, 
\end{eqnarray}
with 
\begin{eqnarray}
    \varrho_k^{(t)} := \sum_{n=1}^N r_{nk}^{(t)} \cdot \nabla_{B_k} \left\{\sum_{i=1}^{M_n} w_i(S_n;\boldsymbol{B}_k^{(t-1)}) \left(\log \lambda_k(t_{n,i-1}) - \int_{t_{n,i-1}}^{t_{n,i}} \lambda_k(u) du \right) \right\},
\end{eqnarray}
where $lr$ is the learning rate.
Then we denote $\boldsymbol B^{(t)} := (\boldsymbol B_k^{(t)})_{k=1}^K$.

\textbf{Tuning of $\rho_1$ and $\rho_2$}.~
Note that $\rho_1$ and $\rho_2$ cannot be too small.
Otherwise, it give vanishing weights $w_{i}(S_n; \boldsymbol{B}_k^{(t-1)})$'s.
In other words, $\operatorname{WTPP}(S_n|\boldsymbol B_k^{(t-1)})$'s are close to zero and do not provide sufficient information for the data sequence. As a result, $q^{(t)}(\mathbf Z)$ will be inaccurate and $\boldsymbol{B}_k^{(t)}$ may not converge to the optimal value. 

\begin{figure}[ht!]
	\centering
    \includegraphics[width=0.9\textwidth]{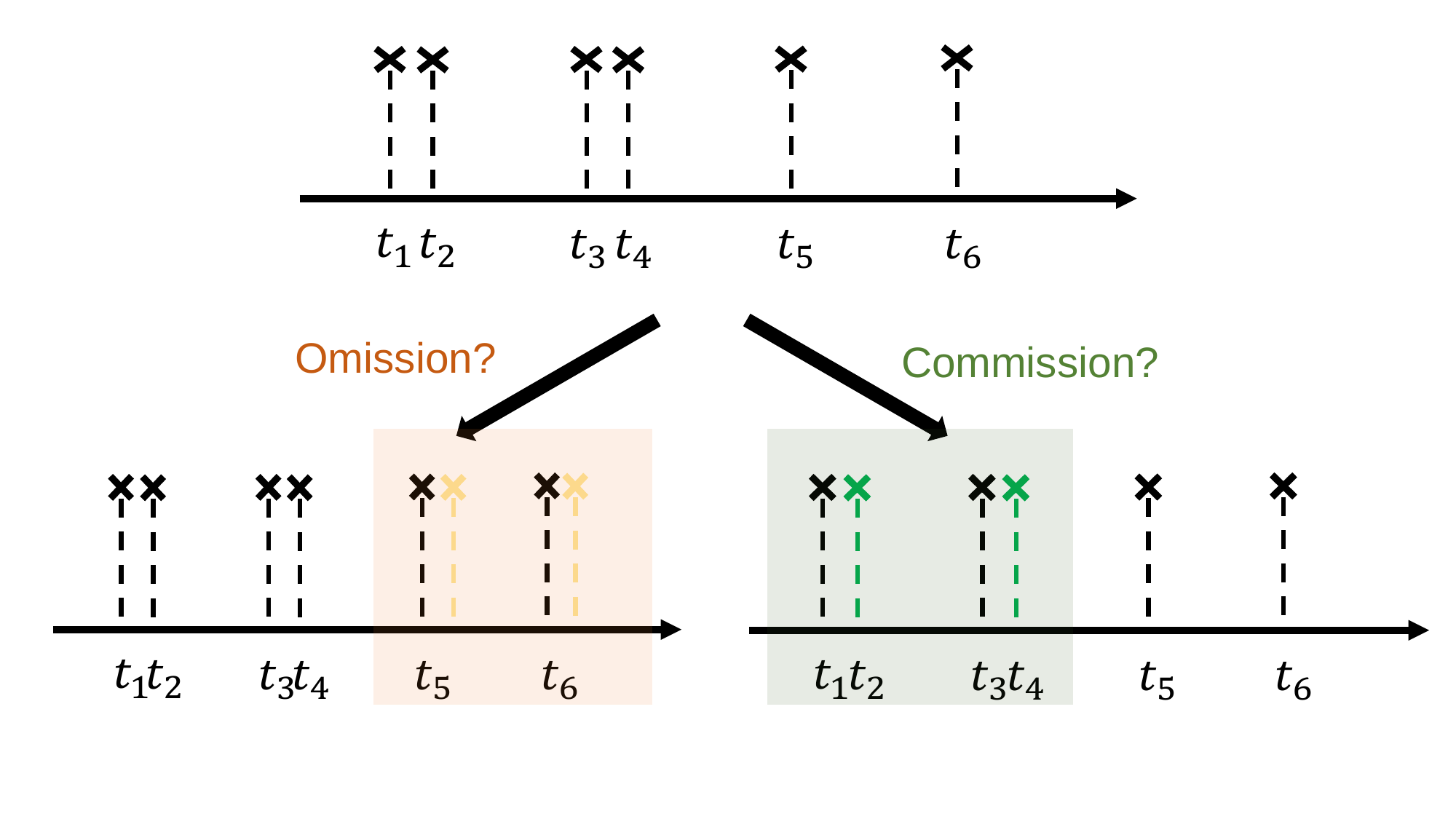}
    \caption{A toy example of the non-identifiability case. The upper plot gives an observed event sequence, which is equally likely to be generated from the event sequence with type-i contamination (bottom left) or type-ii contamination (bottom right). In other words, we may have at least 50 \% chance to give the wrong classification result when the restriction \eqref{eq:adjust:rho} is removed. }\label{fig:indenti}
\end{figure}

In the update, we adjust $\rho_1$ and $\rho_2$ so that 
\begin{eqnarray}\label{eq:adjust:rho}
\frac{1}{n T_0} \sum_{n=1}^N \sum_{i=1}^{M_n} w_i(S_n;\boldsymbol{B}_k^{(t)})(t_{n,i} - t_{n,i-1}) \geq \frac{1}{2}.
\end{eqnarray}
Heuristically speaking, \eqref{eq:adjust:rho} ensures that at least half of event information is reserved. 
This requirement is not too stringent. By the assumption of the contamination mechanism, there are at most $100 \cdot \eta$ percentage of time windows containing outlier events. Therefore, $1/2$ is a very loose lower bound for $1 - \eta$.
On the other hand, if we do not put the constraint \eqref{eq:adjust:rho}, it could lead to a non-identifiability issue.
That is, two types of event contamination could lead to the same observation with high probability. 
A toy example of non-identifiability is given in Figure \ref{fig:indenti}.

The complete computation procedure is summarized in Algorithm \ref{alg1}.

\begin{algorithm}
	\renewcommand{\algorithmicrequire}{\textbf{Input: Sequences $S = \{S_n \}_{n=1}^N$}}
	\renewcommand{\algorithmicensure}{\textbf{Output:}}
	\caption{Clustering with Event Contamination}
	\label{alg1}
	\begin{algorithmic}[1]
		\STATE \textbf{Input} {Sequences $\mathbf S = \{S_n \}_{n=1}^N$}, tolerance parameter $\epsilon$.
            \STATE Initialization of $\boldsymbol{B}^{(0)}$, $\boldsymbol{\pi}^{(0)}$, and $t = 0$.
		\REPEAT
            \STATE Increase time index $t = t + 1$.
		\STATE \textbf{Update of } $\mathbf Z$:
            \STATE \quad Update $q^{(t)}(\mathbf Z)$ by \eqref{eq:posterior:e-step}.
             \STATE \textbf{Update of } $\boldsymbol{B}$ and $\boldsymbol{\pi}$: 
             \STATE \quad $\forall k\in [K]$, update the parameter $\pi_k^{(t)}$ by 
            \eqref{eq:update:pi}.
		\STATE \quad $\forall k\in [K]$, update the parameter $\boldsymbol{B}_k^{(t)}$ by 
            \eqref{eq:mstep}.
            \STATE \textbf{Adjustment of} $\rho_1, \rho_2$: 
            \STATE \quad Adjust $\rho_1$ and $\rho_2$ so that \eqref{eq:adjust:rho} holds.
		\UNTIL $\|\boldsymbol{B}_{k}^{(t)}-\boldsymbol{B}_{k}^{(t-1)}\|\leq\epsilon, ~\forall k\in [K]$.  
		\ENSURE  Estimated parameters $\hat{\boldsymbol{B}}, \hat{\boldsymbol{\pi}}$, Cluster responsibilities $\{\hat{r}_{nk}\}$.
	\end{algorithmic}  
\end{algorithm}

\begin{remark}\label{rmk:W:choose}
In practice, we suggest the following computations for $W_i(S_n; \boldsymbol{B}^{(t)})$.
In the first few iterations, we compute 
\begin{eqnarray}
W_i(S_n;\boldsymbol{B}^{(t)}) := \max_{k \in [K]} w_i(S_n; \boldsymbol{B}^{(t)}).
\end{eqnarray}
This strategy prioritizes the most significant class-specific weight, providing a robust initial classification by emphasizing the strongest class association and mitigating the impact of outliers.

In the rest of iterations, we compute 
\begin{eqnarray}\label{eq:weightall}
W_i(S_n;\boldsymbol{B}^{(t)}) := \sum_{k \in [K]} r_{nk}^{(t)} \cdot w_i(S_n; \boldsymbol{B}^{(t)}).
\end{eqnarray}
This strategy integrates the probabilistic contributions from all classes, reflecting a more balanced and comprehensive assessment of the class memberships.
As the algorithm progresses, the incorporation of class probabilities allows for a finer adjustment between classes, enhancing the overall robustness and accuracy of the clustering algorithm in practice.
\end{remark}

\begin{remark}\label{rmk:weight:role}
    We also like to point out that we use the weight functions differently in updating $q^{(t)}(\mathbf Z)$ and $\boldsymbol{B}^{(t)}$.
    In the former update, we use the overall weight function $W_i(S_n; \boldsymbol{B}^{(t)})$.
    This is because we need to aggregate all $K$ class information to distinguish an event being outlier or not. Then the posterior $q^{(t)}(\mathbf Z)$ can concentrate its mass on the most preferred class labels given the weighted time intervals.
    In contrast, we use the class-specific weight functions $w_i(S_n; \boldsymbol{B}_k^{(t)})$'s in the later update.
    This is because we want to screen out those events which are unlikely to be generated from Class $k$. Especially in the beginning of the algorithm, $r_{nk}^{(t)}$'s are not accurate, using the overall weight could overemphasize those contaminated time intervals.
\end{remark}

\section{Theoretical results}\label{sec:4}

In this section, we provide theoretical explanations of our proposed framework to help readers to gain more insights of how the algorithm works. To start with, we introduce several technical assumptions.

\begin{assumption}\label{asm:1}
    There is a lower bound $\pi_{low} > 0$ for the proportion of each class, that is, $\pi_k \geq \pi_{low}$ for all $k=1,2,\dots,K$.
\end{assumption}

Assumption \ref{asm:1} ensures that no class is drained. It is a common condition in the literature of mixture models. 

\begin{assumption}\label{asm:2}
    The space of working model parameters $\boldsymbol{B}_{k}$'s is bounded. That is, there exists $\Omega_B>0$ such that $\left\|\boldsymbol{B}_{k}\right\|_1 < \Omega_B$ for all $k=1,2,\dots,K$.
\end{assumption}

Assumption \ref{asm:2} is a standard technical condition commonly referenced in statistical theory literature (see \citep{lehmann2006theory, casella2021statistical}). Assuming the boundedness of the parameter space makes the technical argument easier.


\begin{assumption}\label{asm:3}
    There exist $\tau$ and $\Omega$ such that $0 < \tau \leq \lambda_k^{\ast}(t) \leq \Omega$ for all $t \in [0,T]$ and $k=1,2,\dots,K$.
\end{assumption}

Assumption \ref{asm:3} is also a usual technical requirement, as noted in recent works \citep{cai2022latent, fang2023group}. This assumption ensures that the intensity function remains bounded away from zero and is also constrained from above. It also helps to make the analyses easier.


We next define the true working model parameter,
\begin{eqnarray}\label{eq:true:work:para}
    \boldsymbol{B}_k^{\ast} = \arg\max_{ \boldsymbol{B}_k} \mathbb E \bigg[  \int_0^{T_0} (\log \lambda_k(t)) \cdot d N_k(t)  - \int_0^{T_0} \lambda_k(t) dt  \bigg] ~~ \text{for}~ k \in [K]
\end{eqnarray}
with $\lambda_k(t)$ being an intensity function parameterized by $\boldsymbol{B}_k$ and $N_k(t)$ is a TPP following the true underlying intensity $\lambda_k^{\ast}(t)$ with no event contamination. 
In the following, we write $\lambda_{\boldsymbol{B}_k^{\ast}}(t)$ as the intensity $\lambda_k(t)$ with $\boldsymbol{B}_k^{\ast}$ plugged in. 
Then $\lambda_{\boldsymbol{B}_k^{\ast}}(t)$ is the intensity function closest to $\lambda_k^{\ast}(t)$ within the working model space.

\begin{assumption}\label{asm:4}
    For any two different classes $k$ and $k'$ in $[K]$, there exists a constant $C_{gap} > 0$ such that, if event stream $S$ belongs to Class $k$, then it holds $\mathbb{E}[\log \operatorname{TPP}(S | \boldsymbol{B}_{k'}^*)]<\mathbb{E} [\log \operatorname{TPP} (S|\boldsymbol{B}_k^*)] - C_{gap} \cdot L$.
\end{assumption}

In Assumption \ref{asm:4}, the expectation is taken with respect to an event sequence $S$ which follows the intensity $\lambda_k^{\ast}(t)$. This assumption ensures the class identifiability. In other words, it guarantees that, before the event contamination, event streams from different classes can be distinguished by our working model. 

\begin{assumption}\label{asm:5}
    The event contamination follows the mechanism described in Section \ref{sec:mod}.
\end{assumption}

According to Assumption \ref{asm:5}, we know that the proportion of time intervals which could be potentially contaminated does not exceed $\eta$. This assumption is crucial for maintaining the identifiability of the true model parameters and latent classes, as it effectively limits the influence of event outliers on the analysis when $\eta$ is not too large. 
Furthermore, it assumes that the intervals in which outliers may appear are independently and randomly distributed for each event sequence. 
This randomness in the occurrence of outliers helps to prevent systematic biases that could otherwise skew the results. 


\begin{theorem}\label{Thm:Mgrad}
When $\rho_1=\rho_2$, it holds that 
\begin{eqnarray}\label{eq:unbias}
     \left\| \mathbb E \left[ \frac{1}{L} \sum_{i=1}^{M} w_i(S;\boldsymbol{B}_k) \nabla_{B_k} \left(\log \lambda_k(t_{i-1}) - \int_{t_{i-1}}^{t_{i}} \lambda_k(u) du \right) \right] \Bigg|_{\boldsymbol{B}_k = \boldsymbol{B}_k^{\ast}} \right\| = O\left(\frac{\sqrt{H}}{L}\right),
\end{eqnarray}
where $S$ follows the underlying intensity $\lambda_k^{\ast}(t)$.
As a result, when there are no outlier events, $\mathbb E[\varrho(B_k^{\ast})] \rightarrow 0$ as $L \rightarrow \infty$,
where 
\begin{eqnarray}\label{def:grad:all}
\varrho(\boldsymbol{B}_k) := \frac{1}{N L} \sum_{n: Z_n = k} \sum_{i=1}^{M_n} \left\{ w_i(S_n;\boldsymbol{B}_k) \nabla_{B_k} \left(\log \lambda_k(t_{n,i-1}) - \int_{t_{n,i-1}}^{t_{n,i}} \lambda_k(u) du \right) \right\}.
\end{eqnarray}

\end{theorem}

Theorem \ref{Thm:Mgrad} shows that, even after incorporating the weight function, the expected gradient evaluated at the true working model parameters remains asymptotically unbiased when there is no event contamination and $L \rightarrow \infty$. 
In other words, it implies that the true parameter $\boldsymbol{B}^*$ continues to be the (at least local) optimum given sufficiently many observed events. 
This result indicates that our proposed methodology is adaptive and does not deteriorate the performance under the classical setting (i.e., with the absence of event contamination).  

\begin{remark}
    For the special case that both true model and working model are homogeneous Poisson process, then the left hand side of \eqref{eq:unbias} is exactly zero.
\end{remark}

\begin{remark}
    Curious readers may wonder whether we can replace 
    $\sum_i w_i(S;\boldsymbol{B}_k) \nabla_{B_k} (\log \lambda_k(t_{i-1}) - \int_{t_{i-1}}^{t_{i}} \lambda_k(u) du )$ by
    $\sum_i w_i(S;\boldsymbol{B}_k) \nabla_{B_k} (\log \lambda_k({\color{orange}t_{i}}) - \int_{t_{i-1}}^{t_{i}} \lambda_k(u) du )$ in the computation of gradient.
    Empirically, we find they have negligible difference. 
    However, Theorem \ref{Thm:Mgrad} may no longer hold.
\end{remark}


\begin{theorem}\label{Thm:Mout}
Under Assumptions \ref{asm:3} and \ref{asm:5}, with $\rho_1 = \rho_2 = \rho$, for each $k \in [K]$, it holds that 
\begin{eqnarray}
   \| \varrho(\boldsymbol{B}_k^{\ast}) \| 
    = O_p\left(\sqrt{H}\cdot\bigg(\underbrace{\eta}_{\text{contamination}} + \underbrace{\frac{1}{\sqrt{NL}}}_{\text{stochastic}}+\underbrace{\frac{1}{L}}_{\text{using weight}}\bigg)\right),
\end{eqnarray}
where $H$ is the dimension of the parameter $\boldsymbol{B}_k$ .
\end{theorem}

Theorem \ref{Thm:Mout} demonstrates that after incorporating the weight function, the gradient evaluated at the optimal value $\boldsymbol{B}^*$ has an upper bound which consists of three parts, the contamination error, the stochastic error, and the bias induced by weight function. Moreover, the bound is regardless of the number of outlier events added or removed within the contaminated time intervals, suggesting the robustness of the proposed approach.
On the other hand, without the help of weight functions, the impact of outliers on the gradient may not be controlled.


Instead of allowing the arbitrary Type-i and Type-ii contamination, we further assume the specific generation mechanism of contamination adversary. 
\begin{assumption}\label{asm:8}
In the step (c) of the contamination mechanism, we specifically assume 
\begin{itemize}
    \item[] (Type-i) the original events are removed with intensity rate $\lambda_{miss}(t)$;
    \item[] (Type-ii) the new events are added according to the intensity $\lambda_{add}(t)$.
\end{itemize}
\end{assumption}
Assumption \ref{asm:8} says that, when the omission occurs, the event from class $k$ is observed at time $t$ with intensity $\lambda_k^{\ast}(t) - \lambda_{miss}(t)$. When the commission occurs, the intensity at time $t$ is increased to $\lambda_k^{\ast}(t) + \lambda_{add}(t)$.

\begin{assumption}\label{asm:6}
Only one of Type-i or Type-ii contamination is allowed to happen.
\end{assumption}

\begin{theorem}\label{cor:grad}
Under Assumptions \ref{asm:3}, \ref{asm:5}, \ref{asm:8}, and \ref{asm:6} and additionally assume that $\lambda_{miss}(t)$ and $\lambda_{add}(t)$ is proportional to $\lambda_k^{\ast}(t)$, it holds that 
\begin{align*}
    | \mathbb E[\varrho(\boldsymbol{B}_k^{\ast})]| < 
|\mathbb E[\overline \varrho(\boldsymbol{B}_k^{\ast})]|, 
\end{align*}
for sufficiently large $L$, with 
\begin{eqnarray}\label{def:grad:noweight}
\overline \varrho(\boldsymbol{B}_k) := \frac{1}{N L} \sum_{n: Z_n = k} \sum_{i=1}^{M_n}  \left\{ \nabla_{B_k} \left(\log \lambda_k(t_{n,i-1}) - \int_{t_{n,i-1}}^{t_{i}} \lambda_k(u) du \right) \right\},
\end{eqnarray}
where ``$<$" represents the element-wise comparison.
\end{theorem}


Theorem \ref{cor:grad} implies that, after incorporating the weight function, it is expected that the absolute value of the gradient evaluated at the true value $\boldsymbol{B}^{\ast}$ is strictly smaller than that without using the weight functions. 
This reduction in the gradient value means that 
the weight functions can help the gradient move towards the less biased direction when either omission or commission contamination happens.
Theorem \ref{Thm:Mgrad} - Theorem \ref{cor:grad} together characterize the roles of the weight functions in the gradient computation.


Next, we provide the non-asymptotic convergence analysis of the proposed algorithm when the event contamination is allowed. 
{ In addition, we define the class-specific gradient,
\begin{align*}
    g(\boldsymbol{B}_k\mid \boldsymbol{B}^{\ast}):= \mathbb{E}_{S\sim \lambda_k^*} [r_k(S ; \boldsymbol{B}^{\ast})\cdot \nabla\log \operatorname{WTPP}(S\mid \boldsymbol{B}_k) /L ],
\end{align*}
with the weight $W_i(S; \boldsymbol{B}) = \sum_{k}r_k(S ; \boldsymbol{B}^{\ast}) w(S;\boldsymbol B_k)$.
Here 
{ $r_k(S;\boldsymbol{B})={\pi_k \operatorname{WTPP}(S\mid \boldsymbol{B}_k,W)}/{\sum_k\pi_k \operatorname{WTPP}(S\mid \boldsymbol{B}_k, W)}$}.


{
\begin{lemma}\label{lem:lambda}
    Let $\lambda_{k,\max}$ and $\lambda_{k,min}$ be the largest and smallest eigenvalue of $-\nabla_{\boldsymbol{B}_k} g(\boldsymbol{B}_k\mid \boldsymbol{B}_k^*)$.
    Under Assumptions \ref{asm:1}-\ref{asm:3}, it holds that $\lambda_{k,min} > 0$ for all $k \in [K]$.
\end{lemma}
}

\begin{theorem}\label{thm:local}
Suppose Assumptions \ref{asm:1}- \ref{asm:5} hold and the maximum possible contamination proportion is $\eta_{max} \in (0, 1/2)$. 
We choose $\rho_1 = \rho_2 = \rho > 0$ such that 
\begin{eqnarray}\label{eq:rho:require}
\mathbb{E}[(1-X)\phi_{\rho_1,\rho_2}'(X-1)]>0.5/(1-\eta_{max}), ~ X\sim\text{Exp}(1).
\end{eqnarray}
We further assume  the underlying contamination proportion $\eta$ satisfies that 
\begin{eqnarray}\label{eq:eta:require}
(1-\eta)C_{gap} c_{1,\rho} -\eta\int_0^T \lambda_k^*(u) d u+\eta \min\{0, c_{2,\rho} \int_0^T \log(\lambda_k^*(u)/\lambda_j^*(u)) d u\} >0,
\end{eqnarray}
$\forall j \neq k$,
where $c_{1,\rho}$ and $c_{2,\rho}$ are two constants that may depend on $\rho$.
There exists a constant $a>0$ such that 
whenever $\|\boldsymbol{B}_k^{(t)}-\boldsymbol{B}_k^*\|<a$ for $k \in [K]$, it holds
\begin{eqnarray}\label{eq:shrink}
    \left\|\boldsymbol{B}_k^{(t+1)}-\boldsymbol{B}_k^*\right\| \leq \underbrace{\left(\frac{\lambda_{k,max}-\lambda_{k,min}}{\lambda_{k,max}+\lambda_{k,min}}+\gamma\right)}_{\text{shrink rate}}\left\|\boldsymbol{B}_k^{(t)}-\boldsymbol{B}_k^*\right\|+\epsilon^{unif},
\end{eqnarray}
where $\lambda_{\max}$ and $\lambda_{min}$ are the same as in Lemma \ref{lem:lambda}, 
$\gamma$ is a constant satisfying $\gamma \rightarrow 0$ when $L \rightarrow \infty$, and the error $\epsilon^{unif} = O_p(\sqrt{H}\cdot(1/\sqrt{NL}+ \eta + {1/L}))$.
\end{theorem}


\begin{remark}
    In Theorem \ref{thm:local}, \eqref{eq:rho:require} ensures that \eqref{eq:adjust:rho} holds with high probability whenever $\rho^{(t)} \geq \rho$; \eqref{eq:eta:require} guarantees that distinct class can be uniquely identified even with $100 \cdot \eta$ percent of contaminated time windows.
    Smaller $\eta$ makes \eqref{eq:eta:require} easier to hold. Theorem \ref{thm:local} holds under arbitrary contamination, i.e., it does not rely on Assumptions \ref{asm:8} and \ref{asm:6}.
\end{remark}

For large enough $L$, the ``shrink rate" is smaller than 1.
Therefore, Theorem \ref{thm:local} implies that the gap $\|\boldsymbol{B}_k^{(t)}-\boldsymbol{B}_k^*\|$ decreases geometrically until it reaches the same order as $\epsilon^{unif}$. In other words, this result indicates a rapid convergence where the distance between $\boldsymbol{B}_k^{(t)}$ and $\boldsymbol{B}_k^*$ shrinks exponentially towards $\epsilon^{unif}$. Consequently, our robust clustering algorithm enjoys a local linear convergence speed.
As a side product, we also have the following local consistency of class label estimation.


\begin{corollary}\label{col:rnew}
    Under the same conditions of Theorem \ref{thm:local} and  $\|\boldsymbol{B}_k^{(0)} - \boldsymbol{B}_k^{\ast}\| < a$ for all $k$, it holds that $\left\|r_k(S_n ; \hat{\boldsymbol{B}})- r_k^*(S_n)\right\|= O_p(\exp(- G \cdot L)) $ for $n \in [N]$, where $r_k^*(S_n) =  \mathbf{I}_{\{ Z_n^{\ast} = k\}}$ and $G$ is some universal constant.
\end{corollary}

Moreover, we show that our algorithm can detect the contaminated time intervals when the contamination level is high. The details are presented in the supplementary file.

\section{Simulation Study}\label{sec:5}

To demonstrate the feasibility and the efficiency of our new method, we compare it with the baseline method, i.e.,  the same procedure without weight functions, under two types of working models, non-homogeneous Poisson process and self-exciting (Hawkes) process. 

\subsection{Non-homogeneous Poisson processes}


The simulation settings are described as follows. 
The event sequences are generated from a mixture of non-homogeneous Poisson processes with four classes, whose intensity functions are given by
\begin{align*}
    \lambda_1^{\ast}(t)&=3\exp(-{t^2}/{20})+2\exp(-{(t-8)^2}/{20})+\exp(-{(t-20)^2}/{20})+3\exp(-{(t-25)^2}/{3}),\\
    \lambda_2^{\ast}(t)&=2\exp(-{(t-6)^2}/{10})+5\exp(-{(t-20)^2}/{10})+\exp(-{t^2}),\\
    \lambda_3^{\ast}(t)&=5\exp(-{(t-5)^2}/{3})+3\exp(-{(t-12)^2}/{2})+5\exp(-{(t-18)^2}/{3}),\\
    \lambda_4^{\ast}(t)&=5\exp(-{(t-21)^2}/{20})+2\exp(-{(t-12)^2}/{10})+3\exp(-{t^2}/{2}),
\end{align*}
where $t\in [0,T]$ with $T = 24$ (corresponding to 24 hours). 


Type-i contamination is formed by randomly selecting sub-time windows whose total length is no more than $\eta \cdot T$ and deleting all the events within these time windows. 
Type-ii contamination is formed by randomly selecting sub-time windows whose total length is no more than { $\eta \cdot T_0$} and generating new events that follow the random intensity 
$\sum_c U\cdot\exp\{- (t - t_c)^2 /2\sigma^2 \} / \sqrt{2\pi}$ with 
$\sigma=0.05$, $U\sim U[2.5,5]$,
where the random centers $t_c$'s follow the homogeneous Poisson process ($\lambda = 5/12$).  


For each class, we generate 30 event sequences and there are $N =120$ samples in total.
The number of periods $L \in \{1, 2, 4\}$ and the contamination proportion $\eta \in \{0.15, 0.2, 0.25\}$.
We apply our proposed method and two baselines by setting the number of classes equal to 4, 5, or 6.
All cases are repeated for $50$ times.
In the experiments, we choose the number of basis $H$ to be 6 and the stopping criterion is $\|\boldsymbol{B}^{(t)} - \boldsymbol{B}^{(t-1)}\| \leq 0.1$. 

We use the clustering purity \citep{schutze2008introduction} to evaluate the performances of two methods.
To be specific, the purity index is defined as 
\begin{eqnarray}\label{eq:purity}
\operatorname{purity}(\hat{\mathcal S}, \mathcal S^{\ast})=\frac{1}{N} \sum_{k=1}^K \max_{k'\in[K^\ast]} \left|\hat{\mathcal S}_k \cap \mathcal S_{k'}^{\ast} \right|,
\end{eqnarray}
where $\hat{\mathcal S} = \{\hat{\mathcal S}_1, ..., \hat{\mathcal S}_{\hat K}\}$ and $\mathcal S^{\ast} = \{\mathcal S_1^{\ast}, ..., \mathcal S_{K^{\ast}}^{\ast}\}$ are two partitions of the data set according to the estimated labels and true underlying labels.
It is easy to see that the range of purity value is between $0$ and $1$. The higher the purity value is, the better the clustering result is. Moreover, the purity is non-decreasing as $\hat K$ increases. In other words, for a fixed algorithm, the purity will get larger if we wish to cluster the data into more classes. The results are summarized in Table \ref{tab:sim1}.



\begin{table}[]
    \centering
    {
    \begin{tabular}{c|c|c|ccc|ccc}
    \toprule  
    \multicolumn{1}{c}{\multirow{2}*{$\eta$}}&\multicolumn{1}{|c|}{\multirow{2}*{Time}}&\multicolumn{1}{c|}{\multirow{2}*{Algorithm}}& \multicolumn{3}{c|}{Type-i}& \multicolumn{3}{c}{Type-ii}\\
    \hiderowcolors
    \cline{4-9}
    \multicolumn{1}{c}{}&\multicolumn{1}{|c|}{}&\multicolumn{1}{c|}{} & K=4 & K=5 & K=6 & K=4 & K=5 & K=6 \\
    \midrule  
     \multicolumn{1}{c}{}{\multirow{6}*{0.15}}  &\multicolumn{1}{|c|}{\multirow{3}*{L=1}}& Standard & 0.7655 & 0.8233 & 0.8513 & 0.3715 & 0.3960 & 0.4480\\
     \multicolumn{1}{c}{} &\multicolumn{1}{|c|}{} &Robust & 0.9478 & 0.9627 & 0.9620 & 0.8593 & 0.9175 & 0.9420\\
    \cline{2-9}
     \multicolumn{1}{c}{} &\multicolumn{1}{|c|}{\multirow{3}*{L=2}}& Standard & 0.7208 & 0.8145 & 0.8517 & 0.3987 & 0.4193 & 0.4400\\
     \multicolumn{1}{c}{} &\multicolumn{1}{|c|}{} &Robust & 0.9440 & 0.9867 & 0.9878 & 0.8853 & 0.9555 & 0.9830\\
    \cline{2-9}
     \multicolumn{1}{c}{} &\multicolumn{1}{|c|}{\multirow{3}*{L=4}}& Standard & 0.7365 & 0.8283 & 0.8787 & 0.4487 & 0.4913 & 0.4968\\
     \multicolumn{1}{c}{} &\multicolumn{1}{|c|}{} &Robust & 0.9243 & 0.9742 & 0.9883 & 0.8903 & 0.9590 & 0.9728\\
    \midrule  
    \multicolumn{1}{c}{}{\multirow{6}*{0.2}}  &\multicolumn{1}{|c|}{\multirow{3}*{L=1}}& Standard & 0.6960 & 0.7497 & 0.7838 & 0.3553 & 0.3922 & 0.4312 \\
    \multicolumn{1}{c}{} &\multicolumn{1}{|c|}{} &Robust & 0.9307 & 0.9568 & 0.9550 & 0.8608 & 0.9110 & 0.9428\\
    \cline{2-9}
    \multicolumn{1}{c}{} &\multicolumn{1}{|c|}{\multirow{3}*{L=2}}& Standard & 0.6998 & 0.7580 & 0.7862 & 0.3815 & 0.4210 & 0.4470\\
    \multicolumn{1}{c}{} &\multicolumn{1}{|c|}{} &Robust & 0.9565 & 0.9830 & 0.9827 & 0.9047 & 0.9495 & 0.9700\\
    \cline{2-9}
    \multicolumn{1}{c}{} &\multicolumn{1}{|c|}{\multirow{3}*{L=4}}& Standard & 0.6847 & 0.7777 & 0.8157 & 0.4438 & 0.4902 & 0.5043\\
    \multicolumn{1}{c}{} &\multicolumn{1}{|c|}{} &Robust & 0.8983 & 0.9873 & 0.9872 & 0.9068 & 0.9672 & 0.9748\\
    \midrule  
     \multicolumn{1}{c}{}{\multirow{6}*{0.25}}  &\multicolumn{1}{|c|}{\multirow{3}*{L=1}}& Standard & 0.5828 & 0.6605 & 0.7413 & 0.3625 & 0.3882 & 0.4200\\
    \multicolumn{1}{c}{} &\multicolumn{1}{|c|}{} &Robust & 0.9353 & 0.9383 & 0.9422 & 0.8777 & 0.9122 & 0.9427\\
    \cline{2-9}
    \multicolumn{1}{c}{} &\multicolumn{1}{|c|}{\multirow{3}*{L=2}}& Standard & 0.6537 & 0.7348 & 0.7918 & 0.3850 & 0.4117 & 0.4387\\
    \multicolumn{1}{c}{} &\multicolumn{1}{|c|}{} &Robust & 0.9358 & 0.9783 & 0.9870 & 0.9162 & 0.9622 & 0.9625 \\
    \cline{2-9}
    \multicolumn{1}{c}{} &\multicolumn{1}{|c|}{\multirow{3}*{L=4}}& Standard & 0.7108 & 0.7727 & 0.8077 & 0.4467 & 0.4928 & 0.5208\\
    \multicolumn{1}{c}{} &\multicolumn{1}{|c|}{} &Robust & 0.9337 & 0.9837 & 0.9890 & 0.9142 & 0.9570 & 0.9838\\
    \bottomrule 
    \end{tabular}
    }
    \caption{Purity indices returned by two methods under the setting of $\eta=0.15$, $\eta = 0.2$ and $\eta=0.25$ for Non-homogeneous Poisson processes.}
    \label{tab:sim1}
\end{table}

\subsection{Self-exciting processes}

The event sequences are generated from a mixture of non-homogeneous Hawkes processes with four classes,
\[\lambda_{hak,k}(t) = \lambda_k^{\ast}(t) + \sum_{t_j < t} g_k^{\ast}(t - t_j).\]
The detailed formulas are given as follows.
\begin{align*}
    \lambda_1^{\ast}(t)&=3\exp(-{t^2}/{20})+2\exp(-{(t-8)^2}/{20})+\exp(-{(t-20)^2}/{20})+3\exp(-{(t-25)^2}/{3}),\\
    \lambda_2^{\ast}(t)&=2\exp(-{(t-6)^2}/{10})+5\exp(-{(t-20)^2}/{10})+\exp(-{t^2}),\\
    \lambda_3^{\ast}(t)&=5\exp(-{(t-5)^2}/{3})+3\exp(-{(t-12)^2}/{2})+5\exp(-{(t-18)^2}/{3}),\\
    \lambda_4^{\ast}(t)&=5\exp(-{(t-21)^2}/{20})+2\exp(-{(t-12)^2}/{10})+3\exp(-{t^2}/{2}),
\end{align*}
with
{\small
\begin{align*}
g_1^{\ast}(t)&=\frac{0.05}{\sqrt{\pi}}\exp(-{t^2}/{4}), \ g_2^{\ast}(t)=\frac{0.1}{3\sqrt{\pi}/2}\exp(-{t^2}/{9}),\\
g_3^{\ast}(t)&=\frac{0.15}{\sqrt{\pi}}\exp(-{t^2}/{4}), \ g_4^{\ast}(t)=\frac{0.15}{3\sqrt{\pi}/2}\exp(-{t^2}/{9}).
\end{align*}}

In the working model, we choose $\{g_{h'}(\cdot)\}_{h'\in [H']}$ to be the Gaussian kernel basis functions as , where $g_{h'}(t) = \exp(-(t-h'T/H')^2/2\sigma^2)/\sqrt{2\pi}\sigma$, where $\sigma=T/H' $ and $H'=6$ are selected according to \cite{xu2017dirichlet}.
The generation of outlier events and experimental details are the same as in the previous section.
Similarly, purity indices are reported in Table \ref{tab:sim2}. 

\subsection{Results Summary}

From Table \ref{tab:sim1} - \ref{tab:sim2}, we can see that the proposed method works uniformly better than the classical one.
The purity index is significantly higher under both Type-i and Type-ii event contamination settings.
As the number of periods $L$ gets larger, the purity index returned by our algorithm can be very close to 1, while the baseline method cannot.
The gaps between the purity indices of the two methods remain large as the number of classes $K$ increases from 4 to 6. 
When the contamination proportion $\eta$ is increased from 0.15 to 0.25, our method still returns relatively high purity values. 
Additional simulation results of the detection of event outliers and the illustration of the impact of weight functions are provided in the supplementary file.



\begin{table}[]
    \centering
    {
    \begin{tabular}{c|c|c|ccc|ccc}
    \toprule  
    \multicolumn{1}{c}{\multirow{2}*{$\eta$}}&\multicolumn{1}{|c|}{\multirow{2}*{Time}}&\multicolumn{1}{c|}{\multirow{2}*{Algorithm}}& \multicolumn{3}{c|}{Type-i}& \multicolumn{3}{c}{Type-ii}\\
    \hiderowcolors
    \cline{4-9}
    \multicolumn{1}{c}{}&\multicolumn{1}{|c|}{}&\multicolumn{1}{c|}{} & K=4 & K=5 & K=6 & K=4 & K=5 & K=6 \\
    \midrule  
     \multicolumn{1}{c}{}{\multirow{6}*{0.15}}  &\multicolumn{1}{|c|}{\multirow{3}*{L=1}}& Standard & 0.8635 & 0.8788 & 0.8910 & 0.4580 & 0.5053 & 0.5520\\
    \multicolumn{1}{c}{} &\multicolumn{1}{|c|}{} &Robust & 0.9205 & 0.9602 & 0.9663 & 0.8138 & 0.8857 & 0.9185\\
    \cline{2-9}
    \multicolumn{1}{c}{} &\multicolumn{1}{|c|}{\multirow{3}*{L=2}}& Standard & 0.8235 & 0.8523 & 0.8870 & 0.5003 & 0.5330 & 0.5928\\
    \multicolumn{1}{c}{} &\multicolumn{1}{|c|}{} &Robust & 0.9390 & 0.9735 & 0.9878 & 0.9108 & 0.9487 & 0.9713\\
   \cline{2-9}
    \multicolumn{1}{c}{} &\multicolumn{1}{|c|}{\multirow{3}*{L=4}}& Standard & 0.8600 & 0.8932 & 0.9120 & 0.6077 & 0.6783 & 0.6718\\
    \multicolumn{1}{c}{} &\multicolumn{1}{|c|}{} &Robust & 0.9297 & 0.9893 & 0.9837 & 0.9012 & 0.9287 & 0.9573\\
    
    \midrule  
    \multicolumn{1}{c}{}{\multirow{6}*{0.2}}  &\multicolumn{1}{|c|}{\multirow{3}*{L=1}}& Standard & 0.8013 & 0.8333 & 0.8480 & 0.4297 & 0.4463 & 0.5017\\
    \multicolumn{1}{c}{} &\multicolumn{1}{|c|}{} &Robust & 0.9272 & 0.9480 & 0.9562 & 0.8123 & 0.8652 & 0.9103\\
   \cline{2-9}
    \multicolumn{1}{c}{} &\multicolumn{1}{|c|}{\multirow{3}*{L=2}}& Standard & 0.7545 & 0.8343 & 0.8380 & 0.4523 & 0.4987 & 0.5260\\
    \multicolumn{1}{c}{} &\multicolumn{1}{|c|}{} &Robust & 0.9573 & 0.9772 & 0.9855 & 0.9162 & 0.9622 & 0.9625\\
    \cline{2-9}
    \multicolumn{1}{c}{} &\multicolumn{1}{|c|}{\multirow{3}*{L=4}}& Standard & 0.8562 & 0.8730 & 0.8858 & 0.5872 & 0.6107 & 0.6325\\
    \multicolumn{1}{c}{} &\multicolumn{1}{|c|}{} &Robust & 0.9297 & 0.9740 & 0.9933 & 0.8797 & 0.8930 & 0.9620\\
    \midrule  

    \multicolumn{1}{c}{}{\multirow{6}*{0.25}}  &\multicolumn{1}{|c|}{\multirow{3}*{L=1}}& Standard & 0.7520 & 0.7828 & 0.8030 & 0.3977 & 0.4527 & 0.4855\\
    \multicolumn{1}{c}{} &\multicolumn{1}{|c|}{} &Robust & 0.9132 & 0.9383 & 0.9377 & 0.8142 & 0.8610 & 0.8888\\
    \cline{2-9}
    \multicolumn{1}{c}{} &\multicolumn{1}{|c|}{\multirow{3}*{L=2}}& Standard & 0.7065 & 0.7788 & 0.8092 & 0.4535 & 0.4892 & 0.5260\\
    \multicolumn{1}{c}{} &\multicolumn{1}{|c|}{} &Robust & 0.9565 & 0.9838 & 0.9860 & 0.8538 & 0.9188 & 0.9237\\
    \cline{2-9}
    \multicolumn{1}{c}{} &\multicolumn{1}{|c|}{\multirow{3}*{L=4}}& Standard & 0.8222 & 0.8457 & 0.8743 & 0.5755 & 0.6187 & 0.6412 \\
    \multicolumn{1}{c}{} &\multicolumn{1}{|c|}{} &Robust & 0.9492 & 0.9793 & 0.9833 & 0.8287 & 0.9207 & 0.9107\\
    \bottomrule 
    \end{tabular}
    }
    \caption{Purity indices returned by two algorithms under the setting of $\eta=0.15$, $\eta = 0.2$ and $\eta=0.25$ for non-homogeneous Hawkes processes.}
    \label{tab:sim2}
\end{table}

\section{Real Data Application}\label{sec:6}

The IPTV log-data set \citep{6717182} used in our study is collected by China Telecom, a prominent Internet Protocol television (IPTV) provider in Shanghai, China. 
To ensure privacy protection, the study employs anonymized data. This log-data meticulously captures user viewing behaviors and includes anonymous user logs along with timestamps, which are accurate to the second, indicating the start times of viewing sessions. Notably, the log-data is organized on a family basis, with each family being assigned a unique user ID. In cases where families possess multiple televisions, all viewing activities are consolidated under a single user account. We selected 274 users from the dataset and gathered their household structures and watching histories from January 1, 2012, to November 30, 2012, through phone surveys conducted with the assistance of China Telecom. On average, the data reveals that each household logs between 10 and 15 viewing events per day. This extensive dataset provides a robust foundation for analyzing IPTV viewing patterns and behaviors across different household structures.

Similar to the simulation study, we apply the proposed method and the baseline method to the IPTV log-data. 
The working model chosen is the non-homogeneous Poisson process. The number of basis functions $H$ is set to 14, i.e., the time gap between two consecutive knots is 7/14 = 0.5 days. 
Since we do not know the true underlying class labels, we cannot use the purity index to evaluate the performances of the two methods.
Instead, we consider the following $L_1$-index,
\begin{equation}\label{criteria:1}
    \text{L1}_n(\alpha; \text{alg}) = \int_{0}^T \left| \hat{\lambda}_n(t)-\hat{\lambda}_{k(n)}^*(t) \right|\mathbf{I}_{| \hat{\lambda}_n(t)-\hat{\lambda}_{k(n)}^*(t) |<q_\alpha}dt,
\end{equation}
where $k(n)$ is the estimated label of sample $n$, the $\hat{\lambda}_n(t)$ is the estimated intensity function of sample $i$ via cubic spline approximation, and $\hat{\lambda}_{k(n)}^*(t)$ is the estimated intensity function of class $k(n)$. 
In \eqref{criteria:1}, $\alpha$ is used for screening out those sub-time intervals that may contain event outliers. \text{alg} refers to either the proposed method or the baseline.
Finally, we report the median $L_1$ error, i.e., $\text{median}_{1 \leq n \leq N}(\text{L1}_n(\alpha; \text{alg}))$ and 
the comparison rate, i.e.,
$ \sum_{n=1}^N \mathbf 1\{\text{L1}_n(\alpha; \text{alg1}) < \text{L1}_n(\alpha; \text{alg2})\}/ N$.
Therefore, a better algorithm should lead to a smaller median $L_1$ error and a larger comparison rate. 
The final results are summarized in Table \ref{tab:real-1-1}.
We can clearly see that the proposed method has smaller $L_1$ errors and higher comparison rates regardless of choices of $\eta$, $K$, or $\alpha$. 
This suggests that our method is consistently robust and effective. 
For difference choices of $K$, we also report the number of households in each group in Table \ref{tab:group_number}. It can be seen that the standard method can only identify two main groups regardless of the choice of $K$.
By incorporating the proposed weight function, the sizes of different classes will become more even.

\begin{table}[]
    \centering
    \begin{tabular}{c|c|cccccc}
\toprule  
\multicolumn{2}{c|}{L1-error} & K=3 & K=4 & K=5 & K=6 & K=7 & K=8 \\
\hiderowcolors
\midrule  
\multicolumn{1}{c|}{\multirow{3}*{$\alpha = 0.8$}}&  Robust & \textbf{1.4323} & \textbf{1.4288}  & \textbf{1.4276} & \textbf{1.4288} & \textbf{1.4090} & \textbf{1.4011} \\
 \multicolumn{1}{c|}{} & Standard & 1.6141 & 1.5994 & 1.6074 & 1.6026 & 1.5990 & 1.6056 \\
 \multicolumn{1}{c|}{} & Comparison & 0.6934 & 0.8029 & 0.7263 & 0.7226 & 0.6131 & 0.6861\\

\midrule  
\multicolumn{1}{c|}{\multirow{3}*{$\alpha = 0.9$}}&Robust & \textbf{1.6570} & \textbf{1.6525}  & \textbf{1.6570} & \textbf{1.6557} & \textbf{1.6357} & \textbf{1.6375} \\
 \multicolumn{1}{c|}{} &  Standard & 1.8406 & 1.8312 & 1.8372 & 1.8353 & 1.8284 & 1.8336 \\
 \multicolumn{1}{c|}{} &  Comparison & 0.6934 & 0.7920 & 0.7190 & 0.7190 & 0.6131 & 0.6715\\
\bottomrule 
\end{tabular}
    \caption{{ L1-error indices given by all two methods for IPTV data, $\alpha = 0.8$ and $\alpha=0.9$.}}
    \label{tab:real-1-1}
\end{table}

Moreover, we also examine the estimated weight functions, $1 - W_i(S_n; \hat{\boldsymbol{B}})$'s.
Curves of weight functions of two representative households are plotted in Figure \ref{fig:rep}.
Additionally, histograms of weight values are provided in Figure \ref{fig:hist}. The histograms depict the average weight values for each household over the 11-month time period. 
We can clearly see that quite many users have larger values during 01/22/2012 - 01/28/2012 or 09/30/2012 - 10/07/2012. These two time periods exactly correspond to the Chinese spring festival and Chinese national holiday.
In other words, users during holidays may have different TV-watching behaviors compared with their daily lives.

\begin{figure}[htbp]
\centering
\includegraphics[width=0.48\textwidth]{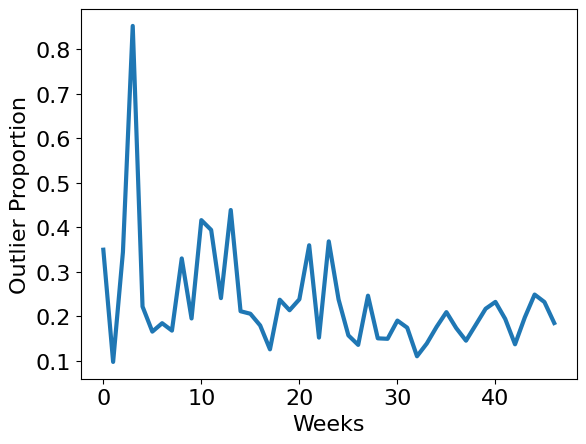}
\includegraphics[width=0.48\textwidth]{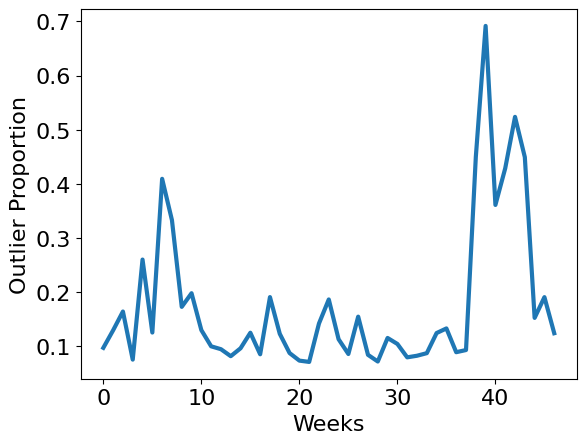}
\caption{The left household ID is 54254406, and the right household ID is 54350089. The former has larger weight values during the spring festival and the latter has larger weight values during national holiday.}\label{fig:rep}
\end{figure}

\begin{figure}[htbp]
\centering
\includegraphics[width=0.32\textwidth]{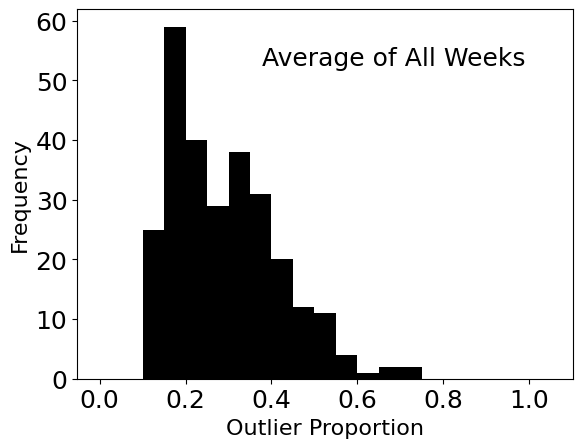}
\includegraphics[width=0.32\textwidth]{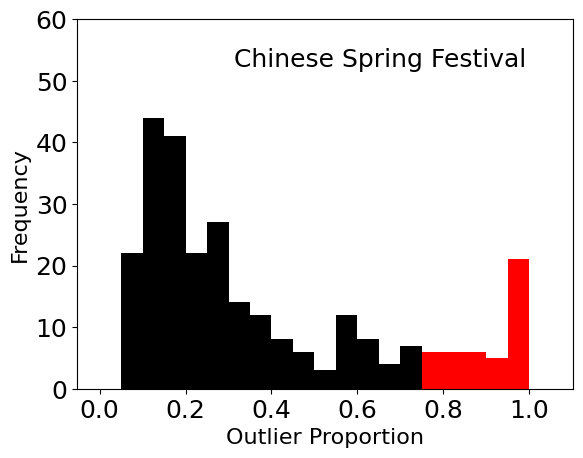}
\includegraphics[width=0.32\textwidth]{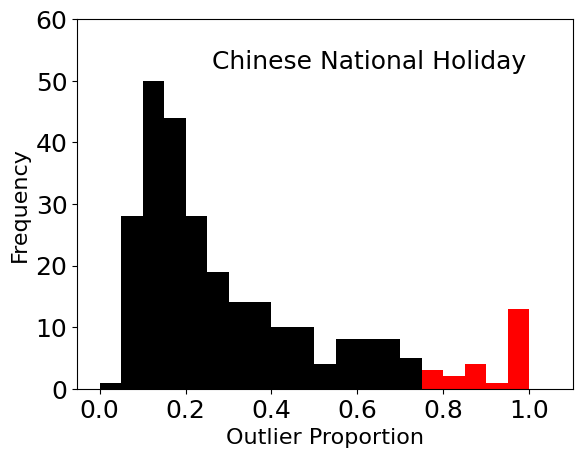}
\caption{The left panel presents a histogram depicting the weight distribution over the whole 11-month time period. The middle panel illustrates a histogram of weights specifically during the Spring Festival period, while the right panel shows the weight distribution histogram for the National Day period.
The red color on the graph indicates the weeks where the detected outlier proportion is greater than $0.75$.
}\label{fig:hist}
\end{figure}

\begin{table}[]
    \centering
    \begin{tabular}{c|p{1.5cm}|p{1.5cm}|p{1.5cm}|p{1.5cm}|p{1.5cm}|p{1.5cm}}
\toprule  
 Algorithm& K=3 & K=4 & K=5 & K=6 & K=7 & K=8 \\
\midrule  
Robust & 124, 82, 68 & 106, 79, 45, 44 & 60, 58, 56, 53, 47 & 61, 51, 44, 41, 39, 38 & 64, 47, 43, 36, 33, 27, 24 & 46, 42, 35, 33, 32, 31, 28, 27 \\
\midrule
Standard & 170, 103, 1 & 152, 92, 30, 0 & 170, 97, 6, 1, 0 & 169, 96, 7, 1, 1, 0 & 166, 88, 17, 2, 1, 0, 0 & 170, 92, 10, 2, 0, 0, 0, 0 \\
\bottomrule 
\end{tabular}
    \caption{Comparison of the number of households in each group that are clustered by the two algorithms.}
    \label{tab:group_number}
\end{table}

\section{Conclusion}\label{sec:7}

In this paper, we provide a simple-but-effective solution to study event stream data when there exist commission or omission outliers. 
We adopt the temporal point process framework by introducing the weight function to adaptively adjust the importance of each observed event. 
The proposed method is shown to have several statistical merits.
In particular, we show that it can return asymptotically unbiased estimation results when no outlier events exist.
On the other hand, when there exist event outliers, our method is much more robust compared with the vanilla one.
To the best of our knowledge, our approach is the first one to simultaneously handle both commission and omission outliers with provable theoretical guarantees.

The weight function proposed in this paper could have a broader impact in learning stream data. 
There are a few possible ways to extend the current work.
In this paper, we do not take into account event type. In future work, the marked temporal point process could be considered for a more flexible and comprehensive framework.
In the main context, we only focus on a clustering problem. 
In fact, our method could be extended to different supervised or unsupervised downstream tasks, including but not limited to, the prediction of the next event arrival, change point detection of user behaviors, etc.

\bibliography{references}


\clearpage

\appendix

{\Large
\begin{center}
Supplementary to ``Learning under Commission or Omission Event Outliers"
\end{center}
}

We provide extra information of $\phi(x)$ in Section \ref{sec:phi}, discussion on evernt outlier detection in Section \ref{sec:dec}, and additional simulation results in Section \ref{sec:addsim}.
In Section \ref{sec:pf1} - \ref{sec:pf2}, we give the proof of Theorems \ref{Thm:Mgrad} - \ref{thm:local}, and \ref{thm:tpr}. The supporting Lemmas we used are proved in Section \ref{sec:pf:lem}.

\section{Choice of $\phi(x)$}\label{sec:phi}

In the implementation, we take $a = 1$, $b = 23/3$ in \eqref{eq:robust_fun0}. 
The shape of $\phi'(x)$ is visualized in Figure \ref{fig:intensity}.

\begin{figure}[htbp]
\centering
\includegraphics[width=0.8\textwidth]{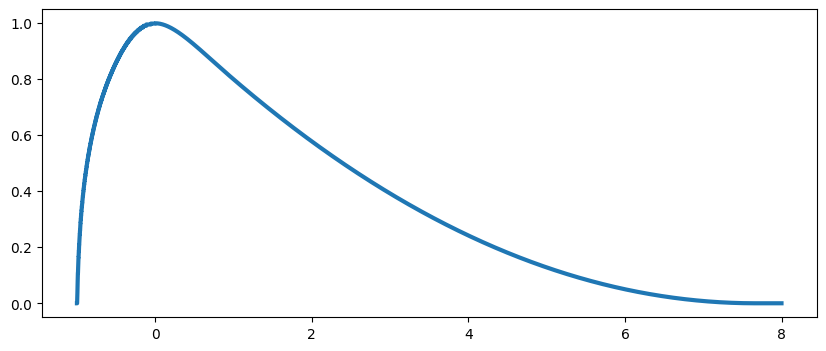}
\caption{ The visualization of $\phi'(x)$. Its domain is $[-1, +\infty)$ and the range is $[0,1]$. }\label{fig:intensity}
\end{figure}

\section{Detection of event outliers}\label{sec:dec}

For event sequence $S_n$, we define its corresponding true positive detection rate and true negative detection rate as
\begin{align}\label{eq:TPR}
    TPR(n)&=\int_{0}^{T_0} \min\{\hat{\omega}_{n}(t),{\omega}_{n}^*(t)\} dt/ \int_{0}^{T_0} {\omega}_{n}^*(t) dt,
\end{align}
and 
\begin{align}\label{eq:TNR}
    TNR(n)&=\int_{0}^{T_0} \min\{1-\hat{\omega}_{n}(t),1-{\omega}_{n}^*(t)\} dt/ \int_{0}^{T_0} 1-{\omega}_{n}^*(t) dt,
\end{align}
where
\begin{align*}
   {\omega}_{n}^*(t) :=& \mathbb{I}(t\in T_{out}) ,\\
    \hat{\omega}_{n}(t) :=& \mathbb{I}_{\{w_i(S_n;\hat{\boldsymbol B}_{k(n)})< \tilde \alpha\}},\text{ when } t\in[t_{i-1},t_{i}],
\end{align*}
$k(n)$ is the estimated class label for event sequence $S_n$, $T_{out}$ is the union set of contaminated time intervals, and $\tilde \alpha \in (0,1)$ is a fixed threshold. 
We define the integrated weight index
\begin{eqnarray}
    \text{IW}(\lambda'; \lambda) := \mathbb{E}_{S \sim \lambda' } \left[\sum_{i=1}^M \mathbf{I}_{\{\phi_{\rho_1, \rho_2}'\left( \int_{t_{i-1}}^{t_i} \lambda(u) d u-1\right)>\widetilde\alpha\}}\cdot(t_i-t_{i-1})\right],
\end{eqnarray}
where $\lambda$ and $\lambda'$ are two intensity function over $[0,T_0]$.
Then we have the following observation.

\begin{proposition}\label{thm:tprfnr}
When $\rho_1 = \rho_2$, it holds that
\begin{align*}
\text{IW}(\lambda^{\ast}; \lambda^{\ast})
> \text{IW}(\lambda'; \lambda^{\ast})
\end{align*}
for arbitrary intensity $\lambda'$ that satisfies $ \lambda'(t) < \lambda^{\ast}(t)$ for all $t \in T_{out}$ or $ \lambda'(t) > \lambda^{\ast}(t)$ for all $t \in T_{out}$.

Moreover, $\text{IW}(\lambda_0 \cdot \lambda^{\ast}; \lambda^{\ast})$ is monotonically decreasing for $\lambda_0 > 1$ and is monotonically increasing for $\lambda_0 \in (0,1)$.
\end{proposition}

Proposition \ref{thm:tprfnr} says that time intervals contaminated by outliers, whether through addition or removal, will have significantly lower weights compared to uncontaminated intervals. 
This result is crucial because it implies that the weight function effectively distinguishes between contaminated and uncontaminated intervals, assigning lower importance to the former. 
This differentiation helps us to detect the contaminated time intervals.

Intuitively, 
the relationships between $TPR(n)$ and $TNR(n)$ with the integrated weight index are through the following equations, 
$\mathbb{E}[TPR(n)] = 1-(\text{IW}(\lambda_n; \lambda^{\ast})-(1-\eta_n)\text{IW}(\lambda^{\ast}; \lambda^{\ast}))/\eta_n T_0 + o_p(1)$ and $\mathbb{E}[TNR(n)] = \text{IW}(\lambda^{\ast}; \lambda^{\ast})/T_0$. Here $\lambda_n$ and $\eta_n$ are the intensity with event outliers and contamination proportion for the $n$-th sequence, respectively.
Thanks to Proposition \ref{thm:tprfnr}, we have the following result.

\begin{theorem}\label{thm:tpr}
Define $\widehat{TP} := N^{-1}\sum_{n=1}^N TPR(n)$. 
Under Assumptions \ref{asm:1} - \ref{asm:6} and $\rho_1 = \rho_2$, it holds that 
$\mathbb E[\widehat{TP}] \rightarrow 1$ as $\lambda_{add}(t) \uparrow \infty$, and $N, L \rightarrow \infty$.
\end{theorem}

Theorem \ref{thm:tpr} shows that the detection power of Algorithm \ref{alg1} increases when there are more contaminated events. 
Especially, when the intensity of Type-ii outliers goes to infinity, the true positive rate will increase to 1. 
On the other hand, we cannot expect the true negative rate , $\widehat{TN} := N^{-1}\sum_{n=1}^N TNR(n)$, to be arbitrarily close to 1. 
This is because that $\mathbb E[\widehat{TN}] \leq \exp\{- (\phi_{\rho_1, \rho_2}')^{-1}(\tilde \alpha)\}$ for any algorithm that returns consistent model parameter estimates.

\section{Additional Simulation Results}\label{sec:addsim}

\subsection{Event outlier detection}

Under the same simulation settings described in Section \ref{sec:5} of the main context, we also report the true positive rate $\widehat{TP}$ and true negative rate $\widehat{TN}$ defined in Theorem \ref{thm:tpr} with {$\tilde \alpha = 0.6$}. 
The results of $\widehat{TP}$ and $\widehat{TN}$ under non-homogeneous Poisson processes and self-exciting processes are summarized in Table \ref{tab:sim1:tp} and Table \ref{tab:sim2:tp}, respectively.

For both types of contamination, the true positive rates become close to 1 when $L$ gets larger.
On the other hand, the true negative rates are well below 1, this phenomenon corroborates the theoretical explanations in Section \ref{sec:4}.

\begin{table}[h]
    \centering
    {
       \begin{tabular}{c|c|c|ccc|ccc}
    \toprule  
    \multicolumn{1}{c}{\multirow{2}*{$\eta$}}&\multicolumn{1}{|c|}{\multirow{2}*{Time}}&\multicolumn{1}{c|}{\multirow{2}*{Algorithm}}& \multicolumn{3}{c|}{Type-i}& \multicolumn{3}{c}{Type-ii}\\
    \hiderowcolors
    \cline{4-9}
    \multicolumn{1}{c}{}&\multicolumn{1}{|c|}{}&\multicolumn{1}{c|}{} & K=4 & K=5 & K=6 & K=4 & K=5 & K=6 \\
    \midrule  
     \multicolumn{1}{c}{}{\multirow{6}*{0.15}}  &\multicolumn{1}{|c|}{\multirow{3}*{L=1}}& TPR & 1 & 0.1 & 1 & 0.6595 & 0.6586 & 0.6631\\
    \multicolumn{1}{c}{} &\multicolumn{1}{|c|}{} &TNR & 0.7863 & 0.7990 & 0.7999 & 0.6769 & 0.7003 & 0.7073\\
    \cline{2-9}
    \multicolumn{1}{c}{} &\multicolumn{1}{|c|}{\multirow{3}*{L=2}}& TPR & 1 & 1 & 1 & 0.7712 & 0.7575 & 0.7528\\
    \multicolumn{1}{c}{} &\multicolumn{1}{|c|}{} &TNR & 0.7814 & 0.7944 & 0.7952 & 0.6869 & 0.7021 & 0.7135\\
    \cline{2-9}
    \multicolumn{1}{c}{} &\multicolumn{1}{|c|}{\multirow{3}*{L=4}}& TPR & 1 & 1 & 1 & 0.8778 & 0.8739 & 0.8695\\
    \multicolumn{1}{c}{} &\multicolumn{1}{|c|}{} &TNR & 0.7769 & 0.7888 & 0.7917 & 0.6898 & 0.7105 & 0.7045\\

    \midrule  
    \multicolumn{1}{c}{}{\multirow{6}*{0.2}}  &\multicolumn{1}{|c|}{\multirow{3}*{L=1}}& TPR & 1 & 1 & 1 & 0.6161 & 0.6062 & 0.5976\\
    \multicolumn{1}{c}{} &\multicolumn{1}{|c|}{} &TNR & 0.7805 & 0.7902 & 0.7987 & 0.6755 & 0.6990 & 0.7093\\
   \cline{2-9}
    \multicolumn{1}{c}{} &\multicolumn{1}{|c|}{\multirow{3}*{L=2}}& TPR & 1 & 0.9998 & 1 & 0.7163 & 0.7113 & 0.7110\\
    \multicolumn{1}{c}{} &\multicolumn{1}{|c|}{} &TNR & 0.7793 & 0.7902 & 0.7900 & 0.6905 & 0.7083 & 0.7148\\
    \cline{2-9}
    \multicolumn{1}{c}{} &\multicolumn{1}{|c|}{\multirow{3}*{L=4}}& TPR & 1 & 0.9999 & 0.9999 & 0.8566 & 0.8479 & 0.8495\\
    \multicolumn{1}{c}{} &\multicolumn{1}{|c|}{} &TNR & 0.7632 & 0.7843 & 0.7867 & 0.6848 & 0.7078 & 0.7156\\

    \midrule
    \multicolumn{1}{c}{}{\multirow{6}*{0.25}}  &\multicolumn{1}{|c|}{\multirow{3}*{L=1}}& TPR & 1 & 1 & 1 & 0.5803 & 0.5615 & 0.5548\\
    \multicolumn{1}{c}{} &\multicolumn{1}{|c|}{} &TNR & 0.7792 & 0.7826 & 0.7882 & 0.6806 & 0.6998 & 0.7082\\
    \cline{2-9}
    \multicolumn{1}{c}{} &\multicolumn{1}{|c|}{\multirow{3}*{L=2}}& TPR & 1 & 1 & 1 & 0.6991 & 0.6895 & 0.6757\\
    \multicolumn{1}{c}{} &\multicolumn{1}{|c|}{} &TNR & 0.7671 & 0.7800 & 0.7834 & 0.6861 & 0.7064 & 0.7123\\
    \cline{2-9}
    \multicolumn{1}{c}{} &\multicolumn{1}{|c|}{\multirow{3}*{L=4}}& TPR & 1 & 1 & 1 & 0.8471 & 0.8400 & 0.8321\\
    \multicolumn{1}{c}{} &\multicolumn{1}{|c|}{} &TNR & 0.7642 & 0.7771 & 0.7791 & 0.6908 & 0.7065 & 0.7122\\
    \bottomrule 
    \end{tabular}
    }
    \caption{{$\widehat{TP}$ and $\widehat{TN}$'s for Non-homogeneous Poisson processes}}
    \label{tab:sim1:tp}
\end{table}

\begin{table}[h]
    \centering
    {
        \begin{tabular}{c|c|c|ccc|ccc}
    \toprule  
    \multicolumn{1}{c}{\multirow{2}*{$\eta$}}&\multicolumn{1}{|c|}{\multirow{2}*{Time}}&\multicolumn{1}{c|}{\multirow{2}*{Algorithm}}& \multicolumn{3}{c|}{Type-i}& \multicolumn{3}{c}{Type-ii}\\
    \hiderowcolors
    \cline{4-9}
    \multicolumn{1}{c}{}&\multicolumn{1}{|c|}{}&\multicolumn{1}{c|}{} & K=4 & K=5 & K=6 & K=4 & K=5 & K=6 \\
    \midrule  
     \multicolumn{1}{c}{}{\multirow{6}*{0.15}}  &\multicolumn{1}{|c|}{\multirow{3}*{L=1}}& TPR & 1 & 1 & 1 & 0.7085 & 0.6911 & 0.6948\\
    \multicolumn{1}{c}{} &\multicolumn{1}{|c|}{} &TNR & 0.7199 & 0.7495 & 0.7503 & 0.5866 & 0.6167 & 0.6285\\
    \cline{2-9}
    \multicolumn{1}{c}{} &\multicolumn{1}{|c|}{\multirow{3}*{L=2}}& TPR & 1 & 1 & 1 & 0.7930 & 0.7870 & 0.7896\\
    \multicolumn{1}{c}{} &\multicolumn{1}{|c|}{} &TNR & 0.7281 & 0.7411 & 0.7505 & 0.6321 & 0.6361 & 0.6555\\
    \cline{2-9}
    \multicolumn{1}{c}{} &\multicolumn{1}{|c|}{\multirow{3}*{L=4}}& TPR & 1 & 1 & 1 & 0.9025 & 0.8997 & 0.8978\\
    \multicolumn{1}{c}{} &\multicolumn{1}{|c|}{} &TNR & 0.7232 & 0.7456 & 0.7447 & 0.6329 & 0.6439 & 0.6487\\
    
    \midrule  
    \multicolumn{1}{c}{}{\multirow{6}*{0.2}}  &\multicolumn{1}{|c|}{\multirow{3}*{L=1}}& TPR & 1 & 1 & 1 & 0.6653 & 0.6485 & 0.6530\\
    \multicolumn{1}{c}{} &\multicolumn{1}{|c|}{} &TNR & 0.7145 & 0.7329 & 0.7469 & 0.5847 & 0.6010 & 0.6199\\
    \cline{2-9}
    \multicolumn{1}{c}{} &\multicolumn{1}{|c|}{\multirow{3}*{L=2}}& TPR & 1 & 1 & 1 & 0.7792 & 0.7647 & 0.7606\\
    \multicolumn{1}{c}{} &\multicolumn{1}{|c|}{} &TNR & 0.7299 & 0.7395 & 0.7416 & 0.6056 & 0.6300 & 0.6409\\
    \cline{2-9}
    \multicolumn{1}{c}{} &\multicolumn{1}{|c|}{\multirow{3}*{L=4}}& TPR & 1 & 1 & 1 & 0.8888 & 0.8893 & 0.8819\\
    \multicolumn{1}{c}{} &\multicolumn{1}{|c|}{} &TNR & 0.7186 & 0.7345 & 0.7402 & 0.6230 & 0.6310 & 0.6579\\

    \midrule  
    \multicolumn{1}{c}{}{\multirow{6}*{0.25}}  &\multicolumn{1}{|c|}{\multirow{3}*{L=1}}& TPR & 1 & 1 & 1 & 0.6365 & 0.6346 & 0.6133\\
    \multicolumn{1}{c}{} &\multicolumn{1}{|c|}{} &TNR & 0.7080 & 0.7250 & 0.7277 & 0.5774 & 0.5964 & 0.6143\\
    \cline{2-9}
    \multicolumn{1}{c}{} &\multicolumn{1}{|c|}{\multirow{3}*{L=2}}& TPR & 1 & 1 & 1 & 0.7553 & 0.7459 & 0.7489\\
    \multicolumn{1}{c}{} &\multicolumn{1}{|c|}{} &TNR & 0.71779 & 0.7342 & 0.7372 & 0.6049 & 0.6223 & 0.6239\\
    \cline{2-9}
    \multicolumn{1}{c}{} &\multicolumn{1}{|c|}{\multirow{3}*{L=4}}& TPR & 1 & 1 & 1 & 0.8794 & 0.8778 & 0.8745\\
    \multicolumn{1}{c}{} &\multicolumn{1}{|c|}{} &TNR & 0.7147 & 0.7279 & 0.7304 & 0.6041 & 0.6307 & 0.6347 \\
    \bottomrule 
    \end{tabular}
    }
    \caption{{ $\widehat{TP}$ and $\widehat{TN}$'s for non-homogeneous Hawkes processes}}
    \label{tab:sim2:tp}
\end{table}

\subsection{Gradient Bias}

In Theorem \ref{cor:grad}, we show that the absolute value of gradient (evaluated at the true parameter) of the proposed method is smaller than that of the vanilla one entry-wisely, when the outlier intensity $\lambda_{miss}(t)$ and $\lambda_{add}(t)$ is proportional to $\lambda_k^{\ast}(t)$. 
In this section, we provide an additional numerical study to illustrate such phenomenon also holds when $\lambda_{miss}(t)$ or 
$\lambda_{add}(t)$ take various form.

In particular, we compute the ratio 
$ \|\varrho(\boldsymbol{B}_k)\|_2 / \|\overline \varrho(\boldsymbol{B}_k)\|_2$, 
where $\varrho(\boldsymbol{B}_k)$ and $\overline \varrho(\boldsymbol{B}_k)$
are defined in \eqref{def:grad:all} and \eqref{def:grad:noweight}, respectively.
Here $\|\|_2$ stands for the $\ell_2$ norm and $\boldsymbol{B}_k$ are randomly chosen from $\boldsymbol{B}_k^* +[e_1,\cdots,e_K]^\top$, where $e_k \sim U(-0.1,0.1)$, $\forall k \in [K]$.
True intensity function $\lambda_k^{\ast} = 2\exp(-{(t-6)^2}/{10})+5\exp(-{(t-20)^2}/{10})+\exp(-{t^2}).$
Eight forms of outlier event intensities, including four type-i and four type-ii contamination, are considered. Their detailed formula are given in Table \ref{tab:sim:grad}.

By Table \ref{tab:sim:grad}, we could see that the ratio is well below 1 and it means that the gradient evaluated around the true value is much smaller than that without weight functions.
This suggests the weight functions in the proposed method could lead to the smaller bias. 
As we increase the outlier intensity (i.e. $c$ changes from 1 to 4), the ratio decreases to zero. It indicates that the performance of the vanilla method (with no weight function) could become very worse.
As a result, the  weight function $W_i(S; \boldsymbol{B})$'s could make a positive impact on the gradient computation.

\begin{table}[bt]
    \centering
    {
    \begin{tabular}{c|c|c|c|c|c}
    \toprule  
    \multicolumn{2}{c|}{\multirow{1}*{Method}}& Robust/ Standard &  Robust/ Standard & Robust/ Standard & Robust/ Standard  \\
    \midrule
    \hiderowcolors
    \multicolumn{1}{c}{\multirow{4}*{Type-ii}}&\multicolumn{1}{|c|}{\multirow{1}*{c}}& \multicolumn{1}{c|}{$c\lambda_k^*(t)$}& \multicolumn{1}{c|}{$c$}& \multicolumn{1}{c|}{$\frac{5c}{\sqrt{2\pi}}\exp(-\frac{(t-t_{b}-\eta T/2)^2}{2})$}&  \multicolumn{1}{c}{$4c\exp(t_b-t)$}\textbf{}\\
    \cline{2-6}
     \multicolumn{1}{c}{}{\multirow{3}*{}}  & \multicolumn{1}{|c|}{1} & 0.5078 & 0.4992 & 0.5788 & 0.4509 \\
    \multicolumn{1}{c}{} &\multicolumn{1}{|c|}{2}  & 0.5002 & 0.4544 & 0.4238 & 0.3780\\
    \multicolumn{1}{c}{} &\multicolumn{1}{|c|}{4}  & 0.4492 & 0.3949 & 0.3541 & 0.3288\\
    
    \midrule  
    \multicolumn{1}{c}{\multirow{4}*{Type-i}}&\multicolumn{1}{|c|}{\multirow{1}*{c}}& \multicolumn{1}{c|}{$(c+1)^{-1}\lambda_k^*(t)-\lambda_k^*(t)$}& \multicolumn{1}{c|}{$-c/2$}& \multicolumn{1}{c|}{$-\frac{2.5c}{\sqrt{2\pi}}\exp(-\frac{(t-t_{b}-\eta T/2)^2}{2})$}& \multicolumn{1}{c}{$-2c\exp(t_b-t)$}\textbf{}\\
    \cline{2-6}
    \multicolumn{1}{c}{}{\multirow{3}*{}}  & \multicolumn{1}{|c|}{1} & 0.4211 & 0.4822 & 0.5333 & 0.6279\\
    \multicolumn{1}{c}{} &\multicolumn{1}{|c|}{2} & 0.3222 & 0.4001 & 0.3959 & 0.4941\\
    \multicolumn{1}{c}{} &\multicolumn{1}{|c|}{4} & 0.2287 & 0.2506 & 0.2663 & 0.3960\\
    \bottomrule 
    \end{tabular}
    }
    \caption{The ratio of $\ell_2$-norm of gradients of the two methods with $\eta=0.2$ and $N=200$. The table lists the $\lambda_{miss}$, $\lambda_{add}$ functions, where the event streams are sampled according to the intensity $\lambda(t) = \max\{\lambda_k^*(t) - \lambda_{miss}(t), 0\}$ or $\lambda(t) =\lambda_k^*(t) + \lambda_{add}(t)$. And outliers are in the interval $[t_b,t_b+\eta T]$ where $t_b$ is randomly generated for each event sequence. 
    Each case are repeated for $50$ times.}
    \label{tab:sim:grad}
\end{table}

\clearpage

\section{Proof of Theorems \ref{Thm:Mgrad} - \ref{cor:grad}}\label{sec:pf1}


\noindent \textbf{Proof of Theorem \ref{Thm:Mgrad}}



For simplicity, we only prove the case when the working model is non-homogeneous Poisson process.
The proof remains unchanged for self-exciting process model, if we treat 
$\sum_{t_j < t} g_{h}(t-t_j)$ as $\kappa_h(t)$.

For $S$ sampled from $\lambda_k^{\ast}(t)$, its gradient is given by
\begin{align}\label{eq:grad}
\boldsymbol{G}:=\left(\begin{array}{c}
\sum_{i=1}^M \phi'\left(\int_{t_{i-1}}^{t_i} \lambda_k(u) d u-1\right)\cdot\left( \frac{\kappa_1(t_{i-1})}{\lambda_{k}(t_{i-1})}-\int_{t_{i-1}}^{t_i} \kappa_1(x)dx\right)-\phi'\left(\int_{0}^{t_1} \lambda_k(u) d u-1\right)\int_{0}^{t_1} \kappa_1(x)dx\\
\vdots\\
\sum_{i=1}^M\phi'\left( \int_{t_{i-1}}^{t_i} \lambda_k(u) d u-1\right)\cdot\left( \frac{\kappa_H(t_{i-1})}{\lambda_{k}(t_{i-1})}-\int_{t_{i-1}}^{t_i} \kappa_H(x)dx\right)-\phi'\left(\int_{0}^{t_1} \lambda_k(u) d u-1\right)\int_{0}^{t_1} \kappa_H(x)dx\\
\end{array}\right)^\top /L.
\end{align}
 
 Consider the first entry in $\boldsymbol{G}$, when $\boldsymbol{B}_k=\boldsymbol{B}_k^*$, it can be written as \[\sum_{i=1}^M \phi'\left(\int_{t_{i-1}}^{t_i} \lambda_k^*(u) d u-1\right)\cdot\left( \frac{\kappa_1(t_{i-1})}{\lambda_{k}^*(t_{i-1})}-\int_{t_{i-1}}^{t_i} \kappa_1(x)dx\right)-\phi'\left(\int_{0}^{t_1} \lambda^*_k(u) d u-1\right)\int_{0}^{t_1} \kappa_1(x)dx.\]
We define $\Lambda_k^*(t):=\int_0^t \lambda_k^*(t)dt$. When $S_n$ is a counting process with intensity function $\lambda_k^*$, we know that $\left\{\Lambda_k^*(t_1),\cdots\right\} $ is a standard Poisson process on $\left[0,\int_0^{L\cdot T}\lambda_k^*(t)dt\right]$.

Then we will use $\Lambda_k^*(t)$ to replace $t$ to discuss. We define $ \widetilde{\lambda_k^*}({\Lambda_k^*}(t)) =  \lambda_k^*(t)$ and $\widetilde{\kappa_h}({\Lambda_k^*}(t)) = \kappa_h^*(t)$. We have 

\begin{align*}
&\mathbb{E}\left[ \sum_{i=1}^M \phi'\left(\int_{t_{i-1}}^{t_i} \lambda_k^*(t) d t-1\right)\cdot\left( \frac{\kappa_1(t_{i-1})}{\lambda_{k}^*(t_{i-1})}-\int_{t_{i-1}}^{t_i} \kappa_1(t)dt\right)-\phi'\left(\int_{0}^{t_1} \lambda^*_k(u) d u-1\right)\int_{0}^{t_1} \kappa_1(x)dx\right]\\
=& \mathbb{E}\left[ \sum_{i=1}^M \phi'\left({\Lambda_k^*}(t_{i}) -{\Lambda_k^*}(t_{i-1}) -1\right)\cdot\left( \frac{\widetilde{\kappa_1}({\Lambda_k^*}(t_{i-1}))}{\widetilde{\lambda_k^*}({\Lambda_k^*}(t_{i-1})) }-\int_{{\Lambda_k^*}(t_{i-1}) }^{{\Lambda_k^*}(t_i) } \frac{\widetilde{\kappa_1}({\Lambda_k^*}(t))}{\widetilde{\lambda_k^*}({\Lambda_k^*}(t)) }d\Lambda_k^*(t) \right)\right]\\
&- \mathbb{E}\left[\phi'\left({\Lambda_k^*}(t_{1})-1\right)\cdot\int_{0}^{{\Lambda_k^*}(t_1) } \frac{\widetilde{\kappa_1}({\Lambda_k^*}(t))}{\widetilde{\lambda_k^*}({\Lambda_k^*}(t)) }d\Lambda_k^*(t)\right].\\
\end{align*}

For simplicity, we write ${\widetilde{\kappa_1}({\Lambda_k^*}(t))}/{\widetilde{\lambda_k^*}({\Lambda_k^*}(t)) } $ as $ \chi(\Lambda_k^*(t))$, $\forall t \in [0,L\cdot T]$, and $\left\{\widetilde{t}_1,\cdots\right\} :=\left\{\Lambda_k^*(t_1),\cdots\right\} $. We define $\widetilde{N}(\widetilde{t}) = N({\Lambda_k^*}^{-1}(\widetilde{t})) $, then we know that $\mathbb{P}(\widetilde{N}(\widetilde{t}+d\widetilde{t})-\widetilde{N}(\widetilde{t})=1)=\mathbb{P}(N({\Lambda_k^*}^{-1}(\widetilde{t}+d\widetilde{t}))-N({\Lambda_k^*}^{-1}(\widetilde{t}))=1)=d\Lambda_k^*({\Lambda_k^*}^{-1}(\widetilde{t}))=d\widetilde{t}$.
Then we get that

{
\begin{align*}
& \mathbb{E}\left[ \sum_{i=1}^M \phi'\left({\Lambda_k^*}(t_{i}) -{\Lambda_k^*}(t_{i-1}) -1\right)\cdot\left(  \chi(\Lambda_k^*(t_{i-1}))-\int_{{\Lambda_k^*}(t_{i-1}) }^{{\Lambda_k^*}(t_i) }  \chi(\Lambda_k^*(t))d\Lambda_k^*(t) \right)\right]\\
=& \mathbb{E}\left[ \sum_{i=1}^M \phi'\left(\widetilde{t}_i -\widetilde{t}_{i-1} -1\right)\cdot\left(  \chi(\widetilde{t}_{i-1})-\int_{\widetilde{t}_{i-1} }^{\widetilde{t}_i}  \chi(\widetilde{s})d\widetilde{s} \right)\right]  \qquad\text{\footnotesize{(variable substitution)}} \\
=& \mathbb{E}\left[ \sum_{i=1}^M \mathbb{E}\left[ \phi'\left(\widetilde{t}_i -\widetilde{t}_{i-1} -1\right)\cdot\left(  \chi(\widetilde{t}_{i-1})-\int_{\widetilde{t}_{i-1} }^{\widetilde{t}_i}  \chi(\widetilde{s})d\widetilde{s} \right)\mid\mathcal{H}_{\widetilde{t}_{i-1}}\right]\right]   \qquad\text{\footnotesize{(take the conditional expectation respectively)}} \\
=& \mathbb{E}\left[ \sum_{i=1}^M f(\widetilde{t}_{i-1})\right]  = \mathbb{E}\left[ \int f(\widetilde{t})  d\widetilde{N}(\widetilde{t}) \right] =  \int_0^{L\Lambda_k^*(T)} f(\widetilde{t})  d\widetilde{t} , \\
\end{align*}
}
where $f(\widetilde{t}_{i-1}):=\mathbb{E}\left[ \phi'\left(\widetilde{t}_i -\widetilde{t}_{i-1} -1\right)\cdot\left(  \chi(\widetilde{t}_{i-1})-\int_{\widetilde{t}_{i-1} }^{\widetilde{t}_i}  \chi(\widetilde{s})d\widetilde{s} \right)\mid\mathcal{H}_{\widetilde{t}_{i-1}}\right] $ is a function of $\widetilde{t}_{i-1}$.

Because $\widetilde{t}_{i} -\widetilde{t}_{i-1}$ follows the standard exponential distribution, we can write $f(\widetilde{t})$ as

$\mathbb{E}_{X\sim \text{EXP(1)}}\left[ \phi'\left(X -1\right)\cdot\left(  \chi(\widetilde{t})-\int_{\widetilde{t} }^{\widetilde{t}+X}  \chi(\widetilde{s})d\widetilde{s} \right)\right]$. We swap the order of integration of $\widetilde{t}$ and $\widetilde{s}$ and get

\begin{align*}
& \left(\int_0^{L\Lambda_k^*(T)} \mathbb{E}_{X\sim \text{EXP(1)}}\left[ \phi'\left(X -1\right)\cdot\left(  \chi(\widetilde{t})-\int_{\widetilde{t} }^{\widetilde{t}+X}  \chi(\widetilde{s})d\widetilde{s} \right)\right]  d\widetilde{t} -\mathbb{E}\left[ \phi'\left(\widetilde{t}_1 -1\right)\cdot\int_{0}^{\widetilde{t}_1 }  \chi(\widetilde{s})d\widetilde{s} \right]\right)/L\\
&=  \mathbb{E}_{X\sim \text{EXP(1)}}\left[ \phi'\left(X -1\right)\left(   \int_0^{L\Lambda_k^*(T)}\chi(\widetilde{t})d\widetilde{t}-\int_{0 }^{X}   \int_0^{L\Lambda_k^*(T)}\chi(\widetilde{t}+\widetilde{s})d\widetilde{t}d\widetilde{s} \right)\right]   /L \\
&-\mathbb{E}_{X\sim \text{EXP(1)}}\left[ \phi'\left(X -1\right) \int_{0}^{X }  \chi(\widetilde{s})d\widetilde{s} \right]/L  \qquad\text{\footnotesize{(swap the order of integration of $\widetilde{t}$ and $\widetilde{s}$)}}\\
&= \int_0^{ L\cdot\Lambda_k^*(T)} \chi(\widetilde{t})d \widetilde{t}\cdot\mathbb{E}_{X\sim \text{EXP(1)}}\left[ \phi'\left(X -1\right)\cdot\left( 1-\int_{0}^{X  }  d\widetilde{s} \right)\right]/L   \quad\text{\footnotesize{(extract the part that is irrelevant to the expectation)}}\\
&+ \left(\mathbb{E}_{X\sim \text{EXP(1)}}\left[ \phi'\left(X -1\right) \int_{0}^{X }\left(\int_{0}^{\widetilde{s}}  \chi(u)du\right) d\widetilde{s} \right]-\mathbb{E}_{X\sim \text{EXP(1)}}\left[ \phi'\left(X -1\right) \int_{0}^{X }  \chi(\widetilde{s})d\widetilde{s} \right]\right)/L \\
&=0+C_{g}/L,
\end{align*}
where $C_g:=\mathbb{E}_{X\sim \text{EXP(1)}}\left[ \phi'\left(X -1\right) \int_{0}^{X }\left(\int_{0}^{\widetilde{s}}  \chi(u)du\right) d\widetilde{s} \right]-\mathbb{E}_{X\sim \text{EXP(1)}}\left[ \phi'\left(X -1\right) \int_{0}^{X }  \chi(\widetilde{s})d\widetilde{s} \right]$.
Here we use the property that $  \int_0^{L\Lambda_k^*(T)}\chi(\widetilde{t})d\widetilde{t} = \int_0^{L\Lambda_k^*(T)-\widetilde{s}}\chi(\widetilde{t}+\widetilde{s})d\widetilde{t}+\int_0^{\widetilde{s}}\chi(\widetilde{t})d\widetilde{t} $ and $\widetilde{t}_1$ follow the standard exponential distribution.

Combining the results, we get that $\mathbb{E} \left[\sum_i \phi'\left(\int_{t_{i-1}}^{t_i} \lambda_k^*(u) d u-1\right)\cdot\left( \frac{\kappa_1(t_{i-1})}{\lambda_{k}^*(t_{i-1})}-\int_{t_{i-1}}^{t_i} \kappa_1(x)dx\right)\right]/L$ goes to $0$ as $L\rightarrow\infty$ when $\rho_1=\rho_2$.

\bigskip

\noindent \textbf{Proof of Theorem \ref{Thm:Mout}}

When $\boldsymbol{B}_k=\boldsymbol{B}_k^*$, the first item of the gradient $\varrho(\boldsymbol{B}_k^{\ast})$ can be written as ${N}^{-1} \sum_{n: Z_n = k}\sum_i \phi'\left(\int_{t_{n,i-1}}^{t_{n,i}} \lambda_k^*(u) d u-1\right)\cdot\left( \frac{\kappa_1(t_{n,i-1})}{\lambda_{k}^*(t_{n,i-1})}-\int_{t_{n,i-1}}^{t_{n,i}} \kappa_1(x)dx\right)/L$ referring to equation \eqref{eq:grad}. 
Next we consider the gradient of one sample $S:=\{t_1,t_2\cdots\,t_M\}$ from cluster $k$. We write $[0,L\cdot T] = T_{in}\cup T_{out}$, while the outliers only appear in $T_{out}$. Assume that when there are no outliers, the events are $\{t'_1,t'_2 \cdots,t'_M\}$, then with a high probability we have
\begin{align*}
& \sum_{i=1}^M \phi'\left(\int_{t_{i-1}}^{t_i} \lambda_k^*(u) d u-1\right)\cdot\left(\frac{\kappa_1(t_{i-1})}{\lambda_{k}^*(t_{i-1})}-\int_{t_{i-1}}^{t_i} \kappa_1(x)dx\right)/L\\
&=  \underbrace{(\sum_{t'_i\in T_{in}} +\sum_{t'_i\in T_{out}})\phi'\left(\int_{t'_{i-1}}^{t'_i} \lambda_k^*(u) d u-1\right)\cdot\left(\frac{\kappa_1(t'_{i-1})}{\lambda_{k}^*(t'_{i-1})}-\int_{t'_{i-1}}^{t'_i} \kappa_1(x)dx\right)/L}_{\delta_0}\\
&+  \underbrace{\sum_{t_i\in T_{out}}\phi'\left(\int_{t_{i-1}}^{t_i} \lambda_k^*(u) d u-1\right)\cdot\left(\frac{\kappa_1(t_{i-1})}{\lambda_{k}^*(t_{i-1})}-\int_{t_{i-1}}^{t_i} \kappa_1(x)dx\right)/L}_{\delta_1}\\
&-  \underbrace{\sum_{t'_i\in T_{out}}\phi'\left(\int_{t'_{i-1}}^{t'_i} \lambda_k^*(u) d u-1\right)\cdot\left(\frac{\kappa_1(t'_{i-1})}{\lambda_{k}^*(t'_{i-1})}-\int_{t'_{i-1}}^{t'_i} \kappa_1(x)dx\right)/L}_{\delta_2}.\\
\end{align*}

We consider ${\delta_0} $ firstly. {When there is no outlier,
we note that  }
\begin{align*}
    \delta_0&=\sum_{i=1}^M \phi'\left(\int_{t'_{i-1}}^{t'_i} \lambda_k^*(u) d u-1\right)\cdot\left(\frac{\kappa_1(t'_{i-1})}{\lambda_{k}^*(t'_{i-1})}-\int_{t'_{i-1}}^{t'_i} \kappa_1(x)dx\right)/L\\
    &=\sum_{l=1}^{L} \sum_{(l-1)T\leq t'_i<lT} \phi'\left(\int_{t'_{i-1}}^{t'_i} \lambda_k^*(u) d u-1\right)\cdot\left(\frac{\kappa_1(t'_{i-1})}{\lambda_{k}^*(t'_{i-1})}-\int_{t'_{i-1}}^{t'_i} \kappa_1(x)dx\right)/L=\sum_{l=1}^{L} X_l/L,
\end{align*}
where $X_l$ are independent. According to Lemma \ref{lem:log-fun}, there exists $c_1>0$ such that
\begin{align*}
    \mathbb{P}\left(|\delta_0-\mathbb{E}(\delta_0)|\geq t\right)\leq 2\exp{\left(-\frac{t L}{c_1}\right)}.
\end{align*}
Then we know that $\delta_0\leq  O(c_1/\sqrt{L}+C_{g}/L) $ with a high probability according to Theorem \ref{Thm:Mgrad}.

Next we consider the bound of $\delta_1$ and $\delta_2$.
When there are outliers, according to the definition in \eqref{eq:robust_fun}, we know that there exists an upper bound $c_{up}$ and a lower bound $c_{low}$ such that $c_{low}\leq(1/x-1)\cdot\phi'(x-1)\leq c_{up}$.

Because of $\sum_{t_i\in T_{out}}\left|t_i-t_{i-1}\right|\leq \eta T$, then we have 
\begin{align*}
\delta_1&= \frac{1}{L}\sum_{t_i\in T_{out}}(t_i-t_{i-1})\cdot\phi'\left(\int_{t_{i-1}}^{t_i} \lambda_k^*(u) d u-1\right)\left(\frac{\kappa_1(t_{i-1})}{\lambda_{k}^*(t_{i-1})}-\int_{t_{i-1}}^{t_i} \kappa_1(x)dx\right)/\left(t_i-t_{i-1}\right)\\
&\leq\frac{1}{L}\sum_{t_i\in T_{out}}(t_i-t_{i-1})\cdot\max_i\{\phi'\left(\int_{t_{i-1}}^{t_i} \lambda_k^*(u) d u-1\right)\left(\frac{\kappa_1(t_{i-1})}{\lambda_{k}^*(t_{i-1})}-\int_{t_{i-1}}^{t_i} \kappa_1(x)dx\right)/\left(t_i-t_{i-1}\right)\}\\
&\leq c_{up}\cdot\eta T\kappa_{\max}/\tau.
\end{align*}
Similarly, we also know that $\delta_1 \geq c_{low}\cdot\eta T\kappa_{\max}/\tau$.
In summary, we know that $|\delta_1|\leq \eta \cdot\max\{|c_{low}|,|c_{up}|\} T\kappa_{\max}/\tau$, and we also have that $|\delta_2|\leq \eta \cdot\max\{|c_{low}|,|c_{up}|\} T\kappa_{\max}/\tau$. We write $ C_\rho:= (|c_{low}|+|c_{up}|)\cdot T\kappa_{\max}/\tau$, and then we know that for each sample $n$,
$\|\sum_{i=1}^M\{ w_i(S_n;\boldsymbol{B}_k) \nabla_{B_k} (\log \lambda_k(t_{n,i-1}) - \int_{t_{n,i-1}}^{t_{n,i}} \lambda_k(u) du ) \}\|/L=O_p(\sqrt{H}\cdot(C_{\rho}\eta + c_1/\sqrt{NL}+C_g/L) )$.
Then we get that $\|\sum_{n: Z_n = k} \sum_i \{ w_i(S_n;\boldsymbol{B}_k) \nabla_{B_k} (\log \lambda_k(t_{n,i-1}) - \int_{t_{n,i-1}}^{t_{n,i}} \lambda_k(u) du ) \}\|/(NL)= O_p(\sqrt{H}\cdot(C_{\rho}\eta + c_1/\sqrt{NL} +C_g/L) )$.

\bigskip

\noindent \textbf{Proof of Theorem \ref{cor:grad}}

First we claim that $\left|\mathbb{E}[(1-X)\phi'(X-1)]\right|<\left|\mathbb{E}[(1-X)]\right|$ when $X\sim\text{Exp}(\lambda)$. 

To see this, we
define $f(\lambda):=\int_0^\infty (1-x)\lambda\exp(-\lambda x)\phi'(x-1)dx \equiv \mathbb{E}[(1-X)\phi'(X-1)]$.
We know that $\int_0^\infty \lambda(1-x)x^{\lambda-1}\exp(-\lambda x)\phi'(x-1)dx =\int_0^\infty \phi'(x-1)d(x^\lambda\exp(-\lambda x))=0$. 

When $\lambda>1$,
\begin{align*}
    f(\lambda)&=\int_0^\infty (1-x)\lambda\exp(-\lambda x)\phi'(x-1)dx-\int_0^\infty \lambda(1-x)x^{\lambda-1}\exp(-\lambda x)\phi'(x-1)dx\\
    &= \int_0^\infty (1-x)(1-x^{\lambda-1})\lambda\exp(-\lambda x)\phi'(x-1)dx\\
    &< \int_0^\infty (1-x)(1-x^{\lambda-1})\lambda\exp(-\lambda x)dx\\
    &=1-1/\lambda \equiv \mathbb{E}[(1-X)],
\end{align*}
and it is easy to know that $f(\lambda)=\int_0^\infty (1-x)(1-x^{\lambda-1})\lambda\exp(-\lambda x)\phi'(x-1)dx>0$.

Similarly, when $ \lambda<1$,
\begin{align*}
    f(\lambda)&=\int_0^\infty (1-x)\lambda\exp(-\lambda x)\phi'(x-1)dx-\int_0^\infty \lambda(1-x)x^{\lambda-1}\exp(-\lambda x)\phi'(x-1)dx\\
    &= -\int_0^\infty (1-x)(x^{\lambda-1}-1)\lambda\exp(-\lambda x)\phi'(x-1)dx\\
    &> -\int_0^\infty (1-x)(x^{\lambda-1}-1)\lambda\exp(-\lambda x)dx\\
    &=1-1/\lambda \equiv \mathbb{E}[(1-X)],
\end{align*}
and $f(\lambda)=-\int_0^\infty (1-x)(x^{\lambda-1}-1)\lambda\exp(-\lambda x)\phi'(x-1)dx<0$.
This concludes the claim.


Assume that the actual intensity function with outliers of sample $n$ is $\widetilde{\lambda}_n(t)$, where $\widetilde{\lambda}_n(t)=h\lambda_{k}^*(t), \forall t\in T_{out}$ and $\widetilde{\lambda}_n(t)=\lambda_{k}^*(t), \forall t\in T_{in}$, and assume that all times $t$ have the same probability of appearing in $T_{out}$, which is $\eta$. Similar to the proof of Theorem \ref{Thm:Mgrad}, when adding any outliers, we have that
\begin{align*}
    &\mathbb{E}\left[ \sum_{i=1}^M \phi'\left(\int_{t_{i-1}}^{t_i} \lambda_k^*(u) d u-1\right)\cdot\left(\frac{\kappa_1(t_{i-1})}{\lambda_{k}^*(t_{i-1})}-\int_{t_{i-1}}^{t_i} \kappa_1(x)dx\right)/L\right]   \\
    =&\mathbb{E}\left[ \sum_{t_i\in T_{in}} \phi'\left(\int_{t_{i-1}}^{t_i} \lambda_k^*(u) d u-1\right)\cdot\left(\frac{\kappa_1(t_{i-1})}{\lambda_{k}^*(t_{i-1})}-\int_{t_{i-1}}^{t_i} \kappa_1(x)dx\right)/L\right] \\
    &+\mathbb{E}\left[ \sum_{t_i\in T_{out}} \phi'\left(\int_{t_{i-1}}^{t_i} \lambda_k^*(u) d u-1\right)\cdot\left(\frac{\kappa_1(t_{i-1})}{\lambda_{k}^*(t_{i-1})}-\int_{t_{i-1}}^{t_i} \kappa_1(x)dx\right)/L\right]\\
    \leq &(1-\eta)C_g/L \qquad\text{\footnotesize{(refer to Theorem \ref{Thm:Mout})}}\\
    +& \eta\int_0^{ L\cdot\Lambda_k^*(T)} \frac{\kappa_1(\Lambda_k^*(t))}{\lambda_{k}^*(\Lambda_k^*(t))}d \Lambda_k^*(t)\cdot h\mathbb{E}_{X\sim \text{EXP(h)}}\left[ \phi'\left(X -1\right)\cdot\left( 1-X \right)\right]/L \\
    &\qquad\text{\footnotesize{($\int_{t_{i-1}}^{t_i} \lambda_k^*(u) d u = \int_{t_{i-1}}^{t_i} \widetilde{\lambda}_n(t) d u/h \sim \text{EXP(h)}$ because of $\widetilde{\lambda}_n(t)=h \lambda_{k}^*(t)$)}}\\
    <& \int_0^{ L\cdot\Lambda_k^*(T)} \eta(h-1)\cdot\frac{\kappa_1(\Lambda_k^*(t))}{\lambda_{k}^*(\Lambda_k^*(t))}d \Lambda_k^*(t)/L+ (1-\eta)C_g/L\\
    &\qquad\text{\footnotesize{(the properties of the function $f(\lambda)$)}}\\
    =&\mathbb{E}\left[ \sum_{i=1}^M \left(\frac{\kappa_1(t_{i-1})}{\lambda_{k}^*(t_{i-1})}-\int_{t_{i-1}}^{t_i} \kappa_1(x)dx\right)/L\right]+(1-\eta)C_g/L.
\end{align*}
That is, the absolute value of the expected robust gradient is always less than that of the original gradient at the true value $ \lambda_{k}^*$ as $L \rightarrow \infty$. The conclusion still holds for the omission case. 

{Similar conclusions can be drawn for other samples $S_n$ and other components $h\in[H]$,} and then we can get that $ | \mathbb E[\varrho(\boldsymbol{B}_k^{\ast})]| < |\mathbb E[\overline \varrho(\boldsymbol{B}_k^{\ast})]|$ as $L\rightarrow\infty$.

\section{Proof of Theorem \ref{thm:local}}\label{sec:pf2}

An overview of the proof is summarized here. We derived an upper bound for {$\|\nabla r_i(S,\boldsymbol{B}^{(t)})\|$} when $\|\boldsymbol{B}^{(t)}_k-\boldsymbol{B}_k^*\| <a/(T\cdot\kappa_{\max})$ using Lemma \ref{lem:weight} and \ref{lem:cluster}. Subsequently, in Lemma \ref{thm:gamma}, we established an upper bound for the gradient of the objective function over M-steps. Lemma \ref{lem:grad:concen} further provided an estimate for the error range of the empirical gradient estimation. By combining these lemma together with the theory of convex optimization, we demonstrated local convergence as stated in Theorem \ref{thm:local}.

{
Define the weight as $r_k(S;\boldsymbol{B})={\pi_k \operatorname{WTPP}(S\mid \boldsymbol{B}_k,W)}/{\sum_{k=1}^K\pi_k \operatorname{WTPP}(S\mid \boldsymbol{B}_k, W)}$.
}

\begin{lemma}\label{lem:weight}
    If $\|\boldsymbol{B}^{(t)}_k-\boldsymbol{B}_k^*\| <a/(T\cdot\kappa_{\max})$, $\forall k\in [K]$, then for $p=0,1,2$ there exists a constant $G>0$ such that
\begin{align*}
    \mathbb{E}_S\left[ r_k(S ; \boldsymbol{B}^{(t)})\left(1-r_k(S ; \boldsymbol{B}^{(t)})\right)\left\|\frac{\partial \log \operatorname{WTPP}({S} \mid \boldsymbol{B}^{(t)}_k, W(S;\boldsymbol B^{(t)}_k))}{\partial \boldsymbol{B}_k^{ }}\right\|^p\right]= O(L(S)^p\exp(-G\cdot L)).
\end{align*}
    
\end{lemma}
\noindent \textbf{Proof of Lemma \ref{lem:weight}}
Without loss of generality, we prove the claim for $k=1$. Taking the expectation of $S$, we get
$$
\begin{aligned}
& \mathbb{E}_S \left[r_1(S ; \boldsymbol{B}^{(t)})\left(1-r_1(S ; \boldsymbol{B}^{(t)})\right)\left\|\frac{\partial \log \operatorname{WTPP}({S} \mid \boldsymbol{B}_1^{(t)}, w(S;\boldsymbol B_1^{(t)}))}{\partial \boldsymbol{B}_1}\right\|^p\right] \\
= & \sum_{k \in[K]} \pi_k \mathbb{E}_{s \sim \mathcal{POI}\left(\boldsymbol{B}_k^*\right)} \left[r_1(S ; \boldsymbol{B}^{(t)})\left(1-r_1(S ; \boldsymbol{B}^{(t)})\right)\left\|\frac{\partial \log \operatorname{WTPP}({S} \mid \boldsymbol{B}_1^{(t)}, w(S;\boldsymbol B_1^{(t)}))}{\partial \boldsymbol{B}_1}\right\|^p \right] \\
\leq & \pi_1  \mathbb{E}_{s \sim \mathcal{POI}\left(\boldsymbol{B}_1^*\right)}\left[ r_1(S ; \boldsymbol{B}^{(t)})\left(1-r_1(S ; \boldsymbol{B}^{(t)})\right)\left\|\frac{\partial \log \operatorname{WTPP}({S} \mid \boldsymbol{B}_1^{(t)}, w(S;\boldsymbol B_1^{(t)}))}{\partial \boldsymbol{B}_1}\right\|^p\right]\\
&+\sum_{k \neq 1} \pi_k  \mathbb{E}_{s \sim \mathcal{POI}\left(\boldsymbol{B}_k^*\right)} \left[r_1(S ; \boldsymbol{B}^{(t)})\left(1-r_1(S ; \boldsymbol{B}^{(t)})\right)\left\|\frac{\partial \log \operatorname{WTPP}({S} \mid \boldsymbol{B}_1^{(t)}, w(S;\boldsymbol B_1^{(t)}))}{\partial \boldsymbol{B}_1}\right\|^p\right] .
\end{aligned}
$$
Let us look at the first term. Define event
$\mathcal{E}_r^{(1)}=\left\{S: S \sim \mathcal{POI}\left(\boldsymbol{B}_1^*\right) ;\left\|\frac{\partial \log \operatorname{WTPP}({S} \mid \boldsymbol{B}_1^*, w(S;\boldsymbol B_1^*))}{\partial \boldsymbol{B}_1}\right\|  \leq r\cdot L\right\}$ for some $r>0$. 
According to the definition in \eqref{eq:robust_fun}, there exists an upper bound $\phi''_{max}<\infty$ of the derivative of $\phi'(\cdot)$. So there exists a constant $\omega_{max}<\infty$ such that 
$\left|w_t(S;\boldsymbol B_1^*)-w_t(S;\boldsymbol B_1^{(t)})\right|<\omega_{max}\cdot a/T$.
Note that
\begin{align}\label{eq:phi_1}
\sum_{i=1}^M \phi'(\int_{t_{i-1}}^{t_i}\lambda_{\boldsymbol{B}_1^{(t)}}(t)dt-1)\leq  c'_{up}\cdot(\int_0^T \lambda_{\boldsymbol{B}_1^*}(t)dt+aT)\leq c'_{up}\cdot(\Omega T):=m_{up},
\end{align}
 where $\phi'(x-1)/x\leq c'_{up}$. 
Similarly, 
\begin{align}\label{eq:phi_2}
\sum_{i=1}^M\left| \phi''(\int_{t_{i-1}}^{t_i}\lambda_{\boldsymbol{B}_1^{(t)}}(t)dt-1)\right|\leq  c''_{up}\cdot(\int_0^T \lambda_{\boldsymbol{B}_1^*}(t)dt+aT)\leq c''_{up}\cdot(\Omega T):=m'_{up},
\end{align}
where $\left|\phi''(x-1)/x\right|\leq c''_{up}$.
Then for $S \in \mathcal{E}_r^{(1)}$, using triangle inequality, we have that
$$
\begin{aligned}
& \left|\sum_{i=1}^M w_i(S;\boldsymbol B_1^{(t)})\cdot\left(\frac{\kappa_h(t_{i-1})}{\lambda_{\boldsymbol{B}_1^{(t)}}(t_{i-1})}-\int_{t_{i-1}}^{t_i}\kappa_h(x)dx\right) \right|\\
\leq & \left|\sum_{i=1}^M w_i(S;\boldsymbol B_1^*)\cdot\left(\frac{\kappa_h(t_{i-1})}{\lambda_{\boldsymbol{B}_1^*}(t_{i-1})}-\int_{t_{i-1}}^{t_i}\kappa_h(x)dx\right)  \right|+\left|\sum_{i=1}^M w_i(S;\boldsymbol B_1^*)\cdot\kappa_h(t_{i-1})\left(\frac{1}{\lambda_{\boldsymbol{B}_1^{(t)}}(t_{i-1})}-\frac{1}{\lambda_{\boldsymbol{B}_1^*}(t_{i-1})}\right)\right|\\
& +\left|\sum_{i=1}^M\left(w_i(S;\boldsymbol B_1^*)-w_i(S;\boldsymbol B_1^{(t)})\right)\cdot\frac{\kappa_h(t_{i-1})}{\lambda_{\boldsymbol{B}_1^{(t)}}(t_{i-1})}\right|+\max_i\left|\left(w_i(S;\boldsymbol B_1^*)-w_i(S;\boldsymbol B_1^{(t)})\right)\right|\cdot L\left|\int_{0}^{T}\kappa_h(x)dx\right|\\
\leq & L\cdot r +\frac{m_{up}}{T\tau^2}\cdot a+\frac{m'_{up}\kappa_{max}}{T\tau}\cdot a +\frac{L\omega_{max}\kappa_{int}}{T}\cdot a := L\cdot r+C_g a,\forall h\in \{1,\cdots,H\},
\end{aligned}
$$
where $\kappa_{int}=\int_{0}^{T}\kappa_h(x)dx$.

Referring to lemma \ref{thm:Estep} and corollary \ref{col:4}, there exists $a_r>0$ such that $\left|r_k(S ; \boldsymbol{B}^{(t)})- r_k^*(S )\right|<a_r $. When $S$ is sampled from cluster $1$, for $S$ within a single period, we have
\begin{align*}
 \left\|W(S;\boldsymbol B^{(t)})- w(S;\boldsymbol B_{1}^*)\right\|_1&\leq \sum_{k\in [K]} \left\|r_k(S ; \boldsymbol{B}^{(t)})w(S;\boldsymbol B_k^{(t)})-r_k^*(S )w(S;\boldsymbol B_k^*)\right\|_1  \\
 &\leq \sum_{k\in [K]} r_k(S ; \boldsymbol{B}^{(t)})\left\|w(S;\boldsymbol B_k^{(t)})-w(S;\boldsymbol B_k^*)\right\|_1+  \left|r_k(S ; \boldsymbol{B}^{(t)})-r_k^*(S )\right|\left\|w(S;\boldsymbol B_{k}^*)\right\|_1 \\
 &\leq \sum_{k\in [K]} r_k(S ; \boldsymbol{B}^{(t)})\left\|\frac{\partial w(S;\boldsymbol B_k^{(t)})}{\partial \boldsymbol B_k}\right\|_1\cdot\left\|\boldsymbol B_k^{(t)}-\boldsymbol B_k^*\right\|_\infty+  a_r\left\|w(S;\boldsymbol B_{k}^*)\right\|_1 \\
 &\leq \max_{k\in [K]} \left\|\frac{\partial w(S;\boldsymbol B_k^{(t)})}{\partial \boldsymbol B_k}\right\|\cdot a/(T\cdot\kappa_{\max})+  a_r K\left\|w(S;\boldsymbol B_{k}^*)\right\|_1 \\
 &\leq m'_{up} a/(T\cdot\kappa_{\max})+  a_r K m_{up}:=C_W(a,a_r),
\end{align*}
where $m'_{up}, m_{up}$ are constants that depend only on $\rho$. Then we have 
\begin{align*}
&\log  \operatorname{WTPP}(S\mid \boldsymbol{B}_1^{(t)}, W(S;\boldsymbol B^{(t)}))=\sum_{i=1}^M W_i(S;\boldsymbol B^{(t)})\cdot\left(\log\lambda_{\boldsymbol{B}_1^{(t)}}(s_i)-\int_{s_{i-1}}^{s_i}\lambda_{\boldsymbol{B}_1^{(t)}}(s)ds\right)\\
=&\log \operatorname{WTPP}(S\mid \boldsymbol{B}_1^*, w(S;\boldsymbol B_{1}^*))+\sum_{i=1}^M W_i(S;\boldsymbol B^{(t)})\cdot\left(\log(\lambda_{\boldsymbol{B}_1^{(t)}}(s_i)/\lambda_{\boldsymbol{B}_1^*}(s_i))-\int_{s_{i-1}}^{s_i}(\lambda_{\boldsymbol{B}_1^{(t)}}(s)-\lambda_{\boldsymbol{B}_1^*}(s))ds\right)\\
&+\sum_{i=1}^M (W_i(S;\boldsymbol B^{(t)})-w_i(S;\boldsymbol B_{1}^*))\cdot\left(\log\lambda_{\boldsymbol{B}_1^*}(s_i)-\int_{s_{i-1}}^{s_i}\lambda_{\boldsymbol{B}_1^*}(s)ds\right)\\
\geq& \log \operatorname{WTPP}(S\mid \boldsymbol{B}_1^*, w(S;\boldsymbol B_1^*)) -(m_{up}\log\left(\frac{\tau+a/T}{\tau}\right)+a)\cdot L -C_W(a,a_r) \cdot ( \Omega T -\log \tau ) L.
\end{align*}

For $k\neq1$ we have $\log \operatorname{WTPP}(S\mid \boldsymbol{B}_k^{(t)}, W(S;\boldsymbol B^{(t)}))-\log \operatorname{WTPP}(S\mid \boldsymbol{B}_k^*, w(S;\boldsymbol B_{1}^*))\leq (m_{up}\log\left((\tau+a/T)/\tau\right)+a)\cdot L +C_W(a,a_r) \cdot( \log \Omega-\tau T  ) L$.

By assumption \ref{asm:4} and lemma \ref{thm:Estep}, we know that $\log \operatorname{WTPP}(S\mid \boldsymbol{B}_k^{(t)}, W(S;\boldsymbol B^{(t)}))\leq\log \operatorname{WTPP}(S\mid \boldsymbol{B}_1^*, w(S;\boldsymbol B_{1}^*)) -C_\eta\cdot L+(m_{up}\log\left(\frac{\tau+a/T}{\tau}\right)+a)\cdot L+C_W(a,a_r) \cdot( \log \Omega-\tau T  ) L$. So we get that 
\begin{align*}
&\mathbb{E}_S\left[(1-r_1(S; \boldsymbol{B}^{(t)}))\left\|\frac{\partial \log \operatorname{WTPP}({S} \mid \boldsymbol{B}_1^{(t)},W)}{\partial \boldsymbol{B}_1}\right\|^p| \mathcal{E}_r^{(1)}\right]\\
&\leq \frac{1-\pi_1}{\pi_1}\frac{ \operatorname{WTPP}(S\mid \boldsymbol{B}_k^{(t)},W)}{ \operatorname{WTPP}(S\mid \boldsymbol{B}_1^{(t)},W)}\left\|\frac{\partial \log \operatorname{WTPP}({S} \mid \boldsymbol{B}_1^{(t)},W)}{\partial \boldsymbol{B}_1}\right\|^p\\
&\leq  \frac{1-\pi_1}{\pi_1}\exp{\left(-C_\eta\cdot L +2 (m_{up}\log\left(\frac{\tau+a/T}{\tau}\right)+a)\cdot L +C_W(a,a_r) \cdot( \log \Omega/\tau +(\Omega-\tau) T  ) L \right)}\\
&\cdot\left(rL+C_g a\right).
\end{align*}

For $ \mathcal{E}_r^{c}$ parts, we refer to Theorem \ref{Thm:Mout}. In the case of $\eta$ proportional outliers, we have
\begin{align*}
    \mathbb{P}\left(\left\|\frac{\partial \log \operatorname{WTPP}({S} \mid \boldsymbol{B}_1^{(t)},W)}{\partial \boldsymbol{B}_1}/L\right\|\geq t+\eta \cdot C_{\rho}\right)\leq 2\exp{\left(-\frac{t L}{c_1}\right)}.
\end{align*}
Taking $ r>2\eta \cdot C_{\rho}$ and $c_0=2c_1$, for $ t\geq r$,
\begin{align*}
    \mathbb{P}\left(\left\|\frac{\partial \log \operatorname{WTPP}({S} \mid \boldsymbol{B}_1^{(t)},W)}{\partial \boldsymbol{B}_1}/L\right\|\geq t\right)\leq 2\exp{\left(-\frac{t L}{c_0}\right)}.
\end{align*}

Obviously, $r_1(S ; \boldsymbol{B}^{(t)})\left(1-r_1(S ; \boldsymbol{B}^{(t)})\right)\leq 1/4$. Then
\begin{align*}
&\mathbb{E}_S\left[ r_1(S ; \boldsymbol{B}^{(t)})\left(1-r_1(S ; \boldsymbol{B}^{(t)})\right)\left\|\frac{\partial \log \operatorname{WTPP}({S} \mid \boldsymbol{B}_1^{(t)},W)}{\partial \boldsymbol{B}_1}\right\|^p \mid \mathcal{E}_r^{c}\right]\\
&\leq \frac{1}{4}\int_{r}^\infty t^p d\mathbb{P}\left(\left\|\frac{\partial \log \operatorname{WTPP}({S} \mid \boldsymbol{B}_1^{(t)},W)}{\partial \boldsymbol{B}_1}\right\|\geq t\cdot L\right)\\
&= \frac{1}{4}\left(r^p\cdot L\mathbb{P}\left(\left\|\frac{\partial \log \operatorname{WTPP}({S} \mid \boldsymbol{B}_1^{(t)},W)}{\partial \boldsymbol{B}_1}\right\|\geq r\cdot L\right)+\int_{r}^\infty pt^{p-1} \mathbb{P}\left(\left\|\frac{\partial \log \operatorname{WTPP}({S} \mid \boldsymbol{B}_1^{(t)},W)}{\partial \boldsymbol{B}_1}\right\|\geq t\cdot L\right)dt\right) \\
&\leq \frac{1}{2}\left(r^p L\exp{\left(-\frac{r L}{c_0}\right)}+\int_{r}^\infty pt^{p-1}\exp{\left(-\frac{t L}{c_0}\right)}dt\right).
\end{align*}

For fixed $r> 0$, when $L\rightarrow\infty$, it's easy to know that $r^p L\exp{\left(-r L/c_0\right)}+\int_{r}^\infty pt^{p-1}\exp{\left(-t L/c_0\right)}dt\rightarrow 0 $. 

Next we consider the remainder of the gradient. For $k\neq 1$,
\begin{align*}
 &\pi_k  \mathbb{E}_{s \sim \mathcal{POI}\left(\boldsymbol{B}_k^*\right)}\left[ r_1(S ; \boldsymbol{B}^{(t)})\left\|\frac{\partial \log \operatorname{WTPP}({S} \mid \boldsymbol{B}_1^{(t)},W)}{\partial \boldsymbol{B}_1}\right\|^p\right]  \\
 &=\underbrace{\int_{\left\|\frac{\partial \log \operatorname{WTPP}(\mathbf{S} \mid \boldsymbol{B}_k^*,w(S;\boldsymbol B_k^*))}{\partial \boldsymbol{B}_k}\right\|<r\cdot L}\frac{\pi_1 \operatorname{WTPP}(S\mid \boldsymbol{B}_1^{(t)},W)\pi_k \operatorname{WTPP}(S\mid \boldsymbol{B}_k^*,W)}{\sum_j\pi_j \operatorname{WTPP}(S\mid \boldsymbol{B}_j^{(t)},W)}\left\|\frac{\partial \log \operatorname{WTPP}({S} \mid \boldsymbol{B}_1^{(t)},W)}{\partial \boldsymbol{B}_1}\right\|^p dS}_{I_1} \\
 &+\underbrace{\int_{\left\|\frac{\partial \log \operatorname{WTPP}(\mathbf{S} \mid \boldsymbol{B}_k^*,w(S;\boldsymbol B_k^*))}{\partial \boldsymbol{B}_k}\right\|>r\cdot L}\frac{\pi_1 \operatorname{WTPP}(S\mid \boldsymbol{B}_1^{(t)},W)\pi_i \operatorname{WTPP}(S\mid \boldsymbol{B}_k^*,W)}{\sum_j\pi_j \operatorname{WTPP}(S\mid \boldsymbol{B}_j^{(t)},W)}\left\|\frac{\partial \log \operatorname{WTPP}({S} \mid \boldsymbol{B}_1^{(t)},W)}{\partial \boldsymbol{B}_1}\right\|^p dS}_{I_2}.
\end{align*}

When $\left\|\partial \log \operatorname{WTPP}({S} \mid \boldsymbol{B}_k^*,w(S;\boldsymbol B_k^*))/\partial \boldsymbol{B}_k\right\|<r\cdot L $, we get that $ \operatorname{WTPP}(S\mid \boldsymbol{B}_k^{(t)},W)/\operatorname{WTPP}(S\mid \boldsymbol{B}_k^*,W)\leq \exp{\left((m_{up}\log\left((\tau+a/T)/\tau\right)+a)\cdot L  +C_W(a,a_r) \cdot( \log \Omega-\tau T  ) L \right)}$ and 
$\operatorname{WTPP}(S\mid \boldsymbol{B}_k^*,W)/\operatorname{WTPP}(S\mid \boldsymbol{B}_k^{(t)},W)\leq \exp{\left((m_{up}\log\left((\tau+a/T)/\tau\right)+a)\cdot L +C_W(a,a_r) \cdot( -\log \tau+\Omega T  ) L \right)}$. Then
\begin{align*}
    I_1&\leq \frac{\pi_k \operatorname{WTPP}(S\mid \boldsymbol{B}_k^*,W)}{\pi_i \operatorname{WTPP}(S\mid \boldsymbol{B}_k^{(t)},W)}\cdot\int_{\left\|\frac{\partial \log \operatorname{WTPP}(\mathbf{S} \mid \boldsymbol{B}_k^*,w)}{\partial \boldsymbol{B}_k}\right\|<r\cdot L }\pi_1 \operatorname{WTPP}(S\mid \boldsymbol{B}_1^{(t)},W)\left\|\frac{\partial \log \operatorname{WTPP}({S} \mid \boldsymbol{B}_1^{(t)},W)}{\partial \boldsymbol{B}_1}\right\|^p dS\\
    &\leq\pi_1\exp{\left((m_{up}\log\left(\frac{\tau+a/T}{\tau}\right)+a)\cdot L  +C_W(a,a_r) \cdot( -\log \tau+\Omega T  ) L \right)}\\
    &\cdot\int_{\left\|\frac{\partial \log \operatorname{WTPP}(\mathbf{S} \mid \boldsymbol{B}_k^*,w)}{\partial \boldsymbol{B}_k}\right\|<r\cdot L } \operatorname{WTPP}(S\mid \boldsymbol{B}_k^*,W)(C_0 L )^pdS\\
    &\cdot\exp{\left(-C_\eta L  +(m_{up}\log\left(\frac{\tau+a/T}{\tau}\right)+a)\cdot L  +C_W(a,a_r) \cdot( \log \Omega-\tau T  ) L \right)}\\
    &\leq \pi_1\exp{\left(-C_\eta\cdot L +2 (m_{up}\log\left(\frac{\tau+a/T}{\tau}\right)+a)\cdot L +C_W(a,a_r) \cdot( \log \Omega/\tau +(\Omega-\tau) T  ) L \right)}\cdot(C_0 L )^p ,
\end{align*}
where $C_0$ is the upper bound of $\left\|\partial \log \operatorname{WTPP}({S} \mid \boldsymbol{B}_k^{(t)},W)/\partial \boldsymbol{B}_k\right\|$ with a high probability $1-\delta$, $\forall k=1,\cdots,K$.

When $\left\|\partial \log \operatorname{WTPP}(\mathbf{S} \mid \boldsymbol{B}_k^*,w(S;\boldsymbol B_k^*))/\partial \boldsymbol{B}_k\right\|>r\cdot L $,
\begin{align*}
    I_2 &= \frac{\pi_1 \operatorname{WTPP}(S\mid \boldsymbol{B}_1^{(t)},W)}{\sum_{j=1}^K\pi_j \operatorname{WTPP}(S\mid \boldsymbol{B}_j^{(t)},W)}\\
    &\cdot\int_{\left\|\frac{\partial \log \operatorname{WTPP}(\mathbf{S} \mid \boldsymbol{B}_k^*,W)}{\partial \boldsymbol{B}_k}\right\|>r\cdot L }\pi_k \operatorname{WTPP}(S\mid \boldsymbol{B}_k^*,W)\left\|\frac{\partial \log \operatorname{WTPP}({S} \mid \boldsymbol{B}_1^{(t)},W)}{\partial \boldsymbol{B}_1^{}}\right\|^p dS\\
    &\leq \int_{\left\|\frac{\partial \log \operatorname{WTPP}(\mathbf{S} \mid \boldsymbol{B}_k^*,w(S;\boldsymbol B_k^*))}{\partial \boldsymbol{B}_k}\right\|>r\cdot L }\pi_k \operatorname{WTPP}(S\mid \boldsymbol{B}_k^*,W)\left\|\frac{\partial \log \operatorname{WTPP}({S} \mid \boldsymbol{B}_1^{(t)},W)}{\partial \boldsymbol{B}_1^{}}\right\|^p dS\\
    &\leq \pi_k(C_0 L )^p\int_{\left\|\frac{\partial \log \operatorname{WTPP}(\mathbf{S} \mid \boldsymbol{B}_k^*,w(S;\boldsymbol B_k^*))}{\partial \boldsymbol{B}_k}\right\|>r\cdot L } \operatorname{WTPP}(S\mid \boldsymbol{B}_k^*,W)dS\\
    &\leq 2\pi_k(C_0 L )^p\exp{\left(-\frac{t L }{c_0}\right)}dS,
\end{align*}
where we use the conclusion obtained above that $\mathbb{P}\left(\left\|\partial \log \operatorname{WTPP}({S} \mid \boldsymbol{B}_k^*,w(S;\boldsymbol B_k^*))/\partial \boldsymbol{B}_k\right\|/L \geq t\right)\leq 2\exp{\left(-t L/c_0\right)}$. Takeing
\begin{align}\label{eq:G}
G=\min\{C_\eta\cdot L -2 (m_{up}\log\left(\frac{\tau+a/T}{\tau}\right)+a)-C_W(a,a_r) \cdot( \log \Omega/\tau +(\Omega-\tau) T ),t/c_0\},
\end{align}
 and then we can get the result.

\begin{lemma}\label{lem:cluster}
    If $\|\boldsymbol{B}_k^{(t)}-\boldsymbol{B}_k^*\| <a/(T\cdot\kappa_{\max})$, that is $\left|\lambda_k(t)-\lambda_k^*(t)\right|< a/T$ for $k=1,2,\dots,K$, then for $p=1,2$ and $\forall k\in [K]$, there exist a constant $G>0$ such that 
    $$\left\|\nabla r_k(S,\boldsymbol{B}^{(t)})\right\|= O(\sqrt{H} L\exp(-G\cdot L)).$$
\end{lemma}
\noindent \textbf{Proof of Lemma \ref{lem:cluster}}
Without loss of generality, we prove the claim for $k=1$. Recall the definition of $r_1(S ; \boldsymbol{B}^{(t)})$ , for any given $S$, consider the function $\boldsymbol{B}\rightarrow r_1(S ; \boldsymbol{B})$, it's easy to know that 
\begin{align*}
\nabla r_1(S;\boldsymbol{B}^{(t)})=\left(\begin{array}{c}
-r_1(S ; \boldsymbol{B}^{(t)})\left(1-r_1(S ; \boldsymbol{B}^{(t)})\right)\frac{\partial \log \operatorname{WTPP}({S} \mid \boldsymbol{B}_1^{(t)},W)}{\partial \boldsymbol{B}_1,W}\\
r_1(S ; \boldsymbol{B}^{(t)})r_2(S ; \boldsymbol{B}^{(t)})\frac{\partial \log \operatorname{WTPP}({S} \mid \boldsymbol{B}_2^{(t)},W)}{\partial \boldsymbol{B}_2^{ }}\\
\vdots\\ 
r_1(S ; \boldsymbol{B}^{(t)})r_K(S ; \boldsymbol{B}^{(t)})\frac{\partial \log \operatorname{WTPP}({S} \mid \boldsymbol{B}_K^{(t)},W)}{\partial \boldsymbol{B}_K}\\ 
\end{array}\right),
\end{align*}
where 
\begin{align*}
\frac{\partial \log \operatorname{WTPP}({S} \mid \boldsymbol{B}_k^{(t)},W)}{\partial \boldsymbol{B}_k}=\left(\begin{array}{c}
\sum_{j=1}^{M}W_j\cdot(\frac{\kappa_1(s_{j-1})}{\lambda_{\boldsymbol{B}_k^{(t)}}(s_{j-1})}-\int_{s_{j-1}}^{s_j}\kappa_1(x)dx)\\
\vdots\\
\sum_{j=1}^{M}W_j\cdot(\frac{\kappa_H(s_{j-1})}{\lambda_{\boldsymbol{B}_k^{(t)}}(s_{j-1})}-\int_{s_{j-1}}^{s_j}\kappa_H(x)dx)\\
\end{array}\right)^\top.
\end{align*}

To calculate the upper bound of $\|\nabla r_k(S,\boldsymbol{B}^{(t)})\|$, let us start by considering the first line. 
Referring to the lemma \ref{lem:weight}, it is easy to know that the first line achieve the requirement. Then let us consider others line. Noticed that 
{\small\begin{align*}
& ~~~~~\mathbb{E}_S\left[ r_1(S ; \boldsymbol{B}^{(t)})r_k(S ; \boldsymbol{B}^{(t)})\left\|\frac{\partial \log \operatorname{WTPP}({S} \mid \boldsymbol{B}_k^{(t)},W)}{\partial \boldsymbol{B}_k}\right\|\right] \\
&\leq \mathbb{E}_S\left[ r_k(S ; \boldsymbol{B}^{(t)})\left(1-r_k(S ; \boldsymbol{B}^{(t)})\right)\left\|\frac{\partial \log \operatorname{WTPP}({S} \mid \boldsymbol{B}_k^{(t)},W)}{\partial \boldsymbol{B}_k}\right\|\right],
\end{align*}}
$\forall k \neq 1$. Thus, the upper bound of line $k$ is the same as line $1$. The proof is complete.

We define $g(\boldsymbol{B}_k^{(t)}\mid \boldsymbol{B}_k^{(t)}):= \mathbb{E}_S 
( r_k(S ; \boldsymbol{B}^{(t)})\cdot\sum_{k=1}^K r_k(S ; \boldsymbol{B}^{(t)}) \nabla\log \operatorname{WTPP}(S\mid \boldsymbol{B}_k^{(t)},  w(S;\boldsymbol B_k^{(t)}) ) /L )$, and 
$g(\boldsymbol{B}_k^{(t)}\mid \boldsymbol{B}_k^*):= \mathbb{E}_S 
( r_k(S ; \boldsymbol{B}^*)\cdot \sum_{k=1}^K r_k(S ; \boldsymbol{B}^*) \nabla\log \operatorname{WTPP}(S\mid \boldsymbol{B}_k^{(t)},  w(S;\boldsymbol B_k^{(t)}) ) /L ) $.

\begin{lemma}
    \label{thm:gamma}
    For $k=\{1,2,\cdots,K\}$, 
    $\|\boldsymbol{B}_k^{(t)}-\boldsymbol{B}_k^*\| <a/(T\cdot\kappa_{\max})$, and then we can get that $\|g(\boldsymbol{B}_k^{(t)}\mid \boldsymbol{B}_k^{(t)})-g(\boldsymbol{B}_k^{(t)}\mid \boldsymbol{B}_k^*)\|\leq \gamma\|\boldsymbol{B}_k^{(t)}-\boldsymbol{B}_k^*\| $, where $\gamma=O(\sqrt{H} L\cdot\exp(-G\cdot L)).$ 
\end{lemma}

\noindent\textbf{Proof of lemma \ref{thm:gamma}}
Without loss of generality, we only consider $k=1$. When $S$ is sampled from cluster $1$,
\begin{align*}
&\left\|g(\boldsymbol{B}_1^{(t)}\mid \boldsymbol{B}_1^{(t)})-g(\boldsymbol{B}_1^{(t)}\mid \boldsymbol{B}_1^*)\right\|\\
=&\Big\|\mathbb{E}_S 
\Big[  (r_1\left(S ; \boldsymbol{B}^{(t)}\right) \sum_j r_j(S ; \boldsymbol{B}^{(t)}) w(S;\boldsymbol{B}^{(t)}_j)-r_1\left(S ; \boldsymbol{B}^*\right) \sum_j r_j(S ; \boldsymbol{B}^{(t)}) w(S;\boldsymbol{B}^{(t)}_j)) \\
& \cdot (\frac{\kappa_h(s_{j-1})}{\lambda_{\boldsymbol{B}_1^{(t)}}(s_{j-1})}-\int_{s_{j-1}}^{s_j}\kappa_h(x)dx ) \Big]\Big\| \\
\leq&\underbrace{\left\|\mathbb{E}_S 
\left[ (r_1\left(S ; \boldsymbol{B}^{(t)}\right)r_1\left(S ; \boldsymbol{B}^{(t)}\right) -r_1\left(S ; \boldsymbol{B}^*\right) r_1\left(S ; \boldsymbol{B}^*\right))\cdot w_j(S;\boldsymbol{B}_k^{(t)}) (\frac{\kappa_h(s_{j-1})}{\lambda_{\boldsymbol{B}_1^{(t)}}(s_{j-1})}-\int_{s_{j-1}}^{s_j}\kappa_h(x)dx ) \right]\right\|}_{I_1} \\
+&\underbrace{\left\|\mathbb{E}_S 
\left[ (r_1\left(S ; \boldsymbol{B}^{(t)}\right)\sum_{j\neq 1} r_{j}\left(S ; \boldsymbol{B}^{(t)}\right) -r_1\left(S ; \boldsymbol{B}^*\right) \sum_{j\neq 1} r_{j}\left(S ; \boldsymbol{B}^*\right))\cdot w_j(S;\boldsymbol{B}_k^{(t)})(\frac{\kappa_h(s_{j-1})}{\lambda_{\boldsymbol{B}_1^{(t)}}(s_{j-1})}-\int_{s_{j-1}}^{s_j}\kappa_h(x)dx )\right]\right\|}_{I_2}.
\end{align*}

Referring to Lemma \ref{lem:cluster}, we are easy to know that $I_1, I_2\sim O(\sqrt{H} L\cdot\exp(-G\cdot L))$, and $I_1,I_2\rightarrow0$ as $L\rightarrow\infty$.

{
\begin{lemma}\label{lem:grad:concen}
    For cluster $k$, we write 
    {\small
    \begin{align*}
        &g(\boldsymbol{B}_k^{(t)}\mid \boldsymbol{B}_k^{(t)})_{\boldsymbol{S},h}:=\frac{1}{N}\sum_{n=1}^N r_k(S_n;\mathbf{B}^{(t)})\sum_{j}  W_j(S_n;\boldsymbol{B}^{(t)})\cdot(\frac{\kappa_h(S_{n,{j-1}})}{\lambda_{\boldsymbol{B}_k^{(t)}}(S_{n,j-1})}-\int_{S_{n,j-1}}^{S_{n,j}}\kappa_h(x)dx )/L(S_n),\\
        & g(\boldsymbol{B}_k^{(t)}\mid \boldsymbol{B}_k^{(t)})= \mathbb{E}_S( r_k(S ; \boldsymbol{B}^{(t)})\cdot \nabla\log \operatorname{WTPP}(S\mid \boldsymbol{B}_k^{(t)},  W(S;\boldsymbol B_k^{(t)}) )) /L(S) ),
    \end{align*}
    }
    where $g(\boldsymbol{B}_k^{(t)}\mid \boldsymbol{B}_k^{(t)})_{\boldsymbol{S},h} $ is the $h-th$ element of $g(\boldsymbol{B}_k^{(t)}\mid \boldsymbol{B}_k^{(t)})_{\boldsymbol{S}} $.
    Then we have $\left\|g(\boldsymbol{B}_k^{(t)}\mid \boldsymbol{B}_k^{(t)})_{\boldsymbol{S}}-g(\boldsymbol{B}_k^{(t)}\mid \boldsymbol{B}_k^{(t)})\right\|=O(\sqrt{H}\cdot(1/\sqrt{NL}+ \eta)):= \epsilon^{unif}$.
    
\end{lemma}
}

\noindent\textbf{Proof of Lemma \ref{lem:grad:concen}}~
For each sample $S_n$, we assume that the sample unaffected by outliers is $\widetilde{S_n}$, and we write $\widetilde{\boldsymbol{S}}:=\{\widetilde{S_1},\cdots \}$.
By triangle inequality, we have 
\begin{align*}
    &\left\|g(\boldsymbol{B}_k^{(t)}\mid \boldsymbol{B}_k^{(t)})_{\boldsymbol{S}}-g(\boldsymbol{B}_k^{(t)}\mid \boldsymbol{B}_k^{(t)})\right\|\\
    &\leq {\left\|g(\boldsymbol{B}_k^{(t)}\mid \boldsymbol{B}_k^{(t)})_{\boldsymbol{S}}-g(\boldsymbol{B}_k^{(t)}\mid \boldsymbol{B}_k^{(t)})_{\widetilde{\boldsymbol{S}}}\right\|}+{\left\|g(\boldsymbol{B}_k^{(t)}\mid \boldsymbol{B}_k^{(t)})_{\widetilde{\boldsymbol{S}}}-g(\boldsymbol{B}_k^{(t)}\mid \boldsymbol{B}_k^{(t)})\right\|}\\
    & = O(\sqrt{H}C_\rho \eta) + O(\sqrt{H}/\sqrt{NL}),
\end{align*}
where $g(\boldsymbol{B}_k^{(t)}\mid \boldsymbol{B}_k^{(t)})_{\widetilde{\boldsymbol{S}}}$ is the gradient of samples $\widetilde{\boldsymbol{S}}$.
Similar to Theorem \ref{Thm:Mout}, we get that the order of the first term is $ O(\sqrt{H}C_\rho \eta)$. We use the fact that each period of each sample in the same cluster is independently and identically distributed, and then we can know that the order of the second term is $O(\sqrt{H}/\sqrt{NL}) $.

\noindent\textbf{Proof of Theorem \ref{thm:local}}~

We only need to prove the situation when $k=1$.
Recall the update rule and definition of $g(\boldsymbol{B}_1^{(t)}\mid \boldsymbol{B}_1^{(t)})$. We know that 
\[\boldsymbol{B}_1^{(t+1)} = 
\boldsymbol{B}_1^{(t)} - lr \cdot \varrho_1^{(t)} = \boldsymbol{B}_1^{(t)} - lr \cdot g(\boldsymbol{B}_1^{(t)}\mid \boldsymbol{B}_1^{(t)})_{\boldsymbol{S}},\]
where $H(S,\boldsymbol{B}^{(t)}_1)$ is the Hessian matrix.
Let $\overline{\boldsymbol{B}}_1:=\mathop{\arg\min}_{\|g(\boldsymbol{\overline B}_1\mid \boldsymbol{\overline B}_1)\|=0} \|\overline{\boldsymbol{B}}_1-{\boldsymbol{B}_1^*}\|$. Then we get that $ \left\|\overline{\boldsymbol{B}}_1-\boldsymbol{B}_1^*\right\|=O(\sqrt{H}/L)$ according to lemma \ref{lem:B_bar}.
By triangle inequality and Lemma \ref{thm:gamma}, we have
$$
\begin{aligned}
&\left\|\boldsymbol{B}_1^{(t+1)}-\boldsymbol{B}_1^*\right\|  =\left\|\boldsymbol{B}_1^{(t)}-\boldsymbol{B}_1^*- lr \cdot  g(\boldsymbol{B}_1^{(t)}\mid \boldsymbol{B}_1^{(t)})_{\boldsymbol{S}}\right\| \\
& \leq\left\|\boldsymbol{B}_1^{(t)}-\overline{\boldsymbol{B}}_1- lr \cdot g(\boldsymbol{B}_1^{(t)}\mid \boldsymbol{B}_1^*)\right\|   +\left\|\overline{\boldsymbol{B}}_1-\boldsymbol{B}_1^*\right\|\\
& +   \left\|lr\cdot(g(\boldsymbol{B}_1^{(t)}\mid \boldsymbol{B}_1^{(t)})-g(\boldsymbol{B}_1^{(t)}\mid \boldsymbol{B}_1^*) )\right\| + \left\|lr\cdot(g(\boldsymbol{B}_1^{(t)}\mid \boldsymbol{B}_1^{(t)})-g(\boldsymbol{B}_1^{(t)}\mid \boldsymbol{B}_1^{(t)})_{\boldsymbol{S}} )\right\| \\
& \leq \frac{\lambda_{max}-\lambda_{min}}{\lambda_{max}+\lambda_{min}}\left\|\boldsymbol{B}_1^{(t)}-\overline{\boldsymbol{B}}_1\right\|+\left\|\overline{\boldsymbol{B}}_1-\boldsymbol{B}_1^*\right\|+\gamma\left\|\boldsymbol{B}_1^{(t)}-\boldsymbol{B}_1^*\right\|+\epsilon^{unif}\\
& \leq \frac{\lambda_{max}-\lambda_{min}}{\lambda_{max}+\lambda_{min}}\left\|\boldsymbol{B}_1^{(t)}-\boldsymbol{B}_1^*\right\|+\left(\frac{\lambda_{max}-\lambda_{min}}{\lambda_{max}+\lambda_{min}}+1\right)\cdot\left\|\overline{\boldsymbol{B}}_1-\boldsymbol{B}_1^*\right\|+\gamma\left\|\boldsymbol{B}_1^{(t)}-\boldsymbol{B}_1^*\right\|+\epsilon^{unif},
\end{aligned}
$$
where $lr=2/(\lambda_{max}+\lambda_{min})$. To see why the second inequality hold, notice that $\nabla g(\boldsymbol{B}_1^{(t)}\mid \boldsymbol{B}_1^*)$ and its neighborhood has the largest eigenvalue $-\lambda_{\min }$ and the smallest eigenvalue $-\lambda_{\max }$. Apply the classical result for gradient descent, with step size $lr=2/(\lambda_{\max }+\lambda_{\min })$ guarantees (see \citet{nesterov2003introductory})
$$
\left\|\boldsymbol{B}_1^{(t)}-\overline{\boldsymbol{B}}_1-lr\cdot g(\boldsymbol{B}_1^{(t)}\mid \boldsymbol{B}_1^*)\right\| \leq \frac{\lambda_{\max }-\lambda_{\min }}{\lambda_{\max }+\lambda_{\min }}\left\|\boldsymbol{B}_1^{(t)}-\overline{\boldsymbol{B}}_1\right\|.
$$

\section{Proof of Theorem \ref{thm:tpr}}
\label{sec:pf3}

\noindent\textbf{Proof of Proposition \ref{thm:tprfnr}}~

For any $\widetilde\alpha\in (0,1)$, there exist $0<x_{\widetilde\alpha}^1<1<x_{\widetilde\alpha}^2\leq\infty$ such that $\phi'(x_{\widetilde\alpha}^1)=\phi'(x_{\widetilde\alpha}^2)=\widetilde\alpha $, and $x_{\widetilde\alpha}^1\exp(-x_{\widetilde\alpha}^1)=x_{\widetilde\alpha}^2\exp(-x_{\widetilde\alpha}^2) $.
For $\lambda\in (0,+\infty)$, we define 
$$
f(\lambda):=\mathbb{E}(X\mathbf{I}_{\{\phi'(X/\lambda-1)>\widetilde\alpha\}})=\int_{\lambda x_{\widetilde\alpha}^1}^{\lambda x_{\widetilde\alpha}^2}x\exp(-x)dx=\exp(-\lambda x_{\widetilde\alpha}^1)-\exp(-\lambda x_{\widetilde\alpha}^2),
$$
where $X\sim \text{Exp}(1)$. 

Referring to Lemma \ref{lem:exp1}, there are outliers when $\lambda\neq 1$. 
$f'(\lambda)=-x_{\widetilde\alpha}^1\exp(-\lambda x_{\widetilde\alpha}^1)+x_{\widetilde\alpha}^2\exp(-\lambda x_{\widetilde\alpha}^2)$. Because of $x_{\widetilde\alpha}^1\exp(-x_{\widetilde\alpha}^1)=x_{\widetilde\alpha}^2\exp(-x_{\widetilde\alpha}^2)$, we know that
\begin{align*}
f'(\lambda)&=(x_{\widetilde\alpha}^2)^{1-\lambda}\cdot(x_{\widetilde\alpha}^2)^\lambda\exp(-\lambda x_{\widetilde\alpha}^2)-(x_{\widetilde\alpha}^1)^{1-\lambda}\cdot(x_{\widetilde\alpha}^1)^\lambda\exp(-\lambda x_{\widetilde\alpha}^1)\\
&=((x_{\widetilde\alpha}^2)^{1-\lambda}-(x_{\widetilde\alpha}^1)^{1-\lambda})\cdot (x_{\widetilde\alpha}^2)^\lambda\exp(-\lambda x_{\widetilde\alpha}^2).
\end{align*}
We are easy to know that $ f'(\lambda)>0$ when $\lambda\in(0,1)$ and $ f'(\lambda)<0$ when $\lambda>1$. That is, $f(\lambda)<f(1)$ for all $\lambda\neq 1$. In particular, $f(\lambda)\rightarrow 0$ as $ \lambda\rightarrow 0$ or $ \lambda\rightarrow\infty$. 

For arbitrary intensity $\lambda'$ that satisfies $ \lambda'(t) < \lambda^{\ast}(t)$ for all $t \in [0,T_0]$ or $ \lambda'(t) > \lambda^{\ast}(t)$ for all $t \in [0,T_0]$, assume that $T_{out}:=\{t\in [0,T_0], \lambda'(t)\neq \lambda^{\ast}(t)\}$, then we have 
\begin{align*}
& \text{IW}(\lambda'; \lambda^{\ast})=\mathbb{E}_{S \sim \lambda' } \left[\sum_{i=1}^M \mathbf{I}_{\{\phi'\left( \int_{t_{i-1}}^{t_i} \lambda^{\ast}(u) d u-1\right)>\widetilde\alpha\}}\cdot(t_i-t_{i-1})\right]\\
&=\mathbb{E}_{S \sim \lambda' } \left[\sum_{i : t_i\in T_{out}} \mathbf{I}_{\{\phi'\left( \int_{t_{i-1}}^{t_i} \lambda^{\ast}(u) d u-1\right)>\widetilde\alpha\}}\cdot(t_i-t_{i-1})\right] + \mathbb{E}_{S \sim \lambda' } \left[\sum_{i : t_i\notin T_{out}} \mathbf{I}_{\{\phi'\left( \int_{t_{i-1}}^{t_i} \lambda^{\ast}(u) d u-1\right)>\widetilde\alpha\}}\cdot(t_i-t_{i-1})\right]\\
&< \mathbb{E}_{S \sim \lambda^{\ast} } \left[\sum_{i : t_i\in T_{out}} \mathbf{I}_{\{\phi'\left( \int_{t_{i-1}}^{t_i} \lambda^{\ast}(u) d u-1\right)>\widetilde\alpha\}}\cdot(t_i-t_{i-1})\right] + \mathbb{E}_{S \sim \lambda^{\ast} } \left[\sum_{i : t_i\notin T_{out}} \mathbf{I}_{\{\phi'\left( \int_{t_{i-1}}^{t_i} \lambda^{\ast}(u) d u-1\right)>\widetilde\alpha\}}\cdot(t_i-t_{i-1})\right]\\
&= \text{IW}(\lambda^{\ast}; \lambda^{\ast}).
\end{align*}

\noindent\textbf{Proof of Theorem \ref{thm:tpr} }~

Recalling the definition \eqref{eq:TPR}, when $\hat{\boldsymbol{B}} \equiv \boldsymbol{B}^{\ast}$,
it can be computed that 
\begin{align}\label{eq:tprnow}
\mathbb{E}[TPR(n)] &=\mathbb{E}\left[ \sum_{i : t_i\in T_{out}} \mathbf{I}_{\{\phi'\left( \int_{t_{i-1}}^{t_i} \lambda^{\ast}(u) d u-1\right)>\widetilde\alpha\}}\cdot(t_i-t_{i-1})\right]/T_{out}\\
&=\mathbb{E}\Big[ \sum_{i : t_i\in T} \mathbf{I}_{\{\phi'\left( \int_{t_{i-1}}^{t_i} \lambda^{\ast}(u) d u-1\right)>\widetilde\alpha\}}(t_i-t_{i-1}) \\
&~~~~~ - \sum_{i : t_i\notin T_{out}} \mathbf{I}_{\{\phi'\left( \int_{t_{i-1}}^{t_i} \lambda^{\ast}(u) d u-1\right)>\widetilde\alpha\}} (t_i-t_{i-1})\Big]/T_{out}\\
&\qquad\text{( $\lambda'(t)=\lambda^{\ast}(t)$ when $t\notin T_{out}$, and the length of $T_{out}$ is $\eta_n\cdot T_0$)}\\
&=1-(\text{IW}(\lambda'; \lambda^{\ast})-(1-\eta_n)\text{IW}(\lambda^{\ast}; \lambda^{\ast}))/\eta_n T_0
\end{align}
and 
$$\mathbb{E}[TNR(n)] = \text{IW}(\lambda^{\ast}; \lambda^{\ast})/T_0,$$
where $\eta_n$ represents the proportion of time that sample $S_n$ is contaminated by outliers. 

Referring to Proposition \ref{thm:tprfnr}, {when $\lambda_n'(t)=\lambda_{miss}(t) \downarrow 0$ or $\lambda_n'(t)=\lambda_{add}(t) \uparrow \infty$ for $t\in T_{out}$, {we know that $\text{IW}(\lambda'; \lambda^{\ast})\rightarrow (1-\eta_n)\text{IW}(\lambda^{\ast}; \lambda^{\ast})$}. Then $\mathbb{E}[TPR(n)] \rightarrow 1$} according to \eqref{eq:tprnow}, and $\widehat{TP} = N^{-1}\sum_{n} TPR(n)\rightarrow 1$ . 


\section{Proof of Supporting Lemmas}
\label{sec:pf:lem}

\noindent \textbf{Proof of Lemma \ref{lem:robust_exp}}
Notice that 
\begin{align*}
    \mathbb{E}_X  \left[(X-1)\cdot\phi'(X-1) \right]=\mathbb{E}_X  \left[(X-1)\cdot\phi'(X-1)\mid X>1 \right]-\mathbb{E}_X  \left[(1-X)\cdot\phi'(X-1)\mid X<1 \right].
\end{align*}
We only need to prove that $ \int_1^\infty (x-1)\exp(-x)\cdot \phi'(x-1) dx =  \int_0^1 (1-x)\exp(-x)\cdot \phi'(x-1) dx$.
According to the definition of $\phi(\cdot)$, it's easy to know that $\int_1^\infty (t-1)\exp(-t)\cdot \phi'(t-1) dt=\int_1^\infty \phi'(t-1) d(-t\exp(-t)) =\int_0^{\exp(-1)} \phi'(t_1(y)-1) dy$, where $t_1(y):=\{t\mid t\geq 1, t\exp(-t)=y\}$. 

Similarly, we also have $\int_0^1 (t-1)\exp(-t)\cdot \phi'(t-1) dt=\int_0^1 \phi'(t-1) d(-t\exp(-t)) =\int_0^{\exp(-1)} \phi'(t_2(y)-1) dy$, where $t_2(y):=\{t\mid 0\leq t\leq1, t\exp(-t)=y\}$. 
Referring to the definition in \eqref{eq:robust_fun}, it's easy to know that $\phi'(t_1(y)-1)= \phi'(t_2(y)-1)$, $\forall 0\leq y\leq\exp(-1)$. Then $\int_0^{\exp(-1)} \phi'(t_1(y)-1) dy= \int_0^{\exp(-1)} \phi'(t_2(y)-1) dy$, and we can get the result. 

\medskip

\noindent \textbf{Proof of Lemma \ref{lem:lambda}}

We consider the case of k=1, the second derivative is
\begin{align*}
&-\sum_{i=1}^{M }\begin{bmatrix}
        \frac{\kappa_1(S_{j,i})}{\lambda_{\boldsymbol{B}_1}(S_{j,i})\cdot L}\\
        \vdots \\
       \frac{\kappa_H(S_{j,i})}{\lambda_{\boldsymbol{B}_1}(S_{j,i})\cdot L}
    \end{bmatrix}
    \cdot
    \begin{bmatrix}
        \phi'\left(\int_{t_{i-1}}^{t_i} \lambda_k^*(u) d u-1\right)\frac{\kappa_1(S_{j,i})}{\lambda_{\boldsymbol{B}_1}(S_{j,i})\cdot L}\\
        \vdots \\
       \phi'\left(\int_{t_{i-1}}^{t_i} \lambda_k^*(u) d u-1\right)\frac{\kappa_H(S_{j,i})}{\lambda_{\boldsymbol{B}_1}(S_{j,i})\cdot L}
    \end{bmatrix}^\top
    \\
&-\sum_{i=1}^{M }\begin{bmatrix}
        \frac{\kappa_1(S_{j,i})}{\lambda_{\boldsymbol{B}_1}(S_{j,i})\cdot L}-\int_{t_{i-1}}^{t_i}\kappa_1(x)dx\\
        \vdots \\
       \frac{\kappa_H(S_{j,i})}{\lambda_{\boldsymbol{B}_1}(S_{j,i})\cdot L}-\int_{t_{i-1}}^{t_i}\kappa_H(x)dx
    \end{bmatrix}
    \cdot
    \begin{bmatrix}
        \phi''\cdot(\frac{\kappa_1(S_{j,i})}{\lambda_{\boldsymbol{B}_1}(S_{j,i})\cdot L}-\int_{t_{i-1}}^{t_i}\kappa_1(x)dx)\\
        \vdots \\
       \phi''\cdot(\frac{\kappa_H(S_{j,i})}{\lambda_{\boldsymbol{B}_1}(S_{j,i})\cdot L}-\int_{t_{i-1}}^{t_i}\kappa_H(x)dx)
    \end{bmatrix}^\top \\
    &:=-H_0(S,\boldsymbol{B}_1)-H_1(S,\boldsymbol{B}_1)
\end{align*}

When outliers are not considered, we only need to prove that $\mathbb{E}\left[\phi''(X-1)\cdot(1-X)^2\right]+\mathbb{E}\left[\phi'(X-1)\right]>0$ referring to the proof of Theorem 1. We note that
$\mathbb{E}\left[\phi''(X-1)\cdot(1-X)^2\right]=\int_0^\infty (x-1)^2e^{-x}\phi''(x-1)dx=\bigg|_0^\infty ((x-1)^2e^{-x}\cdot \phi'(x-1))+\int_0^\infty (x^2-4x+3)e^{-x}\cdot \phi'(x-1)dx$. Then $\mathbb{E}\left[\phi''(X-1)\cdot(1-X)^2\right]+\mathbb{E}\left[\phi'(X-1)\right]=\int_0^\infty (x^2-4x+4)e^{-x}\cdot \phi'(x-1)dx>0$.

Referring to Lemma \ref{lem:exp1}, we know that the second derivative must be negative definite matrix when $\rho_1,\rho_2<\infty$. We write $\mathbb {E}[H_0(S,\boldsymbol{B}_1)]=(\gamma_1-1)\mathbb {E}[H_1(S,\boldsymbol{B}_1)]$, where $\gamma_1:=\int_0^\infty (x-2)^2 e^{-x}\cdot \phi'(x-1)dx/\int_0^\infty e^{-x}\cdot \phi'(x-1)dx$.

When we consider outliers, define $\chi(\lambda):=\mathbb{E}\left[\phi''(X-1)\cdot(1-X)^2\right]+\mathbb{E}\left[\phi'(X-1)\right]=\int_0^\infty \lambda(\lambda x^2-2(\lambda+1)x+(\lambda+3)) e^{-\lambda x}\phi'(x-1)dx$. Referring to Lemma \ref{lem:exp1}, there are outliers when $\lambda\neq 1$. It's easy to know that $\chi(\lambda)=\int_0^\infty ((\lambda x-\lambda-1)^2 +\lambda-1)e^{-\lambda x}\phi'(x-1)dx>0 $ when $\lambda\geq 1$. Obviously the function also has a lower bound $\chi_{low}$ when $\lambda<1$. Thus, for a small enough $\eta$ such that $\chi(1)\cdot(1-\eta)+\chi_{low}\eta >0 $, we can still maintain the convex property under Assumption \ref{asm:5}.

\begin{lemma}\label{lem:B_bar}
    There exist $\|\boldsymbol{\overline B}_k-\boldsymbol{B}_k^\ast\|=O(\sqrt{H}/L)$ such that $\|g(\boldsymbol{\overline B}_k\mid \boldsymbol{\overline B}_k)\|=0$.
\end{lemma}

\noindent \textbf{Proof of Lemma \ref{lem:B_bar}}

Referring to Theorem \ref{Thm:Mgrad}, we know that $ \|g(\boldsymbol{B}_k^*\mid \boldsymbol{B}_k^*)\| = O(\sqrt{H}/L) $, and the smallest eigenvalue of $-\nabla_{\boldsymbol{B}_k} g(\boldsymbol{B}_k\mid \boldsymbol{B}_k^*)$ has a low bound $\lambda_{k,min}>0$ according to lemma \ref{lem:lambda}. 

Due to the continuity of the function $\phi''(\cdot)$, we know that $\|\nabla_{\boldsymbol{B}_k} g(\boldsymbol{B}_k\mid \boldsymbol{B}_k)-\nabla_{\boldsymbol{B}_k} g(\boldsymbol{B}_k^*\mid \boldsymbol{B}_k^*)\|=O(\sqrt{H}/L)$ when $\|\boldsymbol{ B}_k-\boldsymbol{B}_k^\ast\|=O(\sqrt{H}/L)$.
Then we have $\|g(\boldsymbol{B}_k\mid \boldsymbol{B}_k)\| =\| \int \nabla_{\boldsymbol{B}_k} g(\boldsymbol{B}_k\mid \boldsymbol{B}_k) d \boldsymbol{B}_k + g(\boldsymbol{B}_k^*\mid \boldsymbol{B}_k^*) \|=  \|\nabla_{\boldsymbol{B}_k} g(\boldsymbol{B}_k^*\mid \boldsymbol{B}_k^*)\cdot (\boldsymbol{B}_k-\boldsymbol{B}_k^*)+g(\boldsymbol{B}_k^*\mid \boldsymbol{B}_k^*)\|+O(\sqrt{H}/L^2)$.
Thus, $\|g(\boldsymbol{B}_k\mid \boldsymbol{B}_k)\| = O(\sqrt{H}/L^2) $ when $ \boldsymbol{B}_k-\boldsymbol{B}_k^* = -(\nabla_{\boldsymbol{B}_k} g(\boldsymbol{B}_k^*\mid \boldsymbol{B}_k^*))^{-1}\cdot g(\boldsymbol{B}_k^*\mid \boldsymbol{B}_k^*)$, and we get that $ \|\boldsymbol{B}_k-\boldsymbol{B}_k^*\|=O(\sqrt{H}/L)$.

Repeating the above steps, we can get $\boldsymbol{\overline B}_k$ such that $\|g(\boldsymbol{\overline B}_k\mid \boldsymbol{\overline B}_k)\|=0$, where 

$\|\boldsymbol{\overline B}_k-\boldsymbol{B}_k^\ast\| \leq O(\sqrt{H}/L)+O(\sqrt{H}/L^2)+\cdots =O(\sqrt{H}/L)$ as $L\rightarrow\infty$.

\begin{lemma}\label{lem:exp1}
   (Random time change theorem (\cite{brown2002time})). A sequence $X=\left(t_1, \ldots, t_N\right)$ is distributed according to a TPP with compensator $\Lambda^*$ on the interval $[0, T]$ if and only if the sequence $Z=\left(\Lambda^*\left(t_1\right), \ldots, \Lambda^*\left(t_N\right)\right)$ is distributed according to the standard Poisson process on $\left[0, \Lambda^*(T)\right]$.
\end{lemma}

\begin{lemma}[Bernstein's inequality \citep{vershynin2018high}]\label{lemma:Bern}
Let $X_1, \ldots, X_N$ be independent, mean zero, sub-exponential random variables, and $a=\left(a_1, \ldots, a_N\right) \in \mathbb{R}^N$. Then, for every $t \geq 0$, we have
$$
\mathbb{P}\left(\left|\sum_{i=1}^N a_i X_i\right| \geq t\right) \leq 2 \exp \left[-c \min \left(\frac{t^2}{K^2\|a\|_2^2}, \frac{t}{K\|a\|_{\infty}}\right)\right],
$$
where $K=\max _i\left\|X_i\right\|_{\psi_1}$ and $\|X\|_{\psi_1}:=\inf \{t>0: \mathbb{E} \exp (|X| / t) \leq 2\}$.
\end{lemma}

\begin{lemma}\label{lem:log-fun}
  When event sequence $S$ is sampled from the NHP process with parameter $\lambda_*$, with any weight $W\neq0$, its log-likelihood function $\log \operatorname{WTPP}(S\mid \boldsymbol{B}_i,W)$ follows a sub-exponential distribution.
\end{lemma}

\noindent \textbf{Proof of Lemma \ref{lem:log-fun}}

Due to the weight function $W_i<1$, referring to Lemma 7 from \cite{zhang2024robust} and using Lemma \ref{lemma:Bern}, we know that 
\begin{align*}
    \mathbb{P}\left(\left|\log \operatorname{WTPP}(S\mid \boldsymbol{B}_i,W)/L-\mu_{avg}\right| \geq t\right) \leq 2 \exp \left[-c \min \left(\frac {L^2 t^2}{C^2\max\log(\lambda_{*})^2}, \frac{Lt}{C\max\log(\lambda_{*})}\right)\right],
\end{align*}
where $C$ is a finite constant depend on $\boldsymbol{B}_i$ and $W$, and $\mu_{avg}:=\mathbb{E}_{S\sim \lambda_*} \left[\log \operatorname{WTPP}(S\mid \boldsymbol{B}_i,W)/L\right]$.

Similar to the derivative function of $\log \operatorname{WTPP}(S\mid \boldsymbol{B}_i,W)$, there is
\begin{align*}
    \mathbb{P}\left(\left|\frac{\partial\log \operatorname{WTPP}(S\mid \boldsymbol{B}_i,W)}{\partial \boldsymbol{B}_i}/L-\mu_{avg}^1\right| \geq t\right) \leq 2 \exp \left[-c \min \left(\frac {L^2 t^2}{C^2(\max\frac{\kappa_{\max}}{\lambda_{*}(t)})^2}, \frac{Lt}{C\max\frac{\kappa_{\max}}{\lambda_{*}(t)}}\right)\right],
\end{align*}
where $\mu_{avg}^1:=\mathbb{E}_{S\sim \lambda_*} \left[\frac{\partial\log \operatorname{WTPP}(S\mid \boldsymbol{B}_i,W)}{\partial \boldsymbol{B}_i}/L\right].$

\begin{lemma}\label{thm:Estep}
Under assumption \ref{asm:4}, when $S$ is sampled from cluster $k$, for $j\neq k$,
if outlier events occur only for a length of time $\eta\cdot LT$, then there exist $C_\eta>0$ such that
\begin{align*}
\log\operatorname{WTPP}(S\mid \boldsymbol{B}_j, w(S;\boldsymbol B_k))\leq\log\operatorname{WTPP}(S\mid \boldsymbol{B}_k, w(S;\boldsymbol B_k))-C_{\eta}L
\end{align*}
with a high probability $1-4\exp(-C_{gap}L/C_{wtpp})$ at $\boldsymbol{B}_k=\boldsymbol{B}_k^*, \forall k\in [K]$. Here $C_{wtpp}$ is a constant only depending on $\rho$.
\end{lemma}

\noindent \textbf{Proof of Lemma \ref{thm:Estep}}

We first consider the situation when $\eta=0$. Referring to lemma \ref{lem:exp1}, we note that $\log\lambda_{\boldsymbol{B}_k}(t_i)$ and $\int_{t_{i-1}}^{t_i}\lambda_{\boldsymbol{B}_k}(s)ds$ are independent of each other, so it is easy to obtain that $\log\lambda_{\boldsymbol{B}_k}(t_i)$ and $w(S;\boldsymbol B_k):=\phi'(\int_{t_{i-1}}^{t_i}\lambda_{\boldsymbol{B}_k}(s)ds-1)$ are also independent. Referring to lemma \ref{lem:robust_exp}, we have 
\begin{align*}
\mathbb{E}\left[\log \operatorname{WTPP}(S\mid \boldsymbol{B}_k, w(S;\boldsymbol B_k))\right]&=\mathbb{E}\left[\sum_{i=1}^M  w_i(S;\boldsymbol B_k)\cdot\left(\log\lambda_{\boldsymbol{B}_k}(t_i)-\int_{t_{i-1}}^{t_i}\lambda_{\boldsymbol{B}_k}(s)ds\right)\right]\\
&=\mathbb{E}\left[\sum_{i=1}^M w_i(S;\boldsymbol B_k)\cdot\log\lambda_{\boldsymbol{B}_k}(t_i)\right]+\mathbb{E}\left[\sum_{i=1}^M  w_i(S;\boldsymbol B_k)\cdot\int_{t_{i-1}}^{t_i}\lambda_{\boldsymbol{B}_k}(s)ds\right]\\
&=\mathbb{E}\left[\phi'(X-1)\right]\cdot\mathbb{E}\left[\sum_{i=1}^M\log\lambda_{\boldsymbol{B}_k}(t_i)\right]+\mathbb{E}\left[\phi'(X-1)\cdot X\right]\cdot\mathbb{E}\left[\int_{0}^{LT}\lambda_{\boldsymbol{B}_k}(s)ds\right]\\
&=u_0(\phi')\cdot\mathbb{E}\left[\sum_{i=1}^M\log\lambda_{\boldsymbol{B}_k}(t_i)\right]+u_1(\phi')\cdot\mathbb{E}\left[\int_{0}^{LT}\lambda_{\boldsymbol{B}_k}(s)ds\right],
\end{align*}
where $X\sim \text{Exp}(1)$ and $u_0:=\mathbb{E}\left[\phi'(X-1)\right], u_1:=\mathbb{E}\left[\phi'(X-1)\cdot X\right]$.
When $\rho_1=\rho_2$, we have $u_0=u_1$, and then $\mathbb{E}[\log \operatorname{WTPP}(S\mid \boldsymbol{B}_k, w(S;\boldsymbol B_k))]=u_0\cdot \mathbb{E}[\log \operatorname{WTPP}(S\mid \boldsymbol{B}_k)]$. Similarly, $\mathbb{E}[\log \operatorname{WTPP}(S\mid \boldsymbol{B}_j, w(S;\boldsymbol B_k))]=u_0\cdot \mathbb{E}[\log \operatorname{WTPP}(S\mid \boldsymbol{B}_j)]$. Then we have $\mathbb{E}[\log\operatorname{WTPP}(S\mid \boldsymbol{B}_j, w(S;\boldsymbol B_k))]<\mathbb{E}[\log\operatorname{WTPP}(S\mid \boldsymbol{B}_k, w(S;\boldsymbol B_k))]-C_{gap}L\cdot u_0$.

Referring to Lemma \ref{lem:log-fun}, we know that $\log \operatorname{WTPP}(S\mid \boldsymbol{B}_j, w(S;\boldsymbol B_k))/L$ follows a sub-exponential distribution for all $j,k\in[K]$. For given $\rho$, there exists $C_{wtpp}>0$ such that $\forall j,k \in [K]$,
\begin{align*}
\mathbb{P}\left[\left|\log \operatorname{WTPP}(S\mid \boldsymbol{B}_j, w(S;\boldsymbol B_k))/L-\mu_{j,k}\right|>t\right]<2\exp(-tL/ (u_0 C_{wtpp})),
\end{align*}
where $\mu_{j,k}=\mathbb{E}[\log \operatorname{WTPP}(S\mid \boldsymbol{B}_j, w(S;\boldsymbol B_k))/L]$.
Taking $t=C_{gap}\cdot u_0$, we get that
\begin{align*}
\log\operatorname{WTPP}(S\mid \boldsymbol{B}_j, w(S;\boldsymbol B_k))/L<\log\operatorname{WTPP}(S\mid \boldsymbol{B}_k, w(S;\boldsymbol B_k))/L-C_{gap}\cdot u_0/2
\end{align*}
with probability at least $1-4\exp(-C_{gap}L/C_{wtpp})$, and $1-4\exp(-C_{gap}L/C_{wtpp})\rightarrow 1$ as $L\rightarrow 0$.

Next we consider the situation when $\eta>0$.
If $S$ is sampled from cluster $k$, for any $j\neq k$ we have
\begin{align*}
&\log \operatorname{WTPP}({S} \mid \boldsymbol{B}_k, W)-\log \operatorname{WTPP}({S} \mid \boldsymbol{B}_j, W)\\
=& (\sum_{t_i\in T_{in}} +\sum_{t_i\in T_{out}})W_i\left(\log\lambda_{k}^*(t_i)-\int_{t_{i-1}}^{t_i} \lambda_k^*(u) d u  - \log\lambda_{j}^*(t_i)+\int_{t_{i-1}}^{t_i} \lambda_j^*(u) d u \right)\\
\geq&(1-\eta)C_{gap}L u_0/2+\sum_{t_i\in T_{out}}W_i\left(\log\lambda_{k}^*(t_i)-\int_{t_{i-1}}^{t_i} \lambda_k^*(u) d u  - \log\lambda_{j}^*(t_i)+\int_{t_{i-1}}^{t_i} \lambda_j^*(u) d u \right)\\
\geq&(1-\eta)C_{gap}L u_0/2-\eta\int_0^{LT} \lambda_k^*(u) d u+\sum_{t_i\in T_{out}}W_i\left(\log\lambda_{k}^*(t_i)  - \log\lambda_{j}^*(t_i) \right)\\
\geq&(1-\eta)C_{gap}L u_0/2-\eta\int_0^{LT} \lambda_k^*(u) d u+\eta \min\{0, c'_{up}\int_0^T \log(\lambda_k^*(u)/\lambda_j^*(u)) d u\},
\end{align*}
where $\phi'(x-1)/x\leq c'_{up}$.
Thus we can define $C_\eta := (1-\eta)C_{gap} u_0/2-\eta\int_0^T \lambda_k^*(u) d u+\eta \min\{0, c'_{up}\int_0^T \log(\lambda_k^*(u)/\lambda_j^*(u)) d u\} $.

\begin{lemma}\label{col:4}
When $\|\boldsymbol{B}^{(t)}_k-\boldsymbol{B}_k^*\| <a/(T\cdot\kappa_{\max}), \forall k = 1,\cdots,K$, we get {\small $\log\operatorname{WTPP}(S\mid \boldsymbol{B}_j^{(t)}, w(S;\boldsymbol B_k^{(t)}))\leq\log\operatorname{WTPP}(S\mid \boldsymbol{B}_k^{(t)}, w(S;\boldsymbol B_k^{(t)}))-C_{\eta}L+a/T\kappa_{\max}\cdot \max_{\|\boldsymbol{B}-\boldsymbol{B}^*\|<a/(T\cdot\kappa_{\max})}\|\frac{\partial\log\operatorname{WTPP}(S\mid \boldsymbol{B}_j, w(S;\boldsymbol B_k))}{\partial \boldsymbol{B}_j}\|$}, so there exists $a_r>0$ such that $|r_k(S ; \boldsymbol{B}^{(t)})- r_k^*(S )|<a_r$ with a high probability as $L\rightarrow \infty$, where $r_k^*(S)=1$ when $S$ is sampled from cluster $k$.
\end{lemma}
Similarly to the proof of \ref{thm:Estep},  when $S$ is sampled from cluster $k$, we have $\mathbb{E}[\log\operatorname{WTPP}(S\mid \boldsymbol{B}_k^*, w(S;\boldsymbol B_j^*))]<\mathbb{E}[\log\operatorname{WTPP}(S\mid \boldsymbol{B}_j^*, w(S;\boldsymbol B_j^*))], \forall j\neq k$.

\end{document}

%% file: macro.tex
\newcommand{\bma}{\begin{matrix*}[r]}
\newcommand{\ema}{\end{matrix*}}